\documentclass[twoside]{article}

\usepackage{PRIMEarxiv}

\usepackage[utf8]{inputenc} 
\usepackage[T1]{fontenc}    
\usepackage{hyperref}       
\usepackage{url}            
\usepackage{booktabs}       
\usepackage{amsfonts}       
\usepackage{nicefrac}       
\usepackage{microtype}      
\usepackage{lipsum}
\usepackage{fancyhdr}       
\usepackage{graphicx}       
\graphicspath{{media/}}     

\usepackage{amsmath}
\usepackage{tikz}
\usepackage{mathdots}
\usepackage{yhmath}
\usepackage{cancel}
\usepackage{color}
\usepackage{siunitx}
\usepackage{array}
\usepackage{multirow}
\usepackage{amssymb}
\usepackage{textcomp}
\usepackage{gensymb}
\usepackage{tabularx}
\usepackage{extarrows}
\usepackage{booktabs}
\usepackage{float}
\usetikzlibrary{fadings}
\usetikzlibrary{patterns}
\usetikzlibrary{shadows.blur}
\usetikzlibrary{shapes}
\usepackage{adjustbox}
\usepackage{comment}

\title{DiffMorph: Text-less Image Morphing with Diffusion Models 
}

\author{
  Shounak Chatterjee\\
    Kolkata, India\\
{\tt\small shounak.thirti2@gmail.com} \\
}

\begin{document}
\maketitle
\begin{abstract}
Text-conditioned image generation models are a prevalent use of AI image synthesis, yet intuitively controlling output guided by an artist remains challenging. 
Current methods require multiple images and textual prompts for each object to specify them as concepts to generate a single customized image. 

On the other hand, our work, \verb|DiffMorph|, introduces a novel approach that synthesizes images that mix concepts without the use of textual prompts. Our work integrates a sketch-to-image module to incorporate user sketches as input. \verb|DiffMorph| takes an initial image with conditioning artist-drawn sketches to generate a morphed image. 

We employ a pre-trained text-to-image diffusion model and fine-tune it to faithfully reconstruct each image. 
We seamlessly merge images and concepts from sketches into a cohesive composition. 
The image generation capability of our work is demonstrated through our results and a comparison of these with prompt-based image generation.
\end{abstract} 
\keywords{Image Diffusion Model \and Image editing \and Sketch-to-Image}
\begin{figure*}[!ht]
    \centering
    \begin{tabular}{rccc}
         \begin{tabular}{c}
             Sketch Condition $\rightarrow$ \\
             \\
            Input Image$\downarrow$ \\
         \end{tabular}    
         & \includegraphics[width=0.22\textwidth]{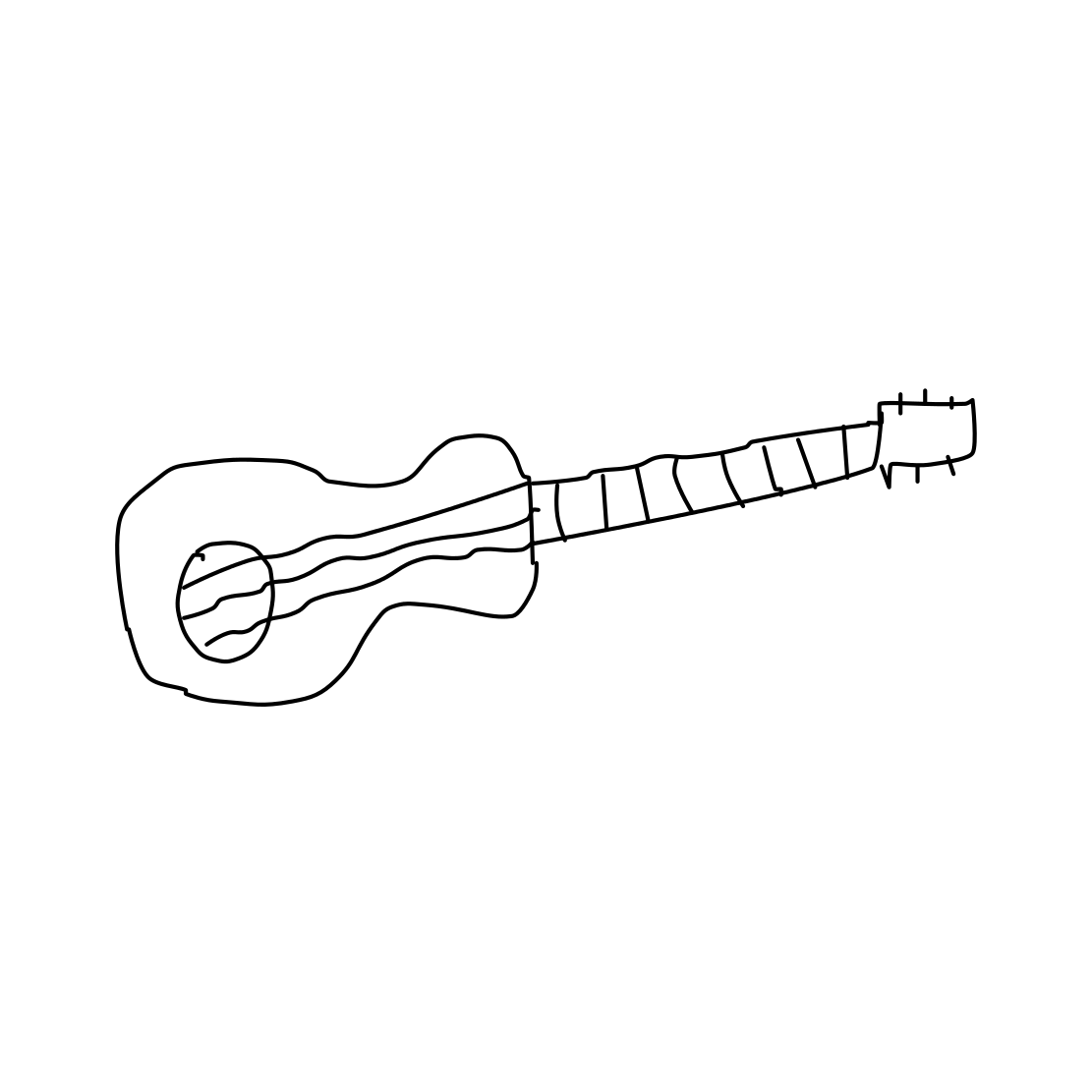} 
         & \includegraphics[width=0.22\textwidth]{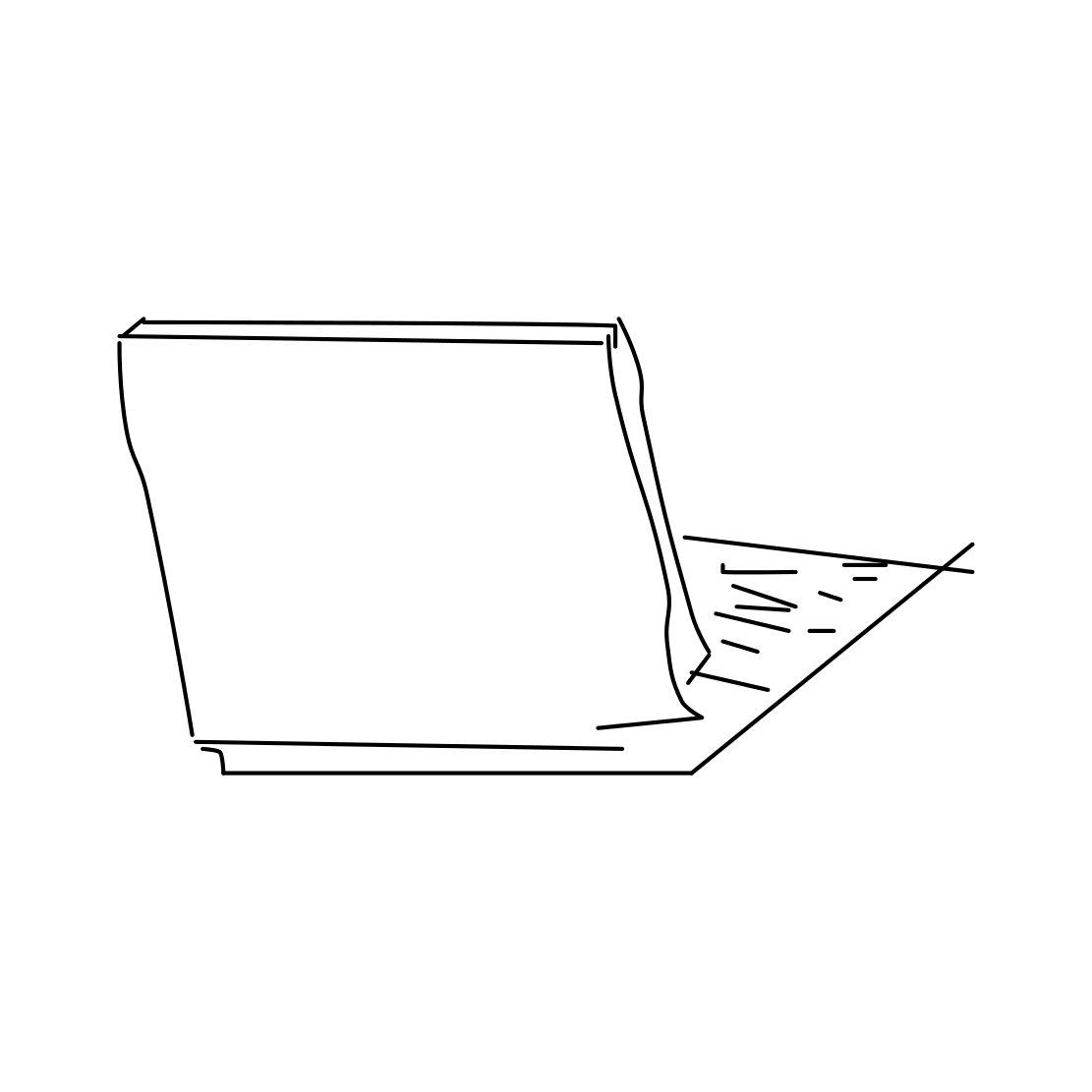} 
         & \includegraphics[width=0.22\textwidth]{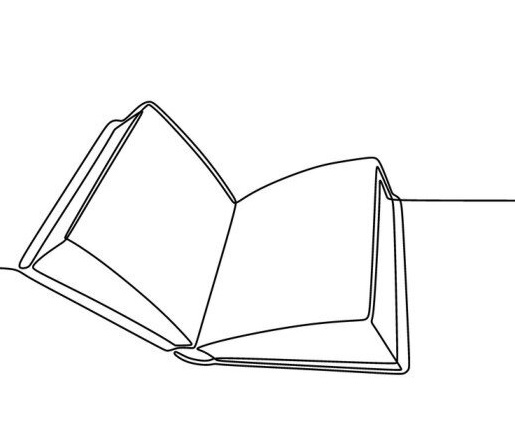}\\
         \includegraphics[width=0.22\textwidth]{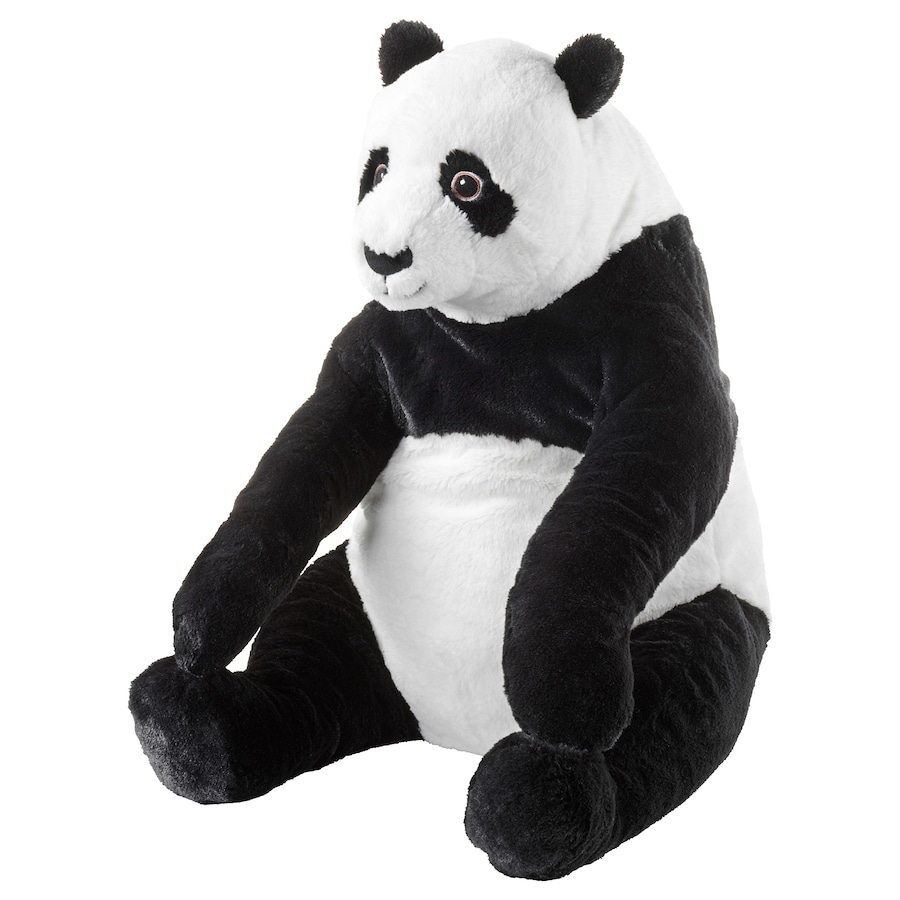}
         & \includegraphics[width=0.22\textwidth]{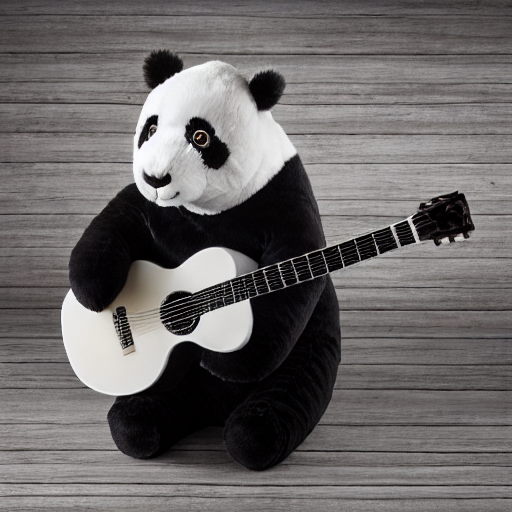} 
         & \includegraphics[width=0.22\textwidth]{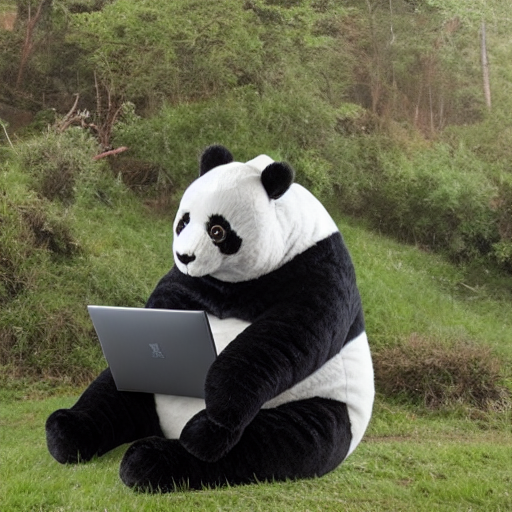} 
         & \includegraphics[width=0.22\textwidth]{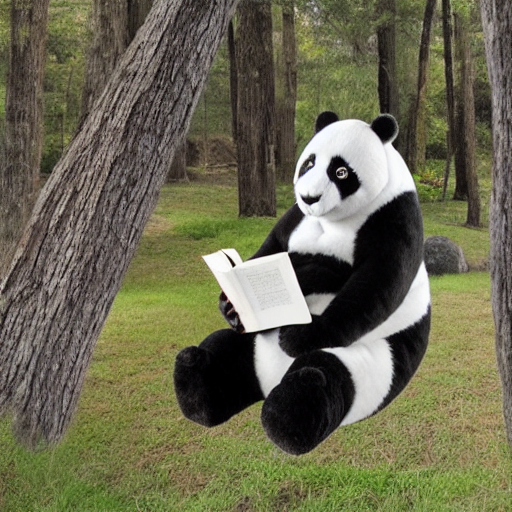}
    \end{tabular}
    \caption{Our method, DiffMorph. Users can add a sketch of any object i.e. concept on top of an Image to generate a new image combined with the sketch. Here we used a panda plushie as a primary concept and added secondary concepts as sketches to transform the image with the provided sketches.}
    \label{fig:Banner}
\end{figure*}
 
\section{Introduction}
Recent advances in the field of large text-to-image models~\cite{dhariwal2021diffusion, kingma2021variational, Rombach_2022_CVPR}, have unequivocally showcased their capabilities in generating images of high fidelity and diversity from text prompts. 
Notwithstanding this diversity, these models exhibit limitations in faithfully reproducing specific objects or \textit{concepts} without careful prompt engineering.
Researchers have sought to enhance the customization capabilities of these models to generate specific concepts~\cite{ruiz2023dreambooth, kumari2023multi}. 
However, this prior work has been limited in terms of the consideration of integrating multiple concepts into the model customization process. 
Common to all these prior research efforts is the prerequisite of employing multiple images corresponding to the targeted concept for effective model customization. 
While some methodologies can function with a single image, their efficacy is limited compared to approaches designed to leverage multiple images.
Recent techniques predominantly adopt one of two principal approaches for model customization; they either represent a concept through a word embedding provided to the text encoder as input~\cite{cohen2022my} or, alternatively, fine-tune the complete weights of a diffusion-based denoiser module~\cite{ruiz2023dreambooth}. 
The former approach is susceptible to challenges in generalizing to unseen prompts, whereas the latter lacks the expressive capacity to generate intricate images.

Despite recent advances, certain features are notably absent in existing methods, particularly with regard to the demand for multiple images for each concept. 
Moreover, the prevalent reliance on text prompts for conditioning image generation in most models remains a limitation to artistic expression. 
This work aims to address these challenges.
Our approach involves the fine-tuning of the diffusion denoiser model to accommodate multiple concepts in a controlled manner, thereby mitigating the risk of overfitting.
Notably, our system stands out by necessitating only one image for each concept, eliminating the conventional requirement for multiple images for individual concepts. 
Furthermore, our approach conditions image generation based on the provided concepts and their relation with each other.

In the proposed system, referred to as \textit{"DiffMorph,"} users initially input an image as a primary concept.
Subsequently, one or more sketches are added as secondary concepts. 
An exclusive sketch-to-image module is incorporated to convert sketches into visually similar images. 
Users also have the option to include images as secondary concepts instead of sketches.
This is an optional feature if the user just wants to morph two images simultaneously.
The system then identifies the classes of the added concepts, generating an embedding that fulfills the requirements of the text prompt. 
Fine-tuning of the pre-trained text-to-image diffusion model is executed with input concept images and their corresponding classes to reconstruct the provided image. 
The fine-tuning process is guided by the area coverage of each concept, with the exception of regularization images, as a preventive measure against overfitting.
This step enables morphing more than two concepts into a single image, as shown in Figure~\ref{fig:Banner}.

In summary, our novel contributions to this work are as follows:
\begin{enumerate}
    \item We investigated the reasons for over-fitting in the fine-tuning of the stable diffusion-based denoiser module and proposed a controlled fine-tuning method.
    \item We proposed a precise sketch-to-image generation model to convert sketches to images to use them as a concept.
    \item Our method works without text prompts and generates images based on the supplied concepts contained in the initial image and sketches.
\end{enumerate}

\section{Related Work}
We discuss the relevant literature on prior methods for image-based diffusion models and sketch-to-image generation models.
\subsection{Diffusion Models}
The concept of Diffusion Models was first proposed by Sohl-Dickstein et al., \cite{pmlr-v37-sohl-dickstein15} and has subsequently been employed widely in current image generation methods \cite{dhariwal2021diffusion, kingma2021variational, balaji2022ediffi, nichol2021glide, ramesh2022hierarchical}.
Diffusion models proved themselves as a powerful generation method that can handle images, text, audio, etc.
These models learn data distribution by executing a Markov Chain~\cite{chung1967markov}, which simulates the reverse data generation process.
The Latent Diffusion Model (LDM) \cite{Rombach_2022_CVPR} performs diffusion operations within the latent image domain \cite{esser2021taming}.
In LDM, instead of performing a diffusion process on the images, the researchers converted the images in low-resolution latent codes.
This conversion helps to reduce the computation cost during training and inference.
Text-to-image diffusion models have set a benchmark performance in image generation by translating text inputs into latent vectors using pre-trained language models, such as CLIP \cite{radford2021learning}.
The model Glide \cite{nichol2021glide} is an example of a text-driven diffusion model, facilitating both image generation and editing.
The Stable Diffusion approach \cite{podell2023sdxl} represents a comprehensive implementation of latent diffusion \cite{Rombach_2022_CVPR}.
Meanwhile, Imagen \cite{NEURIPS2022_ec795aea} conducts pixel diffusion leveraging a pyramid structure, bypassing the need for latent images.
On the commercial front, noteworthy applications include DALL-E2 \cite{dalle} and Midjourney \cite{midj}.

\subsection{Customised Image Generation}
However, a limitation of the diffusion model is its inability to offer granular control over the resultant image.
The image generation process relies solely on text guidance \textit{prompt engineering} to steer the artistic process. 
For a consistent generation of analogous images, it is imperative to instruct a generative Text-2-Image model to synthesize novel images targeting a specific concept, guided by unstructured language—essentially, customizing the model. 
Current customization strategies either refine the denoising network centered around a stable embedding \cite{ruiz2023dreambooth, ruiz2023hyperdreambooth, kawar2023imagic, kumari2023multi} or enhance a collection of word embedding to elucidate the concept \cite{daras2022multiresolution, gal2022image, cohen2022my}. 
An alternate method entails anchoring the supercategory of the concept and employing gated rank-1 editing \cite{tewel2023key}, thus bypassing the need for detailed fine-tuning and subsequent optimization. 
For artists, sketches often resonate more deeply than textual prompts in the artistic process. 
Such sketches afford a more definitive portrayal of specific objects as envisaged by the artist. 
Our method focuses on conditioning image generation using both a submitted image and artistic direction specified as sketch references.

\subsection{Sketch-to-Image Translation}
Conditional Generative Adversarial Networks (GANs) \cite{choi2018stargan, isola2017image, park2019semantic} and transformers \cite{chen2021pre, esser2021taming, ramesh2021zero} possess the capability to recognise the difference between various image domains.
For instance, the Taming Transformer \cite{esser2021taming} adopts a vision transformer methodology, whereas Palette \cite{saharia2022palette} represents a conditional diffusion model developed from scratch. 
Further, the manipulation of pre-trained GANs facilitates the management of specialized image-to-image operations. 
This can be observed with a StyleGAN that can be modulated via supplementary encoders \cite{richardson2021encoding}. 
An extended exploration of these applications can be found in \cite{gal2022stylegan, karras2019style, patashnik2021styleclip}.
ControlNet \cite{zhang2023adding} has retrained the diffusion model by integrating additional conditioning controls, enhancing the precision of photo generation.

\begin{figure*}[h]
    \centering
    \tikzset{every picture/.style={line width=0.75pt}} 
    
    \begin{tikzpicture}[x=0.75pt,y=0.75pt,yscale=-1,xscale=1]
    
    \draw  [fill={rgb, 255:red, 238; green, 168; blue, 168 }  ,fill opacity=1 ] (385,19.2) .. controls (385,14.12) and (389.12,10) .. (394.2,10) -- (525.8,10) .. controls (530.88,10) and (535,14.12) .. (535,19.2) -- (535,160.8) .. controls (535,165.88) and (530.88,170) .. (525.8,170) -- (394.2,170) .. controls (389.12,170) and (385,165.88) .. (385,160.8) -- cycle ;
    \draw   (20,10) -- (90,10) -- (90,80) -- (20,80) -- cycle ;
    \draw   (20,10) -- (90,10) -- (90,95) -- (20,95) -- cycle ;
    
    \draw   (20,104.67) -- (90,104.67) -- (90,174.67) -- (20,174.67) -- cycle ;
    \draw   (20,104.67) -- (90,104.67) -- (90,189.67) -- (20,189.67) -- cycle ;
    
    \draw  [fill={rgb, 255:red, 238; green, 168; blue, 168 }  ,fill opacity=1 ] (310,85) -- (350,70) -- (350,130) -- (310,115) -- (270,130) -- (270,70) -- cycle ;
    \draw   (580,80) -- (679,80) -- (679,170.01) -- (580,170.01) -- cycle ;
    \draw   (580,80) -- (679,80) -- (679,189.3) -- (580,189.3) -- cycle ;
    
    \draw    (90,65) -- (128,65) ;
    \draw [shift={(130,65)}, rotate = 180] [color={rgb, 255:red, 0; green, 0; blue, 0 }  ][line width=0.75]    (10.93,-3.29) .. controls (6.95,-1.4) and (3.31,-0.3) .. (0,0) .. controls (3.31,0.3) and (6.95,1.4) .. (10.93,3.29)   ;
    \draw    (90,135) -- (128,135) ;
    \draw [shift={(130,135)}, rotate = 180] [color={rgb, 255:red, 0; green, 0; blue, 0 }  ][line width=0.75]    (10.93,-3.29) .. controls (6.95,-1.4) and (3.31,-0.3) .. (0,0) .. controls (3.31,0.3) and (6.95,1.4) .. (10.93,3.29)   ;
    \draw    (190,135) -- (220,135) -- (220,112) ;
    \draw [shift={(220,110)}, rotate = 90] [color={rgb, 255:red, 0; green, 0; blue, 0 }  ][line width=0.75]    (10.93,-3.29) .. controls (6.95,-1.4) and (3.31,-0.3) .. (0,0) .. controls (3.31,0.3) and (6.95,1.4) .. (10.93,3.29)   ;
    \draw    (190,65) -- (220,65) -- (220,88) ;
    \draw [shift={(220,90)}, rotate = 270] [color={rgb, 255:red, 0; green, 0; blue, 0 }  ][line width=0.75]    (10.93,-3.29) .. controls (6.95,-1.4) and (3.31,-0.3) .. (0,0) .. controls (3.31,0.3) and (6.95,1.4) .. (10.93,3.29)   ;
    \draw    (90,170) -- (205,170) -- (267.98,131.49) ;
    \draw [shift={(269.68,130.45)}, rotate = 148.56] [color={rgb, 255:red, 0; green, 0; blue, 0 }  ][line width=0.75]    (10.93,-3.29) .. controls (6.95,-1.4) and (3.31,-0.3) .. (0,0) .. controls (3.31,0.3) and (6.95,1.4) .. (10.93,3.29)   ;
    \draw    (90,30) -- (204.68,29.45) -- (268.3,68.95) ;
    \draw [shift={(270,70)}, rotate = 211.83] [color={rgb, 255:red, 0; green, 0; blue, 0 }  ][line width=0.75]    (10.93,-3.29) .. controls (6.95,-1.4) and (3.31,-0.3) .. (0,0) .. controls (3.31,0.3) and (6.95,1.4) .. (10.93,3.29)   ;
    \draw    (350,100) -- (370,100) -- (370,180) -- (550.19,179.92) -- (550.44,140.13) -- (578.19,140.59) ;
    \draw [shift={(580.19,140.63)}, rotate = 180.96] [color={rgb, 255:red, 0; green, 0; blue, 0 }  ][line width=0.75]    (10.93,-3.29) .. controls (6.95,-1.4) and (3.31,-0.3) .. (0,0) .. controls (3.31,0.3) and (6.95,1.4) .. (10.93,3.29)   ;
    \draw  [fill={rgb, 255:red, 198; green, 235; blue, 160 }  ,fill opacity=1 ] (130,56) .. controls (130,52.69) and (132.69,50) .. (136,50) -- (184,50) .. controls (187.31,50) and (190,52.69) .. (190,56) -- (190,74) .. controls (190,77.31) and (187.31,80) .. (184,80) -- (136,80) .. controls (132.69,80) and (130,77.31) .. (130,74) -- cycle ;
    \draw  [fill={rgb, 255:red, 198; green, 235; blue, 160 }  ,fill opacity=1 ] (130,126) .. controls (130,122.69) and (132.69,120) .. (136,120) -- (184,120) .. controls (187.31,120) and (190,122.69) .. (190,126) -- (190,144) .. controls (190,147.31) and (187.31,150) .. (184,150) -- (136,150) .. controls (132.69,150) and (130,147.31) .. (130,144) -- cycle ;
    \draw  [fill={rgb, 255:red, 163; green, 241; blue, 227 }  ,fill opacity=1 ] (200,90) -- (240,90) -- (240,110) -- (200,110) -- cycle ;
    \draw  [fill={rgb, 255:red, 118; green, 165; blue, 212 }  ,fill opacity=0.95 ][line width=0.75]  (460,138) -- (476.7,134.64) -- (476.7,154.78) -- (460,150.75) -- (443.3,154.78) -- (443.3,134.64) -- cycle ;
    \draw    (240,100) -- (268,100) ;
    \draw [shift={(270,100)}, rotate = 180] [color={rgb, 255:red, 0; green, 0; blue, 0 }  ][line width=0.75]    (10.93,-3.29) .. controls (6.95,-1.4) and (3.31,-0.3) .. (0,0) .. controls (3.31,0.3) and (6.95,1.4) .. (10.93,3.29)   ;
    \draw  [line width=0.75]  (390,124.95) -- (423.03,124.95) -- (423.03,158.17) -- (390,158.17) -- cycle ;
    \draw  [line width=0.75]  (497,124.95) -- (530.03,124.95) -- (530.03,158.17) -- (497,158.17) -- cycle ;
    \draw  [line width=0.75]  (400,14.95) -- (433.03,14.95) -- (433.03,48.17) -- (400,48.17) -- cycle ;
    \draw  [fill={rgb, 255:red, 131; green, 182; blue, 79 }  ,fill opacity=1 ] (490,65) -- (530,50) -- (530,110) -- (490,95) -- (450,110) -- (450,50) -- cycle ;
    \draw    (423,145) -- (441,145) ;
    \draw [shift={(443,145)}, rotate = 180] [color={rgb, 255:red, 0; green, 0; blue, 0 }  ][line width=0.75]    (10.93,-3.29) .. controls (6.95,-1.4) and (3.31,-0.3) .. (0,0) .. controls (3.31,0.3) and (6.95,1.4) .. (10.93,3.29)   ;
    \draw    (477,145) -- (495,145) ;
    \draw [shift={(497,145)}, rotate = 180] [color={rgb, 255:red, 0; green, 0; blue, 0 }  ][line width=0.75]    (10.93,-3.29) .. controls (6.95,-1.4) and (3.31,-0.3) .. (0,0) .. controls (3.31,0.3) and (6.95,1.4) .. (10.93,3.29)   ;
    \draw    (512,125) -- (512,115) -- (490,115) -- (490,97) ;
    \draw [shift={(490,95)}, rotate = 90] [color={rgb, 255:red, 0; green, 0; blue, 0 }  ][line width=0.75]    (10.93,-3.29) .. controls (6.95,-1.4) and (3.31,-0.3) .. (0,0) .. controls (3.31,0.3) and (6.95,1.4) .. (10.93,3.29)   ;
    \draw    (433,35) -- (490,35) -- (490,63) ;
    \draw [shift={(490,65)}, rotate = 270] [color={rgb, 255:red, 0; green, 0; blue, 0 }  ][line width=0.75]    (10.93,-3.29) .. controls (6.95,-1.4) and (3.31,-0.3) .. (0,0) .. controls (3.31,0.3) and (6.95,1.4) .. (10.93,3.29)   ;
    \draw  [fill={rgb, 255:red, 163; green, 241; blue, 227 }  ,fill opacity=1 ] (390,70) -- (430,70) -- (430,90) -- (390,90) -- cycle ;
    \draw    (430,80) -- (448,80) ;
    \draw [shift={(450,80)}, rotate = 180] [color={rgb, 255:red, 0; green, 0; blue, 0 }  ][line width=0.75]    (10.93,-3.29) .. controls (6.95,-1.4) and (3.31,-0.3) .. (0,0) .. controls (3.31,0.3) and (6.95,1.4) .. (10.93,3.29)   ;
    \draw  [dash pattern={on 0.84pt off 2.51pt}]  (350,70) -- (388.6,11.9) ;
    \draw  [dash pattern={on 0.84pt off 2.51pt}]  (350,130) -- (390.2,169.5) ;
    \draw (55,45) node  {\includegraphics[width=52.5pt,height=52.5pt]{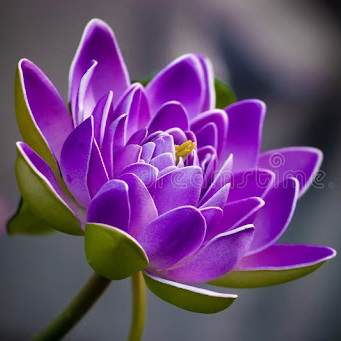}};
    \draw (55,139.67) node  {\includegraphics[width=52.5pt,height=52.5pt]{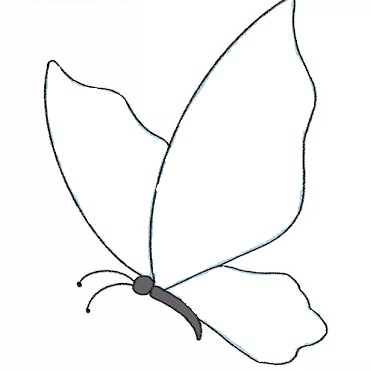}};
    \draw (416.51,31.56) node  {\includegraphics[width=24.77pt,height=24.91pt]{sec/Images/Flower.jpg}};
    \draw (406.51,141.56) node  {\includegraphics[width=24.77pt,height=24.91pt]{sec/Images/ButterSketch.jpg}};
    \draw (513.51,141.56) node  {\includegraphics[width=24.77pt,height=24.91pt]{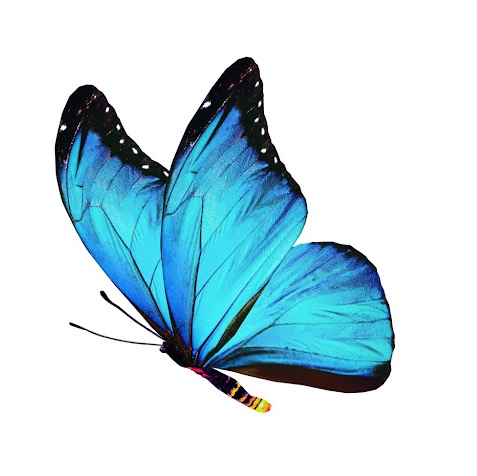}};
    \draw (629.5,125.01) node  {\includegraphics[width=74.25pt,height=67.51pt]{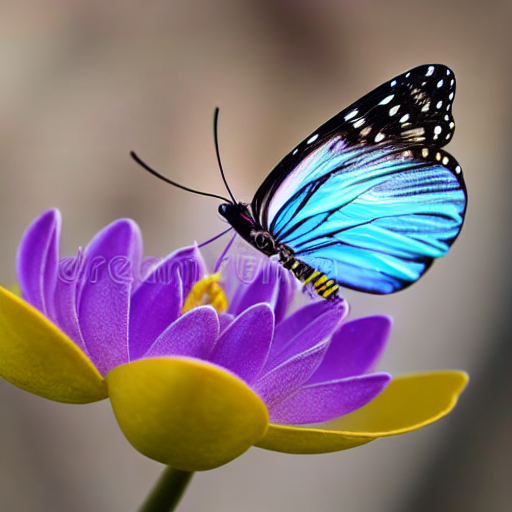}};
    
    \draw (137,60) node [anchor=north west][inner sep=0.75pt]   [align=left] {{\tiny Image Classifier}};
    \draw (137,130) node [anchor=north west][inner sep=0.75pt]   [align=center] {{\tiny Sketch Classifier}};
    \draw (206,93) node [anchor=north west][inner sep=0.75pt]   [align=left] {{\tiny Relation}};
    \draw (203,99) node [anchor=north west][inner sep=0.75pt]   [align=left] {{\tiny Generation}};
    \draw (277,93) node [anchor=north west][inner sep=0.75pt]   [align=left] {DiffMorph};
    \draw (27,82) node [anchor=north west][inner sep=0.75pt]  [font=\scriptsize] [align=left] {Input Image};
    \draw (27,176) node [anchor=north west][inner sep=0.75pt]  [font=\scriptsize] [align=left] {Input Image};
    \draw (597,173) node [anchor=north west][inner sep=0.75pt]  [font=\scriptsize] [align=left] {Output Image};
    \draw (477,71) node [anchor=north west][inner sep=0.75pt]   [align=left] {{\scriptsize Model}};
    \draw (468,80) node [anchor=north west][inner sep=0.75pt]   [align=left] {{\scriptsize Fine-tuning}};
    \draw (398,73) node [anchor=north west][inner sep=0.75pt]   [align=left] {{\tiny Relation}};
    \draw (394,79) node [anchor=north west][inner sep=0.75pt]   [align=left] {{\tiny Embedding}};
    {\begin{minipage}[lt]{28.56pt}\setlength\topsep{0pt}
    \begin{center}
    {\tiny Relation}\\{\tiny Embedding}
    \end{center}
    
    \end{minipage}};
    \draw (399,155) node [anchor=north west][inner sep=0.75pt]   [align=left] {{\tiny ConditionFlow}};
    
    \end{tikzpicture}
\caption{\textbf{Overview of DiffMorph.} The system takes an image and a sketch as a input and determines their respective classes using CLIP classifier~\cite{radford2021learning}. We use our ConditionFlow model to convert the sketch into an image. Subsequently, we fine-tune the Stable Diffusion Model~\cite{Rombach_2022_CVPR} with the images to generate a combined image.}
\label{fig:overview}
\end{figure*}

\section{Approach}
This work aims to produce a morphed 2D image derived from a base image and sequentially update the generated image by adding sketches for conditioning. This allows greater artistic control through sketch-based interaction over the generation process.

\subsection{Sketch-to-Image Generation}
\begin{figure*}
    \begin{center}
    \begin{tabular}{c c}
         \scalebox{0.35}{
            \begin{tikzpicture}[x=0.75pt,y=0.75pt,yscale=-1,xscale=1]
            
                    \draw  [fill={rgb, 255:red, 80; green, 227; blue, 194 }  ,fill opacity=0.35 ] (50,10) -- (50,299.43) -- (10,299.43) -- (10,10) -- cycle ;
                    
                    \draw  [fill={rgb, 255:red, 80; green, 227; blue, 194 }  ,fill opacity=0.35 ] (130,10) -- (130,280) -- (90,280) -- (90,10) -- cycle ;
                    
                    \draw  [fill={rgb, 255:red, 80; green, 227; blue, 194 }  ,fill opacity=0.35 ] (210,10) -- (210,260) -- (170,260) -- (170,10) -- cycle ;
                    
                    \draw  [fill={rgb, 255:red, 80; green, 227; blue, 194 }  ,fill opacity=0.35 ] (290,10) -- (290,240) -- (250,240) -- (250,10) -- cycle ;
                    
                    \draw  [fill={rgb, 255:red, 80; green, 227; blue, 194 }  ,fill opacity=0.35 ] (370,10) -- (370,220) -- (330,220) -- (330,10) -- cycle ;
                    
                    \draw  [fill={rgb, 255:red, 80; green, 227; blue, 194 }  ,fill opacity=0.35 ] (450,10) -- (450,240) -- (410,240) -- (410,10) -- cycle ;
                    
                    \draw  [fill={rgb, 255:red, 80; green, 227; blue, 194 }  ,fill opacity=0.35 ] (530,10) -- (530,260) -- (490,260) -- (490,10) -- cycle ;
                    
                    \draw  [fill={rgb, 255:red, 80; green, 227; blue, 194 }  ,fill opacity=0.35 ] (610,10) -- (610,280) -- (570,280) -- (570,10) -- cycle ;
                    
                    \draw  [fill={rgb, 255:red, 80; green, 227; blue, 194 }  ,fill opacity=0.35 ] (690,10.57) -- (690,300) -- (650,300) -- (650,10.57) -- cycle ;
                    
                    \draw  [fill={rgb, 255:red, 74; green, 144; blue, 226 }  ,fill opacity=0.35 ] (50,500) -- (50,800) -- (10,800) -- (10,500) -- cycle ;
                    
                    \draw  [fill={rgb, 255:red, 74; green, 144; blue, 226 }  ,fill opacity=0.35 ] (130,520) -- (130,800) -- (90,800) -- (90,520) -- cycle ;
                    
                    \draw  [fill={rgb, 255:red, 74; green, 144; blue, 226 }  ,fill opacity=0.35 ] (210,540) -- (210,800) -- (170,800) -- (170,540) -- cycle ;
                    
                    \draw  [fill={rgb, 255:red, 74; green, 144; blue, 226 }  ,fill opacity=0.35 ] (290,560.47) -- (290,800) -- (250,800) -- (250,560.47) -- cycle ;
                    
                    \draw  [fill={rgb, 255:red, 74; green, 144; blue, 226 }  ,fill opacity=0.35 ] (370,580) -- (370,800) -- (330,800) -- (330,580) -- cycle ;
                    
                    \draw    (50,110) -- (88,110) ;
                    \draw [shift={(90,110)}, rotate = 180] [color={rgb, 255:red, 0; green, 0; blue, 0 }  ][line width=0.75]    (10.93,-3.29) .. controls (6.95,-1.4) and (3.31,-0.3) .. (0,0) .. controls (3.31,0.3) and (6.95,1.4) .. (10.93,3.29)   ;
                    \draw    (130,110) -- (168,110) ;
                    \draw [shift={(170,110)}, rotate = 180] [color={rgb, 255:red, 0; green, 0; blue, 0 }  ][line width=0.75]    (10.93,-3.29) .. controls (6.95,-1.4) and (3.31,-0.3) .. (0,0) .. controls (3.31,0.3) and (6.95,1.4) .. (10.93,3.29)   ;
                    \draw    (210,110) -- (248,110) ;
                    \draw [shift={(250,110)}, rotate = 180] [color={rgb, 255:red, 0; green, 0; blue, 0 }  ][line width=0.75]    (10.93,-3.29) .. controls (6.95,-1.4) and (3.31,-0.3) .. (0,0) .. controls (3.31,0.3) and (6.95,1.4) .. (10.93,3.29)   ;
                    \draw    (290,110) -- (328,110) ;
                    \draw [shift={(330,110)}, rotate = 180] [color={rgb, 255:red, 0; green, 0; blue, 0 }  ][line width=0.75]    (10.93,-3.29) .. controls (6.95,-1.4) and (3.31,-0.3) .. (0,0) .. controls (3.31,0.3) and (6.95,1.4) .. (10.93,3.29)   ;
                    \draw    (370,110) -- (408,110) ;
                    \draw [shift={(410,110)}, rotate = 180] [color={rgb, 255:red, 0; green, 0; blue, 0 }  ][line width=0.75]    (10.93,-3.29) .. controls (6.95,-1.4) and (3.31,-0.3) .. (0,0) .. controls (3.31,0.3) and (6.95,1.4) .. (10.93,3.29)   ;
                    \draw    (450,110) -- (488,110) ;
                    \draw [shift={(490,110)}, rotate = 180] [color={rgb, 255:red, 0; green, 0; blue, 0 }  ][line width=0.75]    (10.93,-3.29) .. controls (6.95,-1.4) and (3.31,-0.3) .. (0,0) .. controls (3.31,0.3) and (6.95,1.4) .. (10.93,3.29)   ;
                    \draw    (530,110) -- (568,110) ;
                    \draw [shift={(570,110)}, rotate = 180] [color={rgb, 255:red, 0; green, 0; blue, 0 }  ][line width=0.75]    (10.93,-3.29) .. controls (6.95,-1.4) and (3.31,-0.3) .. (0,0) .. controls (3.31,0.3) and (6.95,1.4) .. (10.93,3.29)   ;
                    \draw    (610,110) -- (648,110) ;
                    \draw [shift={(650,110)}, rotate = 180] [color={rgb, 255:red, 0; green, 0; blue, 0 }  ][line width=0.75]    (10.93,-3.29) .. controls (6.95,-1.4) and (3.31,-0.3) .. (0,0) .. controls (3.31,0.3) and (6.95,1.4) .. (10.93,3.29)   ;
                    \draw    (50,690) -- (88,690) ;
                    \draw [shift={(90,690)}, rotate = 180] [color={rgb, 255:red, 0; green, 0; blue, 0 }  ][line width=0.75]    (10.93,-3.29) .. controls (6.95,-1.4) and (3.31,-0.3) .. (0,0) .. controls (3.31,0.3) and (6.95,1.4) .. (10.93,3.29)   ;
                    \draw    (130,690) -- (168,690) ;
                    \draw [shift={(170,690)}, rotate = 180] [color={rgb, 255:red, 0; green, 0; blue, 0 }  ][line width=0.75]    (10.93,-3.29) .. controls (6.95,-1.4) and (3.31,-0.3) .. (0,0) .. controls (3.31,0.3) and (6.95,1.4) .. (10.93,3.29)   ;
                    \draw    (210,690) -- (248,690) ;
                    \draw [shift={(250,690)}, rotate = 180] [color={rgb, 255:red, 0; green, 0; blue, 0 }  ][line width=0.75]    (10.93,-3.29) .. controls (6.95,-1.4) and (3.31,-0.3) .. (0,0) .. controls (3.31,0.3) and (6.95,1.4) .. (10.93,3.29)   ;
                    \draw    (290,690) -- (328,690) ;
                    \draw [shift={(330,690)}, rotate = 180] [color={rgb, 255:red, 0; green, 0; blue, 0 }  ][line width=0.75]    (10.93,-3.29) .. controls (6.95,-1.4) and (3.31,-0.3) .. (0,0) .. controls (3.31,0.3) and (6.95,1.4) .. (10.93,3.29)   ;
                    \draw   (300,519.5) -- (400,519.5) -- (400,549.5) -- (300,549.5) -- cycle ;
                    
                    \draw  [dash pattern={on 4.5pt off 4.5pt}]  (350,580) -- (350,552) ;
                    \draw [shift={(350,550)}, rotate = 90] [color={rgb, 255:red, 0; green, 0; blue, 0 }  ][line width=0.75]    (10.93,-3.29) .. controls (6.95,-1.4) and (3.31,-0.3) .. (0,0) .. controls (3.31,0.3) and (6.95,1.4) .. (10.93,3.29)   ;
                    \draw  [dash pattern={on 4.5pt off 4.5pt}]  (270,560) -- (270,512) ;
                    \draw [shift={(270,510)}, rotate = 90] [color={rgb, 255:red, 0; green, 0; blue, 0 }  ][line width=0.75]    (10.93,-3.29) .. controls (6.95,-1.4) and (3.31,-0.3) .. (0,0) .. controls (3.31,0.3) and (6.95,1.4) .. (10.93,3.29)   ;
                    \draw  [dash pattern={on 4.5pt off 4.5pt}]  (190,540) -- (190,462) ;
                    \draw [shift={(190,460)}, rotate = 90] [color={rgb, 255:red, 0; green, 0; blue, 0 }  ][line width=0.75]    (10.93,-3.29) .. controls (6.95,-1.4) and (3.31,-0.3) .. (0,0) .. controls (3.31,0.3) and (6.95,1.4) .. (10.93,3.29)   ;
                    \draw  [dash pattern={on 4.5pt off 4.5pt}]  (30,500) -- (30,462) ;
                    \draw [shift={(30,460)}, rotate = 90] [color={rgb, 255:red, 0; green, 0; blue, 0 }  ][line width=0.75]    (10.93,-3.29) .. controls (6.95,-1.4) and (3.31,-0.3) .. (0,0) .. controls (3.31,0.3) and (6.95,1.4) .. (10.93,3.29)   ;
                    \draw  [dash pattern={on 4.5pt off 4.5pt}]  (110,520) -- (110,502) ;
                    \draw [shift={(110,500)}, rotate = 90] [color={rgb, 255:red, 0; green, 0; blue, 0 }  ][line width=0.75]    (10.93,-3.29) .. controls (6.95,-1.4) and (3.31,-0.3) .. (0,0) .. controls (3.31,0.3) and (6.95,1.4) .. (10.93,3.29)   ;
                    \draw  [dash pattern={on 4.5pt off 4.5pt}]  (350,520) -- (350,442) ;
                    \draw [shift={(350,440)}, rotate = 90] [color={rgb, 255:red, 0; green, 0; blue, 0 }  ][line width=0.75]    (10.93,-3.29) .. controls (6.95,-1.4) and (3.31,-0.3) .. (0,0) .. controls (3.31,0.3) and (6.95,1.4) .. (10.93,3.29)   ;
                    \draw  [line width=1.5]  (335,425) .. controls (335,416.72) and (341.72,410) .. (350,410) .. controls (358.28,410) and (365,416.72) .. (365,425) .. controls (365,433.28) and (358.28,440) .. (350,440) .. controls (341.72,440) and (335,433.28) .. (335,425) -- cycle ; \draw  [line width=1.5]  (335,425) -- (365,425) ; \draw  [line width=1.5]  (350,410) -- (350,440) ;
                    \draw  [dash pattern={on 4.5pt off 4.5pt}]  (350,410) -- (350,222) ;
                    \draw [shift={(350,220)}, rotate = 90] [color={rgb, 255:red, 0; green, 0; blue, 0 }  ][line width=0.75]    (10.93,-3.29) .. controls (6.95,-1.4) and (3.31,-0.3) .. (0,0) .. controls (3.31,0.3) and (6.95,1.4) .. (10.93,3.29)   ;
                    \draw  [dash pattern={on 4.5pt off 4.5pt}]  (110,470) .. controls (110.25,441.19) and (99.98,400.73) .. (159.1,365.53) ;
                    \draw [shift={(160,365)}, rotate = 149.56] [color={rgb, 255:red, 0; green, 0; blue, 0 }  ][line width=0.75]    (10.93,-3.29) .. controls (6.95,-1.4) and (3.31,-0.3) .. (0,0) .. controls (3.31,0.3) and (6.95,1.4) .. (10.93,3.29)   ;
                    \draw  [dash pattern={on 4.5pt off 4.5pt}]  (191,365) .. controls (264.88,367.04) and (572.73,375.15) .. (589.77,283.39) ;
                    \draw [shift={(590,282)}, rotate = 98.64] [color={rgb, 255:red, 0; green, 0; blue, 0 }  ][line width=0.75]    (10.93,-3.29) .. controls (6.95,-1.4) and (3.31,-0.3) .. (0,0) .. controls (3.31,0.3) and (6.95,1.4) .. (10.93,3.29)   ;
                    \draw  [line width=1.5]  (160,365) .. controls (160,356.72) and (166.72,350) .. (175,350) .. controls (183.28,350) and (190,356.72) .. (190,365) .. controls (190,373.28) and (183.28,380) .. (175,380) .. controls (166.72,380) and (160,373.28) .. (160,365) -- cycle ; \draw  [line width=1.5]  (160,365) -- (190,365) ; \draw  [line width=1.5]  (175,350) -- (175,380) ;
                    \draw  [color={rgb, 255:red, 208; green, 2; blue, 27 }  ,draw opacity=1 ] [dash pattern={on 4.5pt off 4.5pt}]  (110,280) .. controls (111.24,330.54) and (170.67,295.54) .. (174.89,348.37) ;
                    \draw [shift={(175,350)}, rotate = 266.76] [color={rgb, 255:red, 208; green, 2; blue, 27 }  ][line width=0.75]    (10.93,-3.29) .. controls (6.95,-1.4) and (3.31,-0.3) .. (0,0) .. controls (3.31,0.3) and (6.95,1.4) .. (10.93,3.29)   ;
                    \draw  [line width=1.5]  (285,300) .. controls (285,291.72) and (291.72,285) .. (300,285) .. controls (308.28,285) and (315,291.72) .. (315,300) .. controls (315,308.28) and (308.28,315) .. (300,315) .. controls (291.72,315) and (285,308.28) .. (285,300) -- cycle ; \draw  [line width=1.5]  (285,300) -- (315,300) ; \draw  [line width=1.5]  (300,285) -- (300,315) ;
                    \draw  [color={rgb, 255:red, 208; green, 2; blue, 27 }  ,draw opacity=1 ] [dash pattern={on 4.5pt off 4.5pt}]  (270,240) .. controls (269.88,243.24) and (299.34,261.38) .. (300.01,283.31) ;
                    \draw [shift={(300,285)}, rotate = 272.21] [color={rgb, 255:red, 208; green, 2; blue, 27 }  ][line width=0.75]    (10.93,-3.29) .. controls (6.95,-1.4) and (3.31,-0.3) .. (0,0) .. controls (3.31,0.3) and (6.95,1.4) .. (10.93,3.29)   ;
                    \draw  [dash pattern={on 4.5pt off 4.5pt}]  (270,480) .. controls (270.62,442.21) and (299.54,364.4) .. (299.99,317.41) ;
                    \draw [shift={(300,316)}, rotate = 89.85] [color={rgb, 255:red, 0; green, 0; blue, 0 }  ][line width=0.75]    (10.93,-3.29) .. controls (6.95,-1.4) and (3.31,-0.3) .. (0,0) .. controls (3.31,0.3) and (6.95,1.4) .. (10.93,3.29)   ;
                    \draw  [dash pattern={on 4.5pt off 4.5pt}]  (316,300) .. controls (363.65,299.83) and (427.91,282.36) .. (429.95,241.87) ;
                    \draw [shift={(430,240)}, rotate = 90.17] [color={rgb, 255:red, 0; green, 0; blue, 0 }  ][line width=0.75]    (10.93,-3.29) .. controls (6.95,-1.4) and (3.31,-0.3) .. (0,0) .. controls (3.31,0.3) and (6.95,1.4) .. (10.93,3.29)   ;
                    \draw  [line width=1.5]  (235,335) .. controls (235,326.72) and (241.72,320) .. (250,320) .. controls (258.28,320) and (265,326.72) .. (265,335) .. controls (265,343.28) and (258.28,350) .. (250,350) .. controls (241.72,350) and (235,343.28) .. (235,335) -- cycle ; \draw  [line width=1.5]  (235,335) -- (265,335) ; \draw  [line width=1.5]  (250,320) -- (250,350) ;
                    \draw  [color={rgb, 255:red, 208; green, 2; blue, 27 }  ,draw opacity=1 ] [dash pattern={on 4.5pt off 4.5pt}]  (190,261) .. controls (189.88,264.24) and (247.86,301.4) .. (249.95,324.27) ;
                    \draw [shift={(250,326)}, rotate = 272.21] [color={rgb, 255:red, 208; green, 2; blue, 27 }  ][line width=0.75]    (10.93,-3.29) .. controls (6.95,-1.4) and (3.31,-0.3) .. (0,0) .. controls (3.31,0.3) and (6.95,1.4) .. (10.93,3.29)   ;
                    \draw  [dash pattern={on 4.5pt off 4.5pt}]  (190,429) .. controls (190.62,391.21) and (248.94,401.61) .. (249.99,356.39) ;
                    \draw [shift={(250,355)}, rotate = 89.85] [color={rgb, 255:red, 0; green, 0; blue, 0 }  ][line width=0.75]    (10.93,-3.29) .. controls (6.95,-1.4) and (3.31,-0.3) .. (0,0) .. controls (3.31,0.3) and (6.95,1.4) .. (10.93,3.29)   ;
                    \draw  [dash pattern={on 4.5pt off 4.5pt}]  (266,335) .. controls (318.9,332.58) and (509.25,343.13) .. (510,262.23) ;
                    \draw [shift={(510,261)}, rotate = 89.19] [color={rgb, 255:red, 0; green, 0; blue, 0 }  ][line width=0.75]    (10.93,-3.29) .. controls (6.95,-1.4) and (3.31,-0.3) .. (0,0) .. controls (3.31,0.3) and (6.95,1.4) .. (10.93,3.29)   ;
                    \draw  [line width=1.5]  (66,400) .. controls (66,391.72) and (72.72,385) .. (81,385) .. controls (89.28,385) and (96,391.72) .. (96,400) .. controls (96,408.28) and (89.28,415) .. (81,415) .. controls (72.72,415) and (66,408.28) .. (66,400) -- cycle ; \draw  [line width=1.5]  (66,400) -- (96,400) ; \draw  [line width=1.5]  (81,385) -- (81,415) ;
                    \draw  [dash pattern={on 4.5pt off 4.5pt}]  (30,300) .. controls (30.37,353.73) and (77.91,354.8) .. (79.93,383.23) ;
                    \draw [shift={(80,385)}, rotate = 269.76] [color={rgb, 255:red, 0; green, 0; blue, 0 }  ][line width=0.75]    (10.93,-3.29) .. controls (6.95,-1.4) and (3.31,-0.3) .. (0,0) .. controls (3.31,0.3) and (6.95,1.4) .. (10.93,3.29)   ;
                    \draw  [dash pattern={on 4.5pt off 4.5pt}]  (30,430) .. controls (31.86,394.19) and (19.62,400.69) .. (64.62,400.02) ;
                    \draw [shift={(66,400)}, rotate = 178.99] [color={rgb, 255:red, 0; green, 0; blue, 0 }  ][line width=0.75]    (10.93,-3.29) .. controls (6.95,-1.4) and (3.31,-0.3) .. (0,0) .. controls (3.31,0.3) and (6.95,1.4) .. (10.93,3.29)   ;
                    \draw  [dash pattern={on 4.5pt off 4.5pt}]  (97,400) .. controls (147.25,392.6) and (663.96,432.17) .. (669.95,301.98) ;
                    \draw [shift={(670,300)}, rotate = 90.36] [color={rgb, 255:red, 0; green, 0; blue, 0 }  ][line width=0.75]    (10.93,-3.29) .. controls (6.95,-1.4) and (3.31,-0.3) .. (0,0) .. controls (3.31,0.3) and (6.95,1.4) .. (10.93,3.29)   ;
                    \draw   (-100,130) -- (-30,130) -- (-30,170) -- (-100,170) -- cycle ;
                    
                    \draw    (-30,150) -- (8,150) ;
                    \draw [shift={(10,150)}, rotate = 180] [color={rgb, 255:red, 0; green, 0; blue, 0 }  ][line width=0.75]    (10.93,-3.29) .. controls (6.95,-1.4) and (3.31,-0.3) .. (0,0) .. controls (3.31,0.3) and (6.95,1.4) .. (10.93,3.29)   ;
                    \draw   (-150,640) -- (-70,640) -- (-70,670) -- (-150,670) -- cycle ;
                    
                    \draw  [line width=1.5]  (-50,650) .. controls (-50,641.72) and (-43.28,635) .. (-35,635) .. controls (-26.72,635) and (-20,641.72) .. (-20,650) .. controls (-20,658.28) and (-26.72,665) .. (-35,665) .. controls (-43.28,665) and (-50,658.28) .. (-50,650) -- cycle ; \draw  [line width=1.5]  (-50,650) -- (-20,650) ; \draw  [line width=1.5]  (-35,635) -- (-35,665) ;
                    \draw    (-70,650) -- (8,650) ;
                    \draw [shift={(10,650)}, rotate = 180] [color={rgb, 255:red, 0; green, 0; blue, 0 }  ][line width=0.75]    (10.93,-3.29) .. controls (6.95,-1.4) and (3.31,-0.3) .. (0,0) .. controls (3.31,0.3) and (6.95,1.4) .. (10.93,3.29)   ;
                    \draw    (-70,170) .. controls (-39.06,419.58) and (-33.85,580.85) .. (-34.96,634.41) ;
                    \draw [shift={(-35,636)}, rotate = 271.38] [color={rgb, 255:red, 0; green, 0; blue, 0 }  ][line width=0.75]    (10.93,-3.29) .. controls (6.95,-1.4) and (3.31,-0.3) .. (0,0) .. controls (3.31,0.3) and (6.95,1.4) .. (10.93,3.29)   ;
                    \draw   (220,480) -- (320,480) -- (320,510) -- (220,510) -- cycle ;
                    
                    \draw   (-20,430) -- (80,430) -- (80,460) -- (-20,460) -- cycle ;
                    
                    \draw   (60,470) -- (160,470) -- (160,500) -- (60,500) -- cycle ;
                    
                    \draw   (140,430) -- (240,430) -- (240,460) -- (140,460) -- cycle ;

                    \draw (38.53,42.71) node [anchor=north west][inner sep=0.75pt]  [rotate=-90] [align=left] {Stable Diffusion Encoder Block\_1};
                    \draw (118.53,33) node [anchor=north west][inner sep=0.75pt]  [rotate=-90] [align=left] {Stable Diffusion Encoder Block\_2};
                    \draw (198.53,23) node [anchor=north west][inner sep=0.75pt]  [rotate=-90] [align=left] {Stable Diffusion Encoder Block\_3};
                    \draw (278.53,13) node [anchor=north west][inner sep=0.75pt]  [rotate=-90] [align=left] {Stable Diffusion Encoder Block\_4};
                    \draw (358.53,17) node [anchor=north west][inner sep=0.75pt]  [rotate=-90] [align=left] {Stable Diffusion Middle Block};
                    \draw (438.53,13) node [anchor=north west][inner sep=0.75pt]  [rotate=-90] [align=left] {Stable Diffusion Decoder Block\_4};
                    \draw (518.53,23) node [anchor=north west][inner sep=0.75pt]  [rotate=-90] [align=left] {Stable Diffusion Decoder Block\_3};
                    \draw (598.53,33) node [anchor=north west][inner sep=0.75pt]  [rotate=-90] [align=left] {Stable Diffusion Decoder Block\_2};
                    \draw (678.53,43.29) node [anchor=north west][inner sep=0.75pt]  [rotate=-90] [align=left] {Stable Diffusion Decoder Block\_1};
                    \draw (358,587.5) node [anchor=north west][inner sep=0.75pt]  [rotate=-90] [align=left] {Condition Trainer Middle Block};
                    \draw (278,563.73) node [anchor=north west][inner sep=0.75pt]  [rotate=-90] [align=left] {Condition Trainer Encoder Block\_4};
                    \draw (198,553.5) node [anchor=north west][inner sep=0.75pt]  [rotate=-90] [align=left] {Condition Trainer Encoder Block\_3};
                    \draw (118,543.5) node [anchor=north west][inner sep=0.75pt]  [rotate=-90] [align=left] {Condition Trainer Encoder Block\_2};
                    \draw (38,533.5) node [anchor=north west][inner sep=0.75pt]  [rotate=-90] [align=left] {Condition Trainer Encoder Block\_1};
                    \draw (-65,150) node   [align=left] {\begin{minipage}[lt]{27.2pt}\setlength\topsep{0pt}
                    Input
                    \end{minipage}};
                    \draw (-142.5,646.5) node [anchor=north west][inner sep=0.75pt]   [align=left] {Condition};
                    \draw (315,526) node [anchor=north west][inner sep=0.75pt]   [align=left] {zero conv.};
                    \draw (235,486.5) node [anchor=north west][inner sep=0.75pt]   [align=left] {zero conv.};
                    \draw (-5,436.5) node [anchor=north west][inner sep=0.75pt]   [align=left] {zero conv.};
                    \draw (75,476.5) node [anchor=north west][inner sep=0.75pt]   [align=left] {zero conv.};
                    \draw (155,436.5) node [anchor=north west][inner sep=0.75pt]   [align=left] {zero conv.};

                \end{tikzpicture}
                
            }
             & 
            \scalebox{0.35}{   
    
                \begin{tikzpicture}[x=0.75pt,y=0.75pt,yscale=-1,xscale=1]
    
                    \draw  [fill={rgb, 255:red, 80; green, 227; blue, 194 }  ,fill opacity=0.35 ] (200,0) -- (200,289.43) -- (160,289.43) -- (160,0) -- cycle ;
                    
                    \draw  [fill={rgb, 255:red, 80; green, 227; blue, 194 }  ,fill opacity=0.35 ] (280,0) -- (280,270) -- (240,270) -- (240,0) -- cycle ;
                    
                    \draw  [fill={rgb, 255:red, 80; green, 227; blue, 194 }  ,fill opacity=0.35 ] (360,0) -- (360,250) -- (320,250) -- (320,0) -- cycle ;
                    
                    \draw  [fill={rgb, 255:red, 80; green, 227; blue, 194 }  ,fill opacity=0.35 ] (440,0) -- (440,230) -- (400,230) -- (400,0) -- cycle ;
                    
                    \draw  [fill={rgb, 255:red, 80; green, 227; blue, 194 }  ,fill opacity=0.35 ] (520,0) -- (520,210) -- (480,210) -- (480,0) -- cycle ;
                    
                    \draw  [fill={rgb, 255:red, 80; green, 227; blue, 194 }  ,fill opacity=0.35 ] (600,0) -- (600,230) -- (560,230) -- (560,0) -- cycle ;
                    
                    \draw  [fill={rgb, 255:red, 80; green, 227; blue, 194 }  ,fill opacity=0.35 ] (680,0) -- (680,250) -- (640,250) -- (640,0) -- cycle ;
                    
                    \draw  [fill={rgb, 255:red, 80; green, 227; blue, 194 }  ,fill opacity=0.35 ] (760,0) -- (760,270) -- (720,270) -- (720,0) -- cycle ;
                    
                    \draw  [fill={rgb, 255:red, 80; green, 227; blue, 194 }  ,fill opacity=0.35 ] (840,0.57) -- (840,290) -- (800,290) -- (800,0.57) -- cycle ;
                    
                    \draw  [fill={rgb, 255:red, 74; green, 144; blue, 226 }  ,fill opacity=0.35 ] (200,490) -- (200,790) -- (160,790) -- (160,490) -- cycle ;
                    
                    \draw  [fill={rgb, 255:red, 74; green, 144; blue, 226 }  ,fill opacity=0.35 ] (280,510) -- (280,790) -- (240,790) -- (240,510) -- cycle ;
                    
                    \draw  [fill={rgb, 255:red, 74; green, 144; blue, 226 }  ,fill opacity=0.35 ] (360,530) -- (360,790) -- (320,790) -- (320,530) -- cycle ;
                    
                    \draw  [fill={rgb, 255:red, 74; green, 144; blue, 226 }  ,fill opacity=0.35 ] (440,550.47) -- (440,790) -- (400,790) -- (400,550.47) -- cycle ;
                    
                    \draw  [fill={rgb, 255:red, 74; green, 144; blue, 226 }  ,fill opacity=0.35 ] (520,570) -- (520,790) -- (480,790) -- (480,570) -- cycle ;
                    
                    \draw    (200,100) -- (238,100) ;
                    \draw [shift={(240,100)}, rotate = 180] [color={rgb, 255:red, 0; green, 0; blue, 0 }  ][line width=0.75]    (10.93,-3.29) .. controls (6.95,-1.4) and (3.31,-0.3) .. (0,0) .. controls (3.31,0.3) and (6.95,1.4) .. (10.93,3.29)   ;
                    \draw    (280,100) -- (318,100) ;
                    \draw [shift={(320,100)}, rotate = 180] [color={rgb, 255:red, 0; green, 0; blue, 0 }  ][line width=0.75]    (10.93,-3.29) .. controls (6.95,-1.4) and (3.31,-0.3) .. (0,0) .. controls (3.31,0.3) and (6.95,1.4) .. (10.93,3.29)   ;
                    \draw    (360,100) -- (398,100) ;
                    \draw [shift={(400,100)}, rotate = 180] [color={rgb, 255:red, 0; green, 0; blue, 0 }  ][line width=0.75]    (10.93,-3.29) .. controls (6.95,-1.4) and (3.31,-0.3) .. (0,0) .. controls (3.31,0.3) and (6.95,1.4) .. (10.93,3.29)   ;
                    \draw    (440,100) -- (478,100) ;
                    \draw [shift={(480,100)}, rotate = 180] [color={rgb, 255:red, 0; green, 0; blue, 0 }  ][line width=0.75]    (10.93,-3.29) .. controls (6.95,-1.4) and (3.31,-0.3) .. (0,0) .. controls (3.31,0.3) and (6.95,1.4) .. (10.93,3.29)   ;
                    \draw    (520,100) -- (558,100) ;
                    \draw [shift={(560,100)}, rotate = 180] [color={rgb, 255:red, 0; green, 0; blue, 0 }  ][line width=0.75]    (10.93,-3.29) .. controls (6.95,-1.4) and (3.31,-0.3) .. (0,0) .. controls (3.31,0.3) and (6.95,1.4) .. (10.93,3.29)   ;
                    \draw    (600,100) -- (638,100) ;
                    \draw [shift={(640,100)}, rotate = 180] [color={rgb, 255:red, 0; green, 0; blue, 0 }  ][line width=0.75]    (10.93,-3.29) .. controls (6.95,-1.4) and (3.31,-0.3) .. (0,0) .. controls (3.31,0.3) and (6.95,1.4) .. (10.93,3.29)   ;
                    \draw    (680,100) -- (718,100) ;
                    \draw [shift={(720,100)}, rotate = 180] [color={rgb, 255:red, 0; green, 0; blue, 0 }  ][line width=0.75]    (10.93,-3.29) .. controls (6.95,-1.4) and (3.31,-0.3) .. (0,0) .. controls (3.31,0.3) and (6.95,1.4) .. (10.93,3.29)   ;
                    \draw    (760,100) -- (798,100) ;
                    \draw [shift={(800,100)}, rotate = 180] [color={rgb, 255:red, 0; green, 0; blue, 0 }  ][line width=0.75]    (10.93,-3.29) .. controls (6.95,-1.4) and (3.31,-0.3) .. (0,0) .. controls (3.31,0.3) and (6.95,1.4) .. (10.93,3.29)   ;
                    \draw    (200,680) -- (238,680) ;
                    \draw [shift={(240,680)}, rotate = 180] [color={rgb, 255:red, 0; green, 0; blue, 0 }  ][line width=0.75]    (10.93,-3.29) .. controls (6.95,-1.4) and (3.31,-0.3) .. (0,0) .. controls (3.31,0.3) and (6.95,1.4) .. (10.93,3.29)   ;
                    \draw    (280,680) -- (318,680) ;
                    \draw [shift={(320,680)}, rotate = 180] [color={rgb, 255:red, 0; green, 0; blue, 0 }  ][line width=0.75]    (10.93,-3.29) .. controls (6.95,-1.4) and (3.31,-0.3) .. (0,0) .. controls (3.31,0.3) and (6.95,1.4) .. (10.93,3.29)   ;
                    \draw    (360,680) -- (398,680) ;
                    \draw [shift={(400,680)}, rotate = 180] [color={rgb, 255:red, 0; green, 0; blue, 0 }  ][line width=0.75]    (10.93,-3.29) .. controls (6.95,-1.4) and (3.31,-0.3) .. (0,0) .. controls (3.31,0.3) and (6.95,1.4) .. (10.93,3.29)   ;
                    \draw    (440,680) -- (478,680) ;
                    \draw [shift={(480,680)}, rotate = 180] [color={rgb, 255:red, 0; green, 0; blue, 0 }  ][line width=0.75]    (10.93,-3.29) .. controls (6.95,-1.4) and (3.31,-0.3) .. (0,0) .. controls (3.31,0.3) and (6.95,1.4) .. (10.93,3.29)   ;
                    \draw   (450,510) -- (550,510) -- (550,540) -- (450,540) -- cycle ;
                    
                    \draw  [dash pattern={on 4.5pt off 4.5pt}]  (500,570) -- (500,542) ;
                    \draw [shift={(500,540)}, rotate = 90] [color={rgb, 255:red, 0; green, 0; blue, 0 }  ][line width=0.75]    (10.93,-3.29) .. controls (6.95,-1.4) and (3.31,-0.3) .. (0,0) .. controls (3.31,0.3) and (6.95,1.4) .. (10.93,3.29)   ;
                    \draw  [dash pattern={on 4.5pt off 4.5pt}]  (420,550) -- (420,502) ;
                    \draw [shift={(420,500)}, rotate = 90] [color={rgb, 255:red, 0; green, 0; blue, 0 }  ][line width=0.75]    (10.93,-3.29) .. controls (6.95,-1.4) and (3.31,-0.3) .. (0,0) .. controls (3.31,0.3) and (6.95,1.4) .. (10.93,3.29)   ;
                    \draw  [dash pattern={on 4.5pt off 4.5pt}]  (340,530) -- (340,452) ;
                    \draw [shift={(340,450)}, rotate = 90] [color={rgb, 255:red, 0; green, 0; blue, 0 }  ][line width=0.75]    (10.93,-3.29) .. controls (6.95,-1.4) and (3.31,-0.3) .. (0,0) .. controls (3.31,0.3) and (6.95,1.4) .. (10.93,3.29)   ;
                    \draw  [dash pattern={on 4.5pt off 4.5pt}]  (180,490) -- (180,452) ;
                    \draw [shift={(180,450)}, rotate = 90] [color={rgb, 255:red, 0; green, 0; blue, 0 }  ][line width=0.75]    (10.93,-3.29) .. controls (6.95,-1.4) and (3.31,-0.3) .. (0,0) .. controls (3.31,0.3) and (6.95,1.4) .. (10.93,3.29)   ;
                    \draw  [dash pattern={on 4.5pt off 4.5pt}]  (260,510) -- (260,492) ;
                    \draw [shift={(260,490)}, rotate = 90] [color={rgb, 255:red, 0; green, 0; blue, 0 }  ][line width=0.75]    (10.93,-3.29) .. controls (6.95,-1.4) and (3.31,-0.3) .. (0,0) .. controls (3.31,0.3) and (6.95,1.4) .. (10.93,3.29)   ;
                    \draw  [line width=1.5]  (485,261) .. controls (485,252.72) and (491.72,246) .. (500,246) .. controls (508.28,246) and (515,252.72) .. (515,261) .. controls (515,269.28) and (508.28,276) .. (500,276) .. controls (491.72,276) and (485,269.28) .. (485,261) -- cycle ; \draw  [line width=1.5]  (485,261) -- (515,261) ; \draw  [line width=1.5]  (500,246) -- (500,276) ;
                    \draw  [dash pattern={on 4.5pt off 4.5pt}]  (260,460) .. controls (260.17,380.33) and (260.83,379.67) .. (523,380) ;
                    \draw [shift={(523,380)}, rotate = 180.07] [color={rgb, 255:red, 0; green, 0; blue, 0 }  ][line width=0.75]    (10.93,-3.29) .. controls (6.95,-1.4) and (3.31,-0.3) .. (0,0) .. controls (3.31,0.3) and (6.95,1.4) .. (10.93,3.29)   ;
                    \draw  [line width=1.5]  (525,380) .. controls (525,371.72) and (531.72,365) .. (540,365) .. controls (548.28,365) and (555,371.72) .. (555,380) .. controls (555,388.28) and (548.28,395) .. (540,395) .. controls (531.72,395) and (525,388.28) .. (525,380) -- cycle ; \draw  [line width=1.5]  (525,380) -- (555,380) ; \draw  [line width=1.5]  (540,365) -- (540,395) ;
                    \draw  [line width=1.5]  (566,340) .. controls (566,331.72) and (572.72,325) .. (581,325) .. controls (589.28,325) and (596,331.72) .. (596,340) .. controls (596,348.28) and (589.28,355) .. (581,355) .. controls (572.72,355) and (566,348.28) .. (566,340) -- cycle ; \draw  [line width=1.5]  (566,340) -- (596,340) ; \draw  [line width=1.5]  (581,325) -- (581,355) ;
                    \draw  [dash pattern={on 4.5pt off 4.5pt}]  (420,470) .. controls (420.17,300.52) and (419.84,300.99) .. (523.43,300.01) ;
                    \draw [shift={(525,300)}, rotate = 179.46] [color={rgb, 255:red, 0; green, 0; blue, 0 }  ][line width=0.75]    (10.93,-3.29) .. controls (6.95,-1.4) and (3.31,-0.3) .. (0,0) .. controls (3.31,0.3) and (6.95,1.4) .. (10.93,3.29)   ;
                    \draw  [line width=1.5]  (525,300) .. controls (525,291.72) and (531.72,285) .. (540,285) .. controls (548.28,285) and (555,291.72) .. (555,300) .. controls (555,308.28) and (548.28,315) .. (540,315) .. controls (531.72,315) and (525,308.28) .. (525,300) -- cycle ; \draw  [line width=1.5]  (525,300) -- (555,300) ; \draw  [line width=1.5]  (540,285) -- (540,315) ;
                    \draw  [dash pattern={on 4.5pt off 4.5pt}]  (342,419) .. controls (338.83,342.33) and (338.83,341) .. (566,340) ;
                    \draw [shift={(566,340)}, rotate = 179.75] [color={rgb, 255:red, 0; green, 0; blue, 0 }  ][line width=0.75]    (10.93,-3.29) .. controls (6.95,-1.4) and (3.31,-0.3) .. (0,0) .. controls (3.31,0.3) and (6.95,1.4) .. (10.93,3.29)   ;
                    \draw  [line width=1.5]  (216,400) .. controls (216,391.72) and (222.72,385) .. (231,385) .. controls (239.28,385) and (246,391.72) .. (246,400) .. controls (246,408.28) and (239.28,415) .. (231,415) .. controls (222.72,415) and (216,408.28) .. (216,400) -- cycle ; \draw  [line width=1.5]  (216,400) -- (246,400) ; \draw  [line width=1.5]  (231,385) -- (231,415) ;
                    \draw  [dash pattern={on 4.5pt off 4.5pt}]  (180,290) .. controls (180.37,343.73) and (227.91,353.45) .. (229.93,382.21) ;
                    \draw [shift={(230,384)}, rotate = 269.76] [color={rgb, 255:red, 0; green, 0; blue, 0 }  ][line width=0.75]    (10.93,-3.29) .. controls (6.95,-1.4) and (3.31,-0.3) .. (0,0) .. controls (3.31,0.3) and (6.95,1.4) .. (10.93,3.29)   ;
                    \draw  [dash pattern={on 4.5pt off 4.5pt}]  (180,420) .. controls (180.82,400.73) and (182.75,400.34) .. (214.05,400.02) ;
                    \draw [shift={(216,400)}, rotate = 179.42] [color={rgb, 255:red, 0; green, 0; blue, 0 }  ][line width=0.75]    (10.93,-3.29) .. controls (6.95,-1.4) and (3.31,-0.3) .. (0,0) .. controls (3.31,0.3) and (6.95,1.4) .. (10.93,3.29)   ;
                    \draw  [dash pattern={on 4.5pt off 4.5pt}]  (247,400) .. controls (314.83,399) and (359.17,410.33) .. (590,410) .. controls (819.68,409.67) and (820.83,409.67) .. (820.01,291.79) ;
                    \draw [shift={(820,290)}, rotate = 89.6] [color={rgb, 255:red, 0; green, 0; blue, 0 }  ][line width=0.75]    (10.93,-3.29) .. controls (6.95,-1.4) and (3.31,-0.3) .. (0,0) .. controls (3.31,0.3) and (6.95,1.4) .. (10.93,3.29)   ;
                    \draw   (50,80) -- (120,80) -- (120,120) -- (50,120) -- cycle ;
                    
                    \draw    (120,100) -- (158,100) ;
                    \draw [shift={(160,100)}, rotate = 180] [color={rgb, 255:red, 0; green, 0; blue, 0 }  ][line width=0.75]    (10.93,-3.29) .. controls (6.95,-1.4) and (3.31,-0.3) .. (0,0) .. controls (3.31,0.3) and (6.95,1.4) .. (10.93,3.29)   ;
                    \draw   (0,664) -- (80,664) -- (80,694) -- (0,694) -- cycle ;
                    
                    \draw  [line width=1.5]  (100,680) .. controls (100,671.72) and (106.72,665) .. (115,665) .. controls (123.28,665) and (130,671.72) .. (130,680) .. controls (130,688.28) and (123.28,695) .. (115,695) .. controls (106.72,695) and (100,688.28) .. (100,680) -- cycle ; \draw  [line width=1.5]  (100,680) -- (130,680) ; \draw  [line width=1.5]  (115,665) -- (115,695) ;
                    \draw    (80,680) -- (158,680) ;
                    \draw [shift={(160,680)}, rotate = 180] [color={rgb, 255:red, 0; green, 0; blue, 0 }  ][line width=0.75]    (10.93,-3.29) .. controls (6.95,-1.4) and (3.31,-0.3) .. (0,0) .. controls (3.31,0.3) and (6.95,1.4) .. (10.93,3.29)   ;
                    \draw    (85,122) .. controls (115.94,371.58) and (117.23,608.32) .. (116.04,663.39) ;
                    \draw [shift={(116,665)}, rotate = 271.38] [color={rgb, 255:red, 0; green, 0; blue, 0 }  ][line width=0.75]    (10.93,-3.29) .. controls (6.95,-1.4) and (3.31,-0.3) .. (0,0) .. controls (3.31,0.3) and (6.95,1.4) .. (10.93,3.29)   ;
                    \draw   (370,470) -- (470,470) -- (470,500) -- (370,500) -- cycle ;
                    
                    \draw   (130,420) -- (230,420) -- (230,450) -- (130,450) -- cycle ;
                    
                    \draw   (210,460) -- (310,460) -- (310,490) -- (210,490) -- cycle ;
                    
                    \draw   (290,420) -- (390,420) -- (390,450) -- (290,450) -- cycle ;
                    
                    \draw  [dash pattern={on 4.5pt off 4.5pt}]  (500,506) -- (500,278) ;
                    \draw [shift={(500,276)}, rotate = 90] [color={rgb, 255:red, 0; green, 0; blue, 0 }  ][line width=0.75]    (10.93,-3.29) .. controls (6.95,-1.4) and (3.31,-0.3) .. (0,0) .. controls (3.31,0.3) and (6.95,1.4) .. (10.93,3.29)   ;
                    \draw  [dash pattern={on 4.5pt off 4.5pt}]  (500,246) -- (500,212) ;
                    \draw [shift={(500,210)}, rotate = 90] [color={rgb, 255:red, 0; green, 0; blue, 0 }  ][line width=0.75]    (10.93,-3.29) .. controls (6.95,-1.4) and (3.31,-0.3) .. (0,0) .. controls (3.31,0.3) and (6.95,1.4) .. (10.93,3.29)   ;
                    \draw  [dash pattern={on 4.5pt off 4.5pt}]  (557,300) .. controls (582.04,300.33) and (580.85,300.99) .. (580.01,231.06) ;
                    \draw [shift={(580,230)}, rotate = 89.33] [color={rgb, 255:red, 0; green, 0; blue, 0 }  ][line width=0.75]    (10.93,-3.29) .. controls (6.95,-1.4) and (3.31,-0.3) .. (0,0) .. controls (3.31,0.3) and (6.95,1.4) .. (10.93,3.29)   ;
                    \draw  [dash pattern={on 4.5pt off 4.5pt}]  (597,340) .. controls (653.6,340.33) and (660.69,308.97) .. (660.02,251.74) ;
                    \draw [shift={(660,250)}, rotate = 89.18] [color={rgb, 255:red, 0; green, 0; blue, 0 }  ][line width=0.75]    (10.93,-3.29) .. controls (6.95,-1.4) and (3.31,-0.3) .. (0,0) .. controls (3.31,0.3) and (6.95,1.4) .. (10.93,3.29)   ;
                    \draw  [dash pattern={on 4.5pt off 4.5pt}]  (551,380) .. controls (739.22,380.33) and (740.82,380.99) .. (740.01,271.66) ;
                    \draw [shift={(740,270)}, rotate = 89.57] [color={rgb, 255:red, 0; green, 0; blue, 0 }  ][line width=0.75]    (10.93,-3.29) .. controls (6.95,-1.4) and (3.31,-0.3) .. (0,0) .. controls (3.31,0.3) and (6.95,1.4) .. (10.93,3.29)   ;
                    
                    \draw (188.53,32.71) node [anchor=north west][inner sep=0.75pt]  [rotate=-90] [align=left] {Stable Diffusion Encoder Block\_1};
                    \draw (268.53,23) node [anchor=north west][inner sep=0.75pt]  [rotate=-90] [align=left] {Stable Diffusion Encoder Block\_2};
                    \draw (348.53,13) node [anchor=north west][inner sep=0.75pt]  [rotate=-90] [align=left] {Stable Diffusion Encoder Block\_3};
                    \draw (428.53,3) node [anchor=north west][inner sep=0.75pt]  [rotate=-90] [align=left] {Stable Diffusion Encoder Block\_4};
                    \draw (508.53,7) node [anchor=north west][inner sep=0.75pt]  [rotate=-90] [align=left] {Stable Diffusion Middle Block};
                    \draw (588.53,3) node [anchor=north west][inner sep=0.75pt]  [rotate=-90] [align=left] {Stable Diffusion Decoder Block\_4};
                    \draw (668.53,13) node [anchor=north west][inner sep=0.75pt]  [rotate=-90] [align=left] {Stable Diffusion Decoder Block\_3};
                    \draw (748.53,23) node [anchor=north west][inner sep=0.75pt]  [rotate=-90] [align=left] {Stable Diffusion Decoder Block\_2};
                    \draw (828.53,33.29) node [anchor=north west][inner sep=0.75pt]  [rotate=-90] [align=left] {Stable Diffusion Decoder Block\_1};
                    \draw (508,577.5) node [anchor=north west][inner sep=0.75pt]  [rotate=-90] [align=left] {Condition Trainer Middle Block};
                    \draw (428,553.73) node [anchor=north west][inner sep=0.75pt]  [rotate=-90] [align=left] {Condition Trainer Encoder Block\_4};
                    \draw (348,543.5) node [anchor=north west][inner sep=0.75pt]  [rotate=-90] [align=left] {Condition Trainer Encoder Block\_3};
                    \draw (268,533.5) node [anchor=north west][inner sep=0.75pt]  [rotate=-90] [align=left] {Condition Trainer Encoder Block\_2};
                    \draw (188,523.5) node [anchor=north west][inner sep=0.75pt]  [rotate=-90] [align=left] {Condition Trainer Encoder Block\_1};
                    \draw (85,100) node   [align=left] {\begin{minipage}[lt]{27.2pt}\setlength\topsep{0pt}
                    Input
                    \end{minipage}};
                    \draw (7.5,670.5) node [anchor=north west][inner sep=0.75pt]   [align=left] {Condition};
                    \draw (385,476.5) node [anchor=north west][inner sep=0.75pt]   [align=left] {zero conv.};
                    \draw (145,426.5) node [anchor=north west][inner sep=0.75pt]   [align=left] {zero conv.};
                    \draw (225,466.5) node [anchor=north west][inner sep=0.75pt]   [align=left] {zero conv.};
                    \draw (305,426.5) node [anchor=north west][inner sep=0.75pt]   [align=left] {zero conv.};
                    \draw (465,516) node [anchor=north west][inner sep=0.75pt]   [align=left] {zero conv.};

                \end{tikzpicture}
    
            }\\
            (a) ControlNet Architecture & (b) Modified Architecture\\
            \multicolumn{2}{c}{}
        \end{tabular}
    \end{center}
    \caption{Comparison of ControlNet vs ConditionFlow Architecture. The connections marked in red in 1.(a) are removed in 1.(b) as our suggested update.}
    \label{fig:architecture}
\end{figure*}
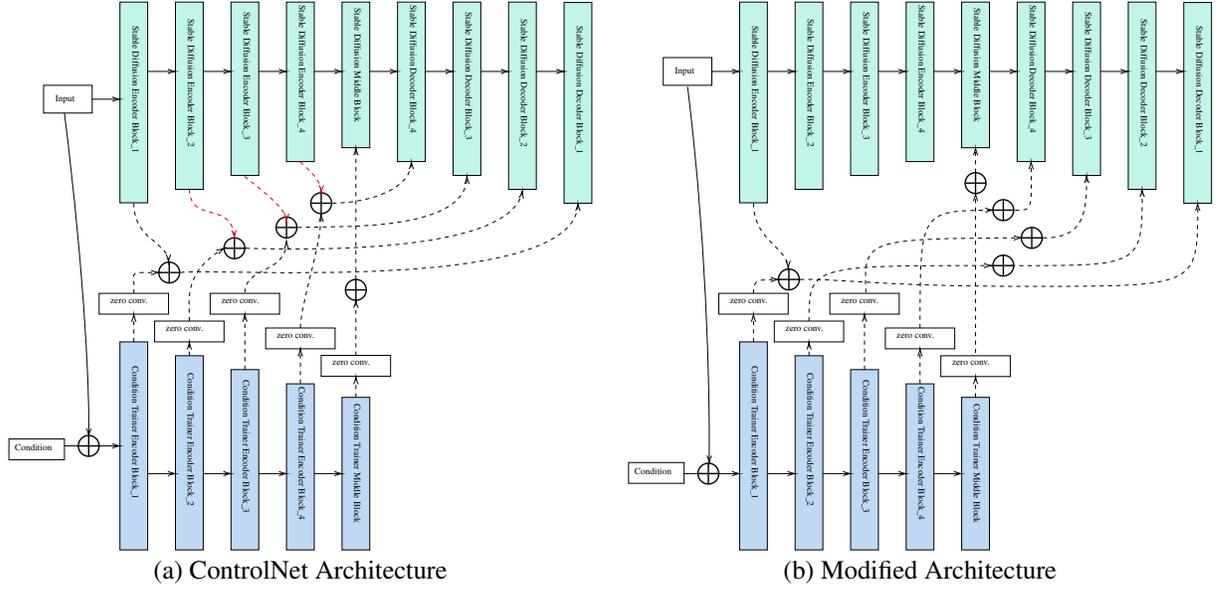

ControlNet~\cite{zhang2023adding} by Zhang et al. represents a new approach for image generation employing the diffusion technique. It generates a new image based on an external constraint, such as edges, depth, $\alpha$-pose, or sketches. To realize this, the condition is routed through an encoder block. The output from each encoding layer is seamlessly integrated with the corresponding layer of a denoising U-Net model, as illustrated in Figure~\ref{fig:architecture}(a).

Within the conventional U-Net architecture, skip connections are pivotal in the encoder-decoder framework. These connections function by combining the outputs of the encoder segment with the relevant decoder segment, thereby facilitating the retrieval of spatial information potentially compromised during the downsampling phase. This spatial data delineates the relative positioning of pixel pairs, typically characterized by parameters such as distance $d$, and orientation $\theta$. The described parameters clarify the positioning of a pixel relative to another, as detailed in \cite{BANKMAN2009261}. Notably, even amidst this downsampling, the encoder-decoder segment preserves vital feature data, encompassing elements like texture, color, and edges, as cited in \cite{cciccek20163d}.

Within the framework of ControlNet~\cite{zhang2023adding}, the researchers concatenate both the encoded condition and the input with the decoding segment. Such a fusion ensures that spatial attributes from both the image and the condition are incorporated. When the input is conditioned upon an $\alpha$-pose or a depth map, the model also requires the input data. Conversely, our investigations indicate that spatial information derived from images might not always be cardinal for conditions like sketches.  In our model, referred to as \textit{ConditionFlow}, depicted in Figure~\ref{fig:architecture}(b), we have excised the skip connections embodying the spatial features of the input image. We find that images produced through our approach parallel or even surpass the quality of those generated by the baseline ControlNet model. 
We tested a range of settings where combinations of skip connections are removed. However, our chosen approach gives the most optimal result. Please see our comparative study for the settings in section~\ref{sec: condflow}. 

We assume the regular denoising U-Net block as $\mathcal{F}$ with $\Theta$ parameters. With a given noisy image $x\in \mathbb{R}^{h\times w\times c}$ with ${h, w, c}$ being height, weight and number channels and $y$ being the segmented noise, the network block becomes

\begin{equation}\label{eq:1}
y=\mathcal{F}( x,\Theta ).
\end{equation}

As there are concatenations of features from the encoder to decoder block in U-Net, we can rewrite  Eqn.~\ref{eq:1} as follows
\begin{equation}\label{eq:2}
y=\mathcal{G}( x,\Theta ) -x^\prime,
\end{equation}
where $\mathcal{G}$ is the neural network block without the skip connections. As skip connections are encoded input~\cite{he2016deep} that are typically added to the decoder block, we can represent these as $x^\prime$ as shown in Eqn.~\ref{eq:2}. 

In ControlNet~\cite{zhang2023adding}, the parameters in $\Theta$ are locked and then cloned into a trainable copy $\Theta_c$. The copied $\Theta_c$ is trained in another neural network where blocks are connected by a unique type of convolution layer called “zero convolution”. This “zero convolution” is introduced by the authors which is a $1\times1$ convolution layer with both weight and bias initialized with zeros denoted by $\mathbb{Z}$ with parameters $\{\Theta_{z1}, \Theta_{z2}\}$ for two different instances. Then the structure becomes
\begin{equation}\label{eq:3}
    y_{CN} =\mathcal{F}( x,\Theta ) +\mathcal{Z}(\mathcal{F}( c+\mathcal{Z}( c;\ \Theta _{z1}) ;\Theta _{c}) ;\Theta _{z2}).
\end{equation}

Therefore, without the spatial features from the images, the updated structure of Eqn.~\ref{eq:3} becomes
\begin{equation}\label{eq:4}
    y_{CF} =\mathcal{G}( x,\Theta ) +\mathcal{Z}(\mathcal{F}( c+\mathcal{Z}( c;\ \Theta_{z1}) ;\Theta_{c}) ;\Theta_{z2}).
\end{equation}
We use the same encoding function for the sketch conditions $c$, so the zero convolution side remains the same. However, the final decoder block differs and is denoted as $\mathcal{G}$ previously. Therefore, Eqn.~\ref{eq:4} is rewritten from Eqn.~\ref{eq:3} without the spatial features from the input noisy images

\begin{equation}
    \therefore \ y_{CF} \simeq y_{CN} -x^\prime.\nonumber
\end{equation}

Since we have locked the actual parameters and trained the parameters for the condition, the loss function remains the same as ControlNet~\cite{zhang2023adding}
\begin{equation}
    \mathcal{L} =\mathbb{E}_{z_0,t,\ c_{f} ,c_{t} ,\ \epsilon \sim \mathcal{N}( 0,1)}\left[ \| \epsilon -\epsilon _{\theta }( z_{t} ,t,c_{f} ,c_{t}) \| _{2}^{2}\right].\nonumber
\end{equation}
where $z_0$ is the input image, which becomes $z_t$ by adding noise in the image diffusion algorithm, where $t$ represents the number of time steps of noise addition. $c_t$ and $c_t$ are text prompt and sketch conditions, respectively. $\epsilon _{\theta }$ is the network to predict noise of $z_t$.

\subsection{Conditioning Image with Multiple Concepts}
Historically, numerous methodologies have been proposed for model fine-tuning, primarily focusing on refining the diffusion model to generate a specific object related to a designated class or concept.
The method generally originated from either a single image~\cite{kawar2023imagic, balaji2022ediffi} or requires multiple images~\cite{ruiz2023dreambooth, brooks2023instructpix2pix, kumari2023multi}. A recurrent challenge with these techniques is their tendency to over-fit the model~\cite{tewel2023key}. A consequence of over-fitting is that images subsequently generated by the adjusted model often mirror the initial object image closely, resulting in new conditions introduced later exerting minimal impact on the generated output. Rather than depending on an array of images, our method requires merely one image per concept and handles multiple classes. Post fine-tuning, the enhanced model is capable of generating images that accommodate all the classes.

\begin{figure}
    \centering
    \begin{tabular}{c}
        \scalebox{1.2}{
            \tikzset{every picture/.style={line width=0.75pt}} 
            \begin{tikzpicture}[x=0.75pt,y=0.75pt,yscale=-1,xscale=1]
            \draw  [fill={rgb, 255:red, 238; green, 228; blue, 137 }  ,fill opacity=1 ] (28.14,40) -- (43.83,40) -- (43.83,87.35) -- (28.14,87.35) -- cycle ;
            \draw  [fill={rgb, 255:red, 241; green, 191; blue, 115 }  ,fill opacity=1 ] (67.38,52.63) .. controls (67.38,45.65) and (73.03,40) .. (80,40) -- (141.07,40) .. controls (148.04,40) and (153.7,45.65) .. (153.7,52.63) -- (153.7,90.51) .. controls (153.7,97.48) and (148.04,103.13) .. (141.07,103.13) -- (80,103.13) .. controls (73.03,103.13) and (67.38,97.48) .. (67.38,90.51) -- cycle ;
            \draw    (43.83,63.67) -- (65.38,63.67) ;
            \draw [shift={(67.38,63.67)}, rotate = 180] [color={rgb, 255:red, 0; green, 0; blue, 0 }  ][line width=0.75]    (10.93,-3.29) .. controls (6.95,-1.4) and (3.31,-0.3) .. (0,0) .. controls (3.31,0.3) and (6.95,1.4) .. (10.93,3.29)   ;
            \draw    (47.76,99.19) -- (65.97,80.88) ;
            \draw [shift={(67.38,79.46)}, rotate = 134.84] [color={rgb, 255:red, 0; green, 0; blue, 0 }  ][line width=0.75]    (10.93,-3.29) .. controls (6.95,-1.4) and (3.31,-0.3) .. (0,0) .. controls (3.31,0.3) and (6.95,1.4) .. (10.93,3.29)   ;
            \draw (208.63,71.57) node  {\includegraphics[width=47.08pt,height=47.35pt]{sec/Images/Flower.jpg}};
            \draw    (153.7,64.07) -- (175.63,64.07) ;
            \draw [shift={(177.63,64.07)}, rotate = 180] [color={rgb, 255:red, 0; green, 0; blue, 0 }  ][line width=0.75]    (10.93,-3.29) .. controls (6.95,-1.4) and (3.31,-0.3) .. (0,0) .. controls (3.31,0.3) and (6.95,1.4) .. (10.93,3.29)   ;
            \draw    (177.24,79.46) -- (156.09,80.18) ;
            \draw [shift={(154.09,80.25)}, rotate = 358.05] [color={rgb, 255:red, 0; green, 0; blue, 0 }  ][line width=0.75]    (10.93,-3.29) .. controls (6.95,-1.4) and (3.31,-0.3) .. (0,0) .. controls (3.31,0.3) and (6.95,1.4) .. (10.93,3.29)   ;
            \draw (338.11,71.57) node  {\includegraphics[width=47.08pt,height=47.35pt]{sec/Images/Flower.jpg}};
            \draw  [dash pattern={on 7.5pt off 2.25pt}]  (306.72,71.57) -- (242.02,71.57) ;
            \draw [shift={(240.02,71.57)}, rotate = 360] [color={rgb, 255:red, 0; green, 0; blue, 0 }  ][line width=0.75]    (10.93,-3.29) .. controls (6.95,-1.4) and (3.31,-0.3) .. (0,0) .. controls (3.31,0.3) and (6.95,1.4) .. (10.93,3.29)   ;
            \draw  [dash pattern={on 7.5pt off 3.75pt}]  (339,40) -- (339,16) -- (37,15) -- (37,37) ;
            \draw [shift={(37,39)}, rotate = 270] [color={rgb, 255:red, 0; green, 0; blue, 0 }  ][line width=0.75]    (10.93,-3.29) .. controls (6.95,-1.4) and (3.31,-0.3) .. (0,0) .. controls (3.31,0.3) and (6.95,1.4) .. (10.93,3.29)   ;

            \draw (27,93) node [anchor=north west][inner sep=0.75pt]   [align=left] {$c_x$};
            \draw (247,60) node [anchor=north west][inner sep=0.75pt]   [align=left] {\begin{minipage}[lt]{38.76pt}\setlength\topsep{0pt}
            \begin{center}
            {\tiny Reconstruction }\\{\tiny Loss}
            \end{center}
            
            \end{minipage}};
            \draw (72.32,49) node [anchor=north west][inner sep=0.75pt]   [align=left] {\begin{minipage}[lt]{54.4pt}\setlength\topsep{0pt}
            \begin{center}
            {\scriptsize Pre-trained }\\{\scriptsize Stable Diffusion}\\{\scriptsize Model}
            \end{center}
            
            \end{minipage}};
            \draw (32,75) node [anchor=north west][inner sep=0.75pt]  [rotate=-270] [align=left] {{\tiny Noise}};
            \end{tikzpicture}
        }
        \\
        \\
        \scalebox{1.2}{
            \tikzset{every picture/.style={line width=0.75pt}} 
            \begin{tikzpicture}[x=0.75pt,y=0.75pt,yscale=-1,xscale=1]
            \draw  [fill={rgb, 255:red, 238; green, 228; blue, 137 }  ,fill opacity=1 ] (28.14,40) -- (43.83,40) -- (43.83,87.35) -- (28.14,87.35) -- cycle ;
            \draw  [fill={rgb, 255:red, 241; green, 191; blue, 115 }  ,fill opacity=1 ] (67.38,52.63) .. controls (67.38,45.65) and (73.03,40) .. (80,40) -- (141.07,40) .. controls (148.04,40) and (153.7,45.65) .. (153.7,52.63) -- (153.7,90.51) .. controls (153.7,97.48) and (148.04,103.13) .. (141.07,103.13) -- (80,103.13) .. controls (73.03,103.13) and (67.38,97.48) .. (67.38,90.51) -- cycle ;
            \draw    (43.83,63.67) -- (65.38,63.67) ;
            \draw [shift={(67.38,63.67)}, rotate = 180] [color={rgb, 255:red, 0; green, 0; blue, 0 }  ][line width=0.75]    (10.93,-3.29) .. controls (6.95,-1.4) and (3.31,-0.3) .. (0,0) .. controls (3.31,0.3) and (6.95,1.4) .. (10.93,3.29)   ;
            \draw    (47.76,99.19) -- (65.97,80.88) ;
            \draw [shift={(67.38,79.46)}, rotate = 134.84] [color={rgb, 255:red, 0; green, 0; blue, 0 }  ][line width=0.75]    (10.93,-3.29) .. controls (6.95,-1.4) and (3.31,-0.3) .. (0,0) .. controls (3.31,0.3) and (6.95,1.4) .. (10.93,3.29)   ;
            \draw (208.63,71.57) node  {\includegraphics[width=47.08pt,height=47.35pt]{sec/Images/ButterImg.jpg}};
            \draw    (153.7,64.07) -- (175.63,64.07) ;
            \draw [shift={(177.63,64.07)}, rotate = 180] [color={rgb, 255:red, 0; green, 0; blue, 0 }  ][line width=0.75]    (10.93,-3.29) .. controls (6.95,-1.4) and (3.31,-0.3) .. (0,0) .. controls (3.31,0.3) and (6.95,1.4) .. (10.93,3.29)   ;
            \draw    (177.24,79.46) -- (156.09,80.18) ;
            \draw [shift={(154.09,80.25)}, rotate = 358.05] [color={rgb, 255:red, 0; green, 0; blue, 0 }  ][line width=0.75]    (10.93,-3.29) .. controls (6.95,-1.4) and (3.31,-0.3) .. (0,0) .. controls (3.31,0.3) and (6.95,1.4) .. (10.93,3.29)   ;
            \draw (338.11,71.57) node  {\includegraphics[width=47.08pt,height=47.35pt]{sec/Images/ButterImg.jpg}};
            \draw  [dash pattern={on 7.5pt off 2.25pt}]  (306.72,71.57) -- (242.02,71.57) ;
            \draw [shift={(240.02,71.57)}, rotate = 360] [color={rgb, 255:red, 0; green, 0; blue, 0 }  ][line width=0.75]    (10.93,-3.29) .. controls (6.95,-1.4) and (3.31,-0.3) .. (0,0) .. controls (3.31,0.3) and (6.95,1.4) .. (10.93,3.29)   ;
            \draw  [dash pattern={on 7.5pt off 3.75pt}]  (339,40) -- (339,16) -- (37,15) -- (37,37) ;
            \draw [shift={(37,39)}, rotate = 270] [color={rgb, 255:red, 0; green, 0; blue, 0 }  ][line width=0.75]    (10.93,-3.29) .. controls (6.95,-1.4) and (3.31,-0.3) .. (0,0) .. controls (3.31,0.3) and (6.95,1.4) .. (10.93,3.29)   ;

            \draw (27,93) node [anchor=north west][inner sep=0.75pt]   [align=left] {$c_s$};
            \draw (247,60) node [anchor=north west][inner sep=0.75pt]   [align=left] {\begin{minipage}[lt]{38.76pt}\setlength\topsep{0pt}
            \begin{center}
            {\tiny Reconstruction }\\{\tiny Loss}
            \end{center}
            
            \end{minipage}};
            \draw (72.32,49) node [anchor=north west][inner sep=0.75pt]   [align=left] {\begin{minipage}[lt]{54.4pt}\setlength\topsep{0pt}
            \begin{center}
            {\scriptsize Pre-trained }\\{\scriptsize Stable Diffusion}\\{\scriptsize Model}
            \end{center}
            
            \end{minipage}};
            \draw (32,75) node [anchor=north west][inner sep=0.75pt]  [rotate=-270] [align=left] {{\tiny Noise}};
            \end{tikzpicture}
        }
        \\
        \\
        \scalebox{1.15}{
        \tikzset{every picture/.style={line width=0.75pt}} 

        \begin{tikzpicture}[x=0.75pt,y=0.75pt,yscale=-1,xscale=1]
        
        \draw  [fill={rgb, 255:red, 241; green, 191; blue, 115 }  ,fill opacity=1 ] (200,203.9) .. controls (200,196.22) and (206.22,190) .. (213.9,190) -- (278.6,190) .. controls (286.28,190) and (292.5,196.22) .. (292.5,203.9) -- (292.5,245.6) .. controls (292.5,253.28) and (286.28,259.5) .. (278.6,259.5) -- (213.9,259.5) .. controls (206.22,259.5) and (200,253.28) .. (200,245.6) -- cycle ;
        
        \draw (55,225) node  {\includegraphics[width=52.5pt,height=52.5pt]{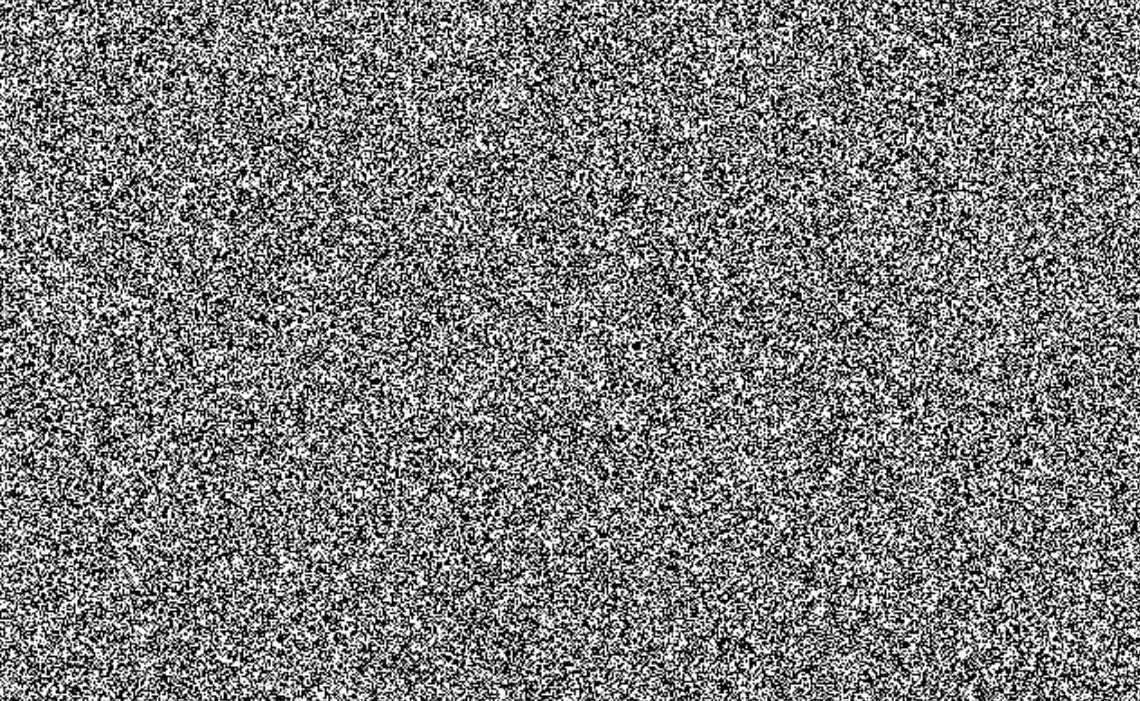}};
        \draw   [fill={rgb, 255:red, 174; green, 235; blue, 218 }  ,fill opacity=1 ](100,230) -- (170,230) -- (170,259.5) -- (100,259.5) -- cycle ;
        \draw    (90,210) -- (198,210) ;
        \draw [shift={(200,210)}, rotate = 180] [color={rgb, 255:red, 0; green, 0; blue, 0 }  ][line width=0.75]    (10.93,-3.29) .. controls (6.95,-1.4) and (3.31,-0.3) .. (0,0) .. controls (3.31,0.3) and (6.95,1.4) .. (10.93,3.29)   ;
        \draw    (170,246) -- (198,245.63) ;
        \draw [shift={(200,245.6)}, rotate = 179.24] [color={rgb, 255:red, 0; green, 0; blue, 0 }  ][line width=0.75]    (10.93,-3.29) .. controls (6.95,-1.4) and (3.31,-0.3) .. (0,0) .. controls (3.31,0.3) and (6.95,1.4) .. (10.93,3.29)   ;
        \draw    (292.5,227.5) -- (327.5,227.97) ;
        \draw [shift={(329.5,228)}, rotate = 180.77] [color={rgb, 255:red, 0; green, 0; blue, 0 }  ][line width=0.75]    (10.93,-3.29) .. controls (6.95,-1.4) and (3.31,-0.3) .. (0,0) .. controls (3.31,0.3) and (6.95,1.4) .. (10.93,3.29)   ;

        \draw (365.5,224.5) node  {\includegraphics[width=52.5pt,height=52.5pt]{sec/Images/ButterFlower.png}};
        
        \draw (108,232) node [anchor=north west][inner sep=0.75pt]   [align=left] {\begin{minipage}[lt]{41.48pt}\setlength\topsep{0pt}
        {\tiny Relation between}
        \begin{center}
        {\tiny  $c_x$ and $c_s$}
        \end{center}
        
        \end{minipage}};
        \draw (207.25,200) node [anchor=north west][inner sep=0.75pt]   [align=left] {\begin{minipage}[lt]{54.4pt}\setlength\topsep{0pt}
        \begin{center}
        {\scriptsize Fine-tuned }\\{\scriptsize Stable Diffusion}\\{\scriptsize Model}
        \end{center}
        
        \end{minipage}};
        \draw (353.5,258) node [anchor=north west][inner sep=0.75pt]   [align=left] {{\tiny Output}};

        \end{tikzpicture}
        
        }    
    \end{tabular}
    \caption{We optimize the Stable Diffusion Model~\cite{Rombach_2022_CVPR} with the determined classes to reconstruct the input images. Form the classes we generate a relation between them, which is used to generate the output}
    \label{fig:fine-tune}
\end{figure}

\begin{figure*}[ht]
        \centering
        \begin{tabular}{ccccc}
            Sketch & ControlNet & ConditionFlow & Setting 1 & Setting 2\\
            \includegraphics[width=0.15\textwidth]{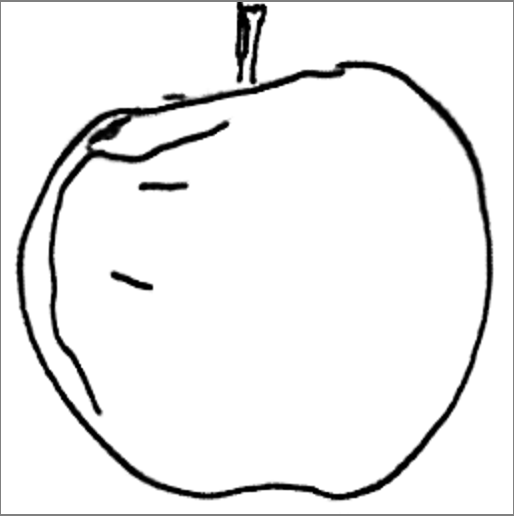}& 
            \includegraphics[width=0.15\textwidth]{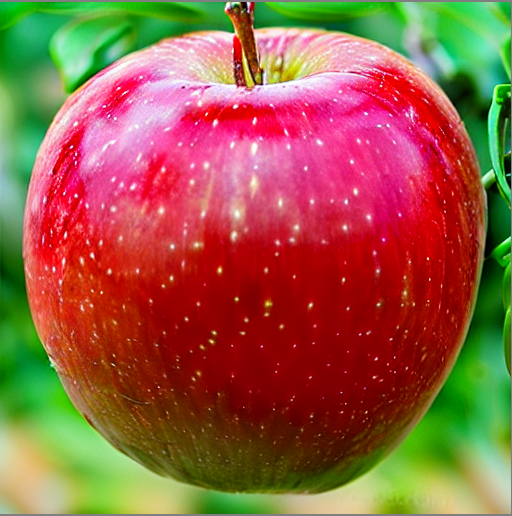}&
            \includegraphics[width=0.15\textwidth]{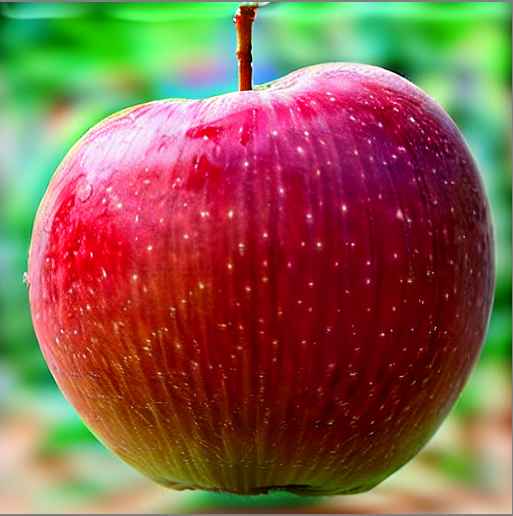}&
            \includegraphics[width=0.15\textwidth]{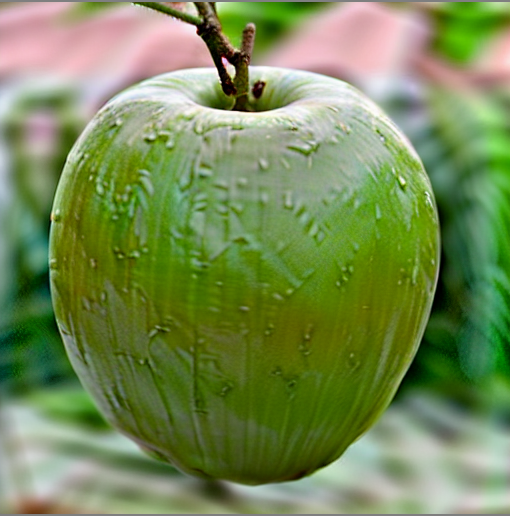}&
            \includegraphics[width=0.15\textwidth]{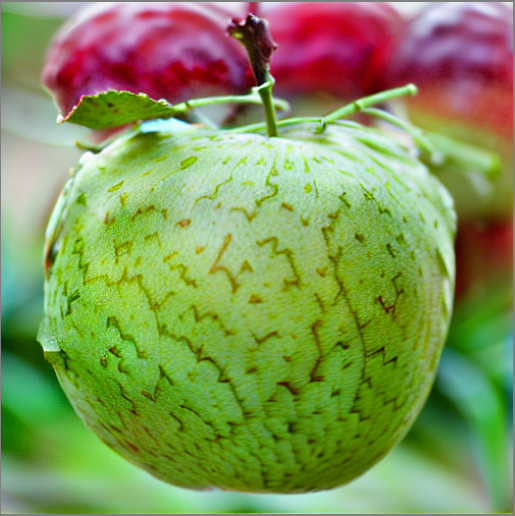}\\
             \includegraphics[width=0.15\textwidth]{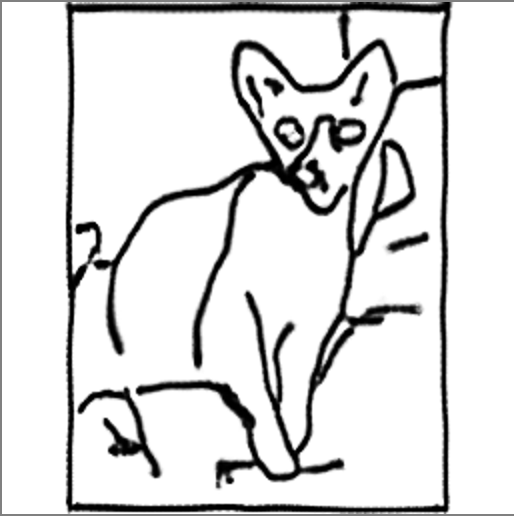}& 
            \includegraphics[width=0.15\textwidth]{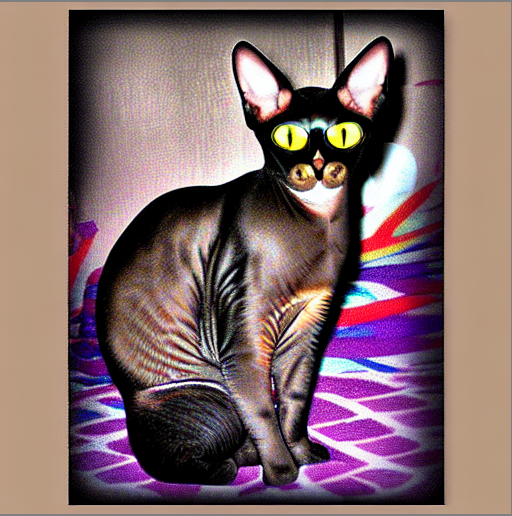}&
            \includegraphics[width=0.15\textwidth]{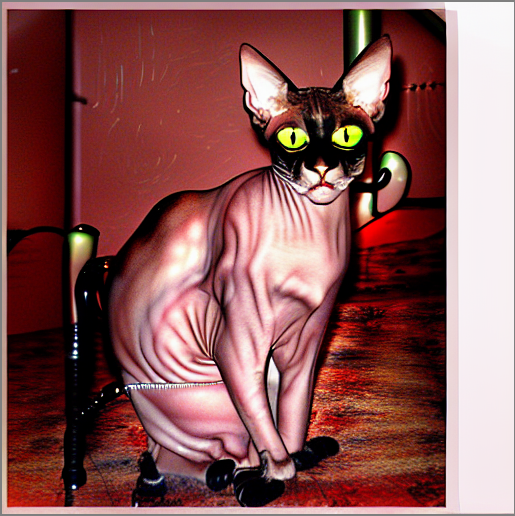}&
            \includegraphics[width=0.15\textwidth]{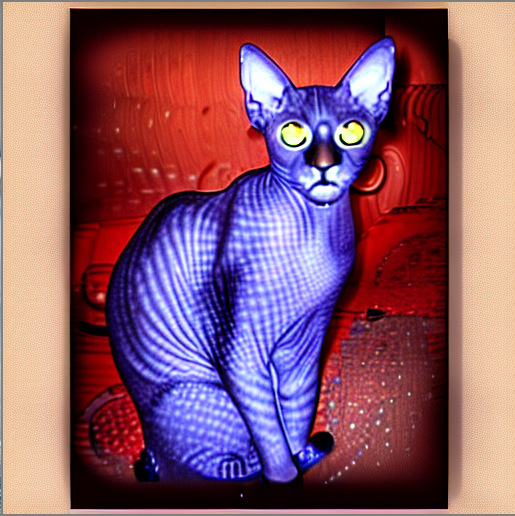}&
            \includegraphics[width=0.15\textwidth]{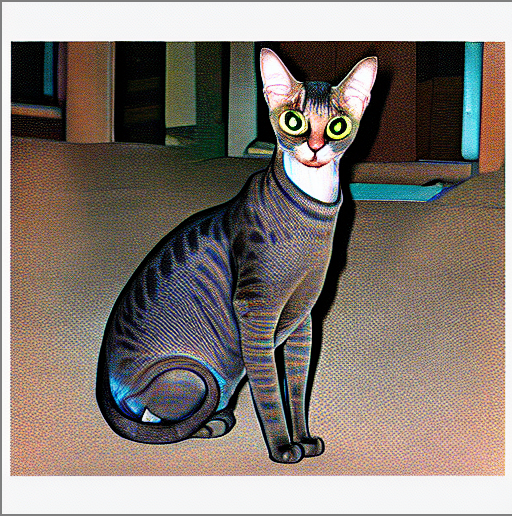}\\
             \includegraphics[width=0.15\textwidth]{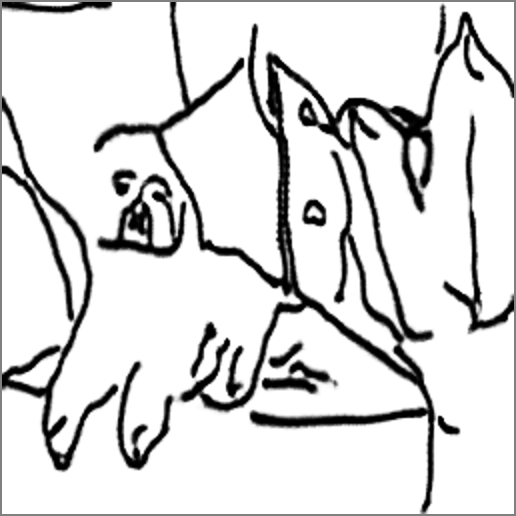}& 
            \includegraphics[width=0.15\textwidth]{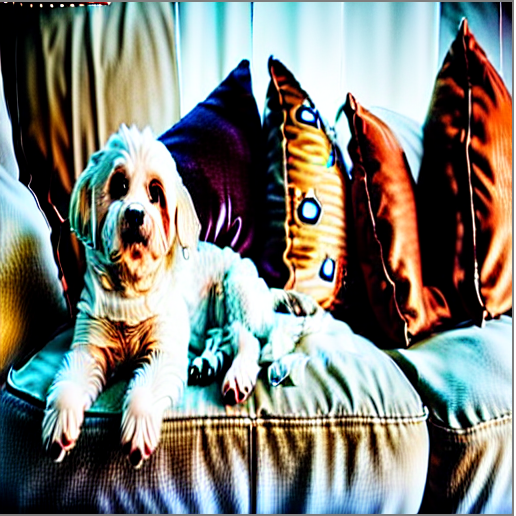}&
            \includegraphics[width=0.15\textwidth]{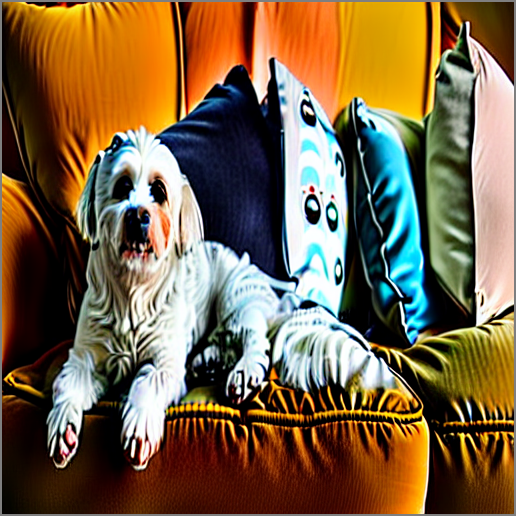}&
            \includegraphics[width=0.15\textwidth]{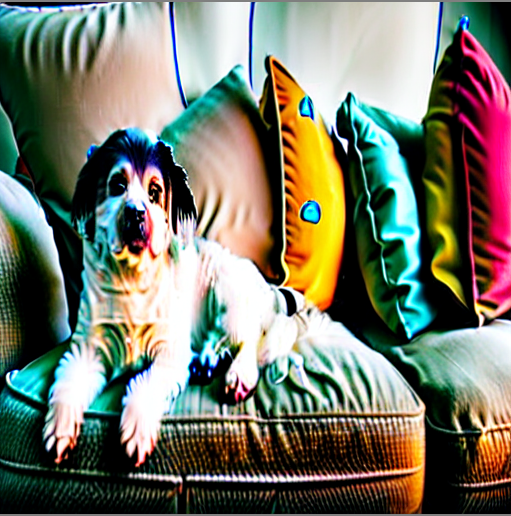}&
            \includegraphics[width=0.15\textwidth]{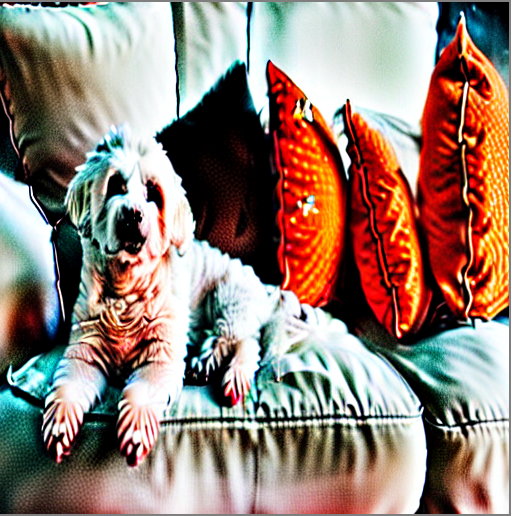}
        \end{tabular}
        \caption{Here we are comparing different settings we tested for better sketch-to-image generation. ControlNet and ConditionFlow columns represent the outputs generated from the baseline model and our modified model. We also tested 2 other configurations. Those are explained in the Section.~\ref{sec: condflow}. We listed their outputs here.}
        \label{fig:abl-cf}
    \end{figure*}

Initially, we ascertain the classes of the provided images and sketches from the CLIP classification~\cite{radford2021learning}, subsequently optimizing the relation between the determined classes using ConceptNet~\cite{speer2017conceptnet}. Using the ConceptNet we list the common relation between the determined classes and select the topmost relation for conditioning of the image generation. Hence, to offer artistic control over the image customization, we abstain from offering any external prompts, instead we use the generate the relation between the provided sketch and image as a condition for custom image generation. After this, we generate an image from the sketch using our proposed \textit{ConditionFlow} model. Subsequently, we adjust the level of fine-tuning of the model. Normally the model is fine-tuned for its customization and to recreate a given image. We found in our experiments that if we run the fine-tune process various duration the reconstructrability of the image varies and with a controlled fine-tuning we can customize the model to reconstruct more than one image. In this fine-tuning process we are tying to minimize our reconstruction loss in each iteration. Consequently, our loss function for fine-tuning for two images is

\begin{gather}
\mathcal{L} (x,s,c_{x} ,c_{s} ,\theta )=\mathbb{E}_{t,t',\epsilon ,\epsilon '}[ w_{t} * \| \epsilon _{\theta } (x_{t} ,t,c_{x} )-\epsilon \| _{2}^{2} + \notag\\
 w_{t'} * \lambda \| \epsilon _{\theta } (x_{t'} ,t',c_{s} )-\epsilon '\| _{2}^{2}],\nonumber
\end{gather}
where $x$ represents the supplied image, while $s$ stands for the sketch, and $x_t$ denotes the image generated from the sketch. The terms $c_x$ and $c_s$ are classes of the image and the sketch, respectively. The values for $\lambda$ and $w_t, w_t'$ are determined based on the areas that the image and the sketch occupy and experimental hyper-parameters. These parameters are used in the $\epsilon _{\theta }$ network to predict the noises of $x_t$ and $x_{t'}$.

It is noteworthy to mention that the contemporary state-of-the-art model for customization, as proposed by Tewel et al.~\cite{tewel2023key}, requires approximately 4 minutes for execution. Their method locks concept cross-attention keys to their superordinate category and utilizes gated rank-1 editing, which controls the influence of the newly learned concept on the image generation model. In contrast, our approach fine-tunes the weight of the image generation module of the stable diffusion model and minimizes the loss function to reconstruct the input images. This process of loss minimization is faster than the technique introduced by Tewel et al. Our method needs 1-1.5 minutes for each concept to fine-tune the model. Additionally, the hyper-parameters generated from our distinctive area coverage approach restrict the model customizing duration, which mitigates the over-fitting of the model. Consequently, our method facilitates the generation of morphed images spanning multiple classes. For an overview of the pipeline at inference, see Figure~\ref{fig:overview}.

\begin{figure*}
    \centering
    \begin{tabular}{cccccc}
    Examples & DreamBooth & Textual Inversion & Custom Diffusion & Ours & External Condition\\
    \begin{adjustbox}{max width=0.14\textwidth}
    \begin{tabular}{cc}
        \includegraphics{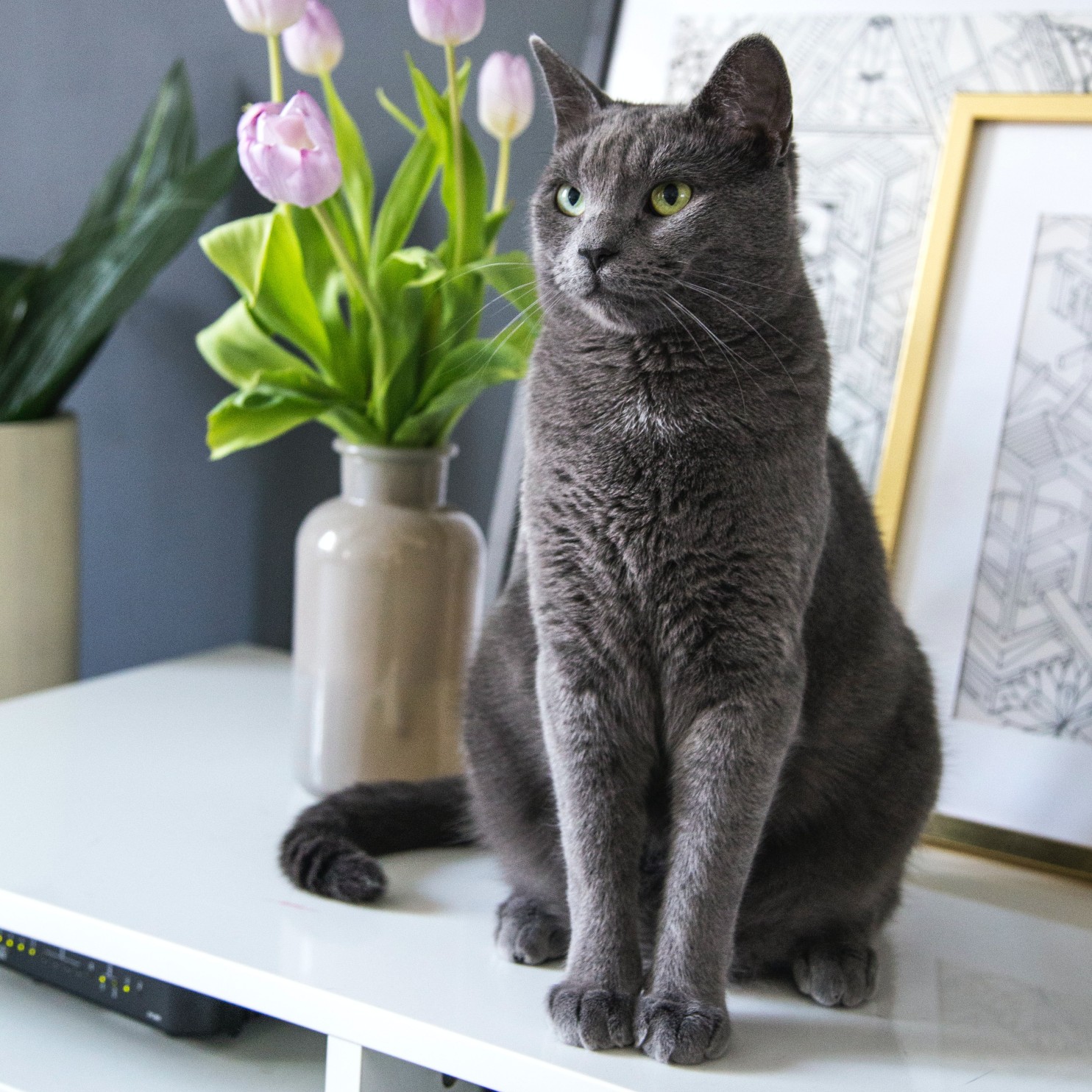} &
        \includegraphics{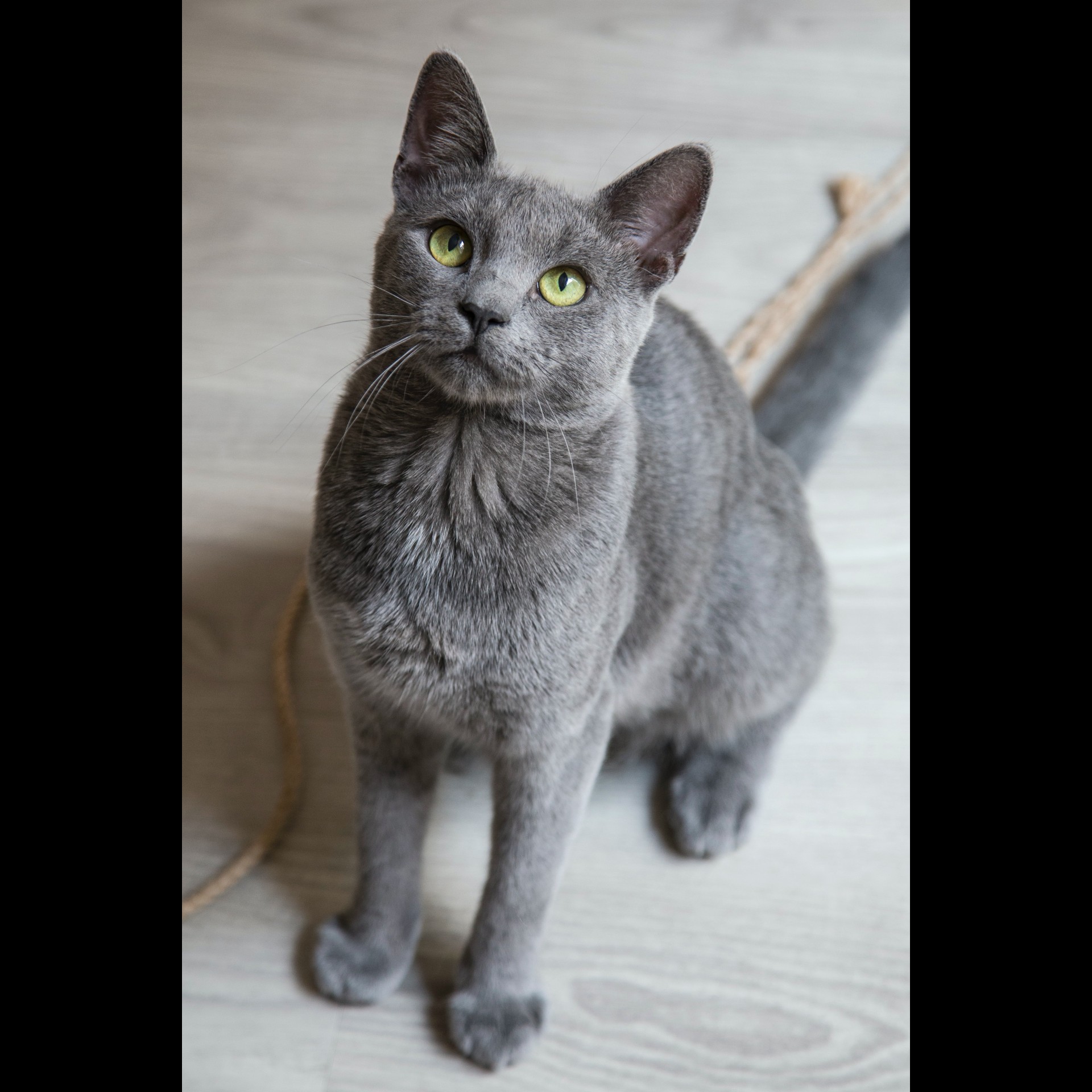}\\
        \includegraphics{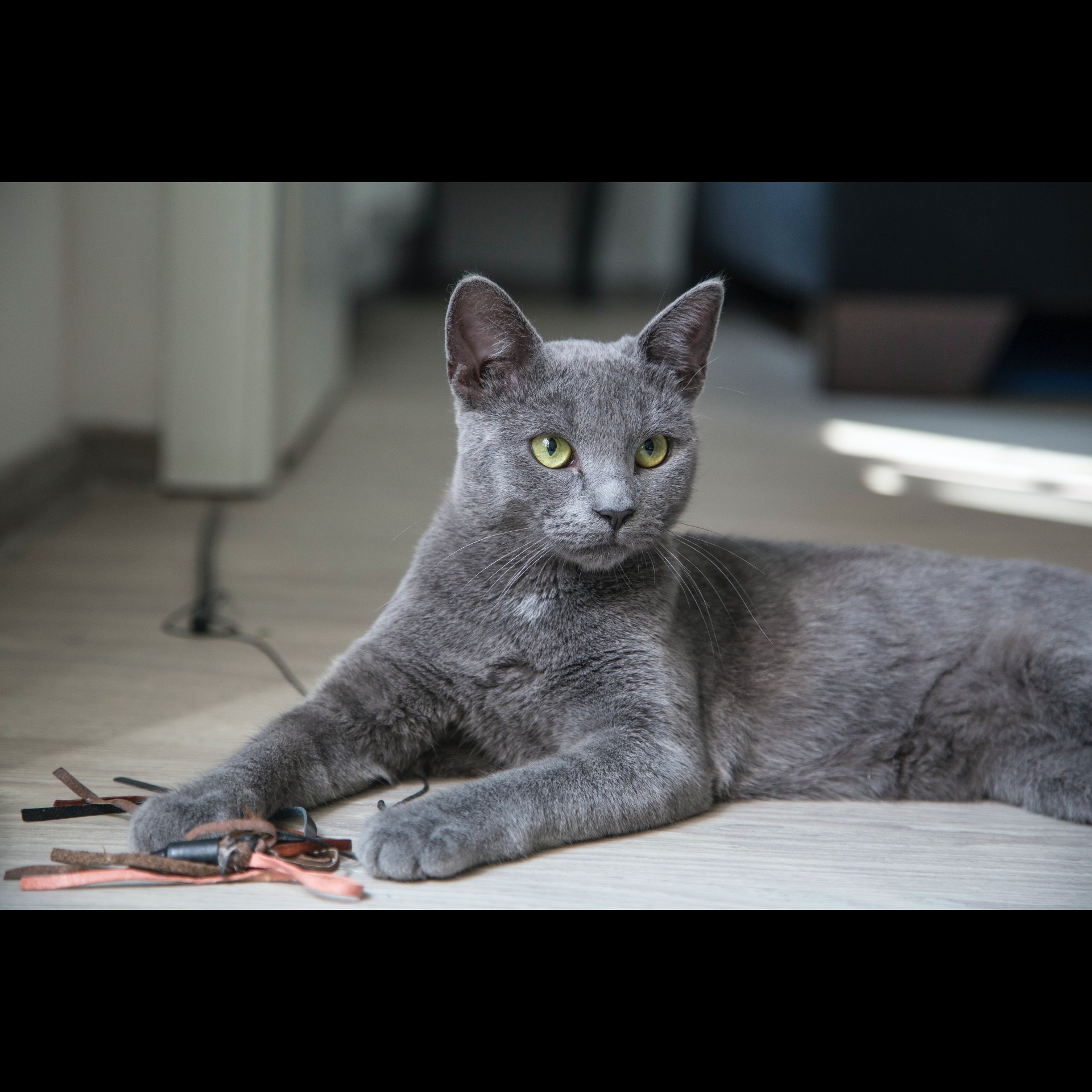} &
        \includegraphics{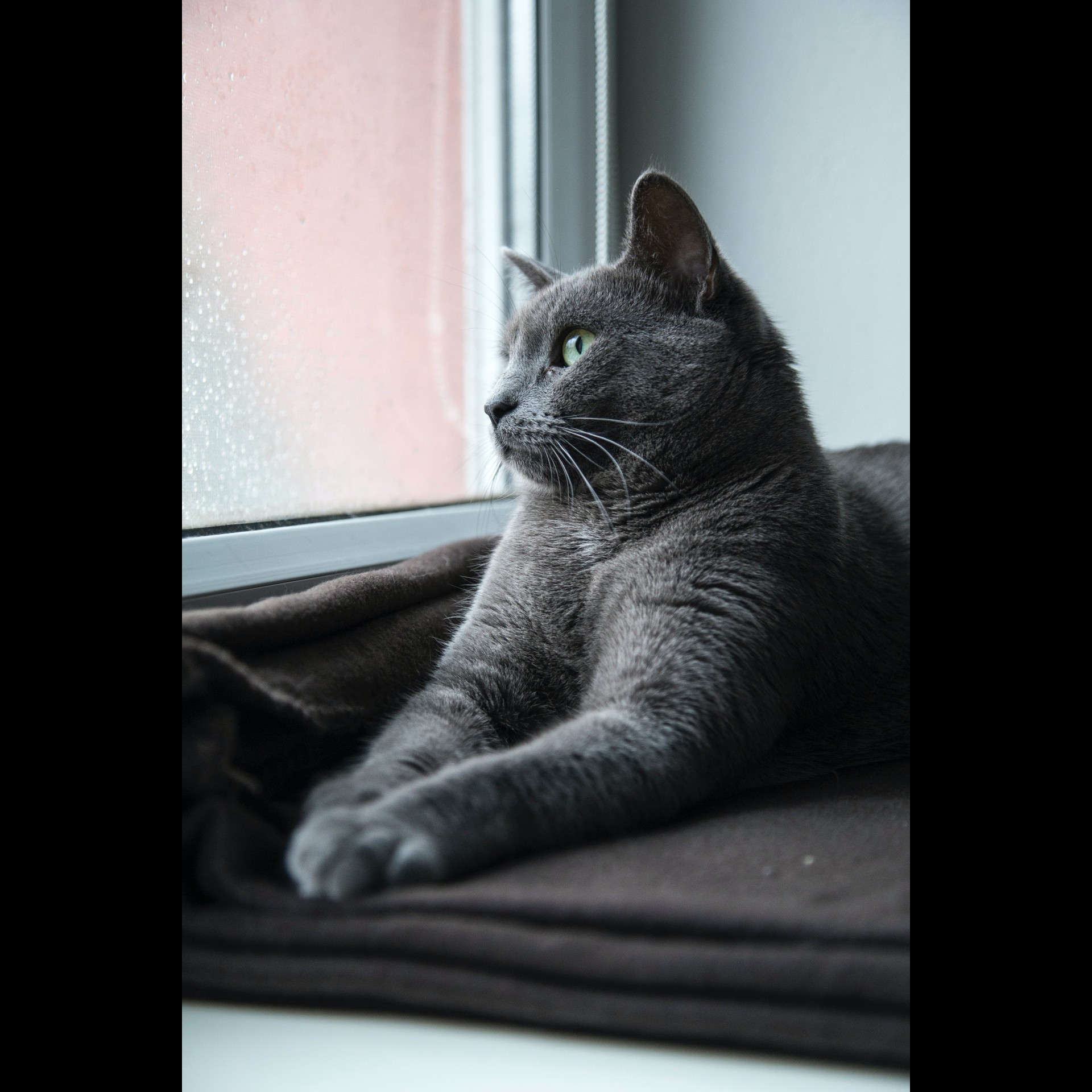}\\ 
    \end{tabular}
    \end{adjustbox}&
    \begin{adjustbox}{max width=0.14\textwidth}
    \begin{tabular}{cc}
        \multicolumn{2}{c}{\includegraphics[scale=2]{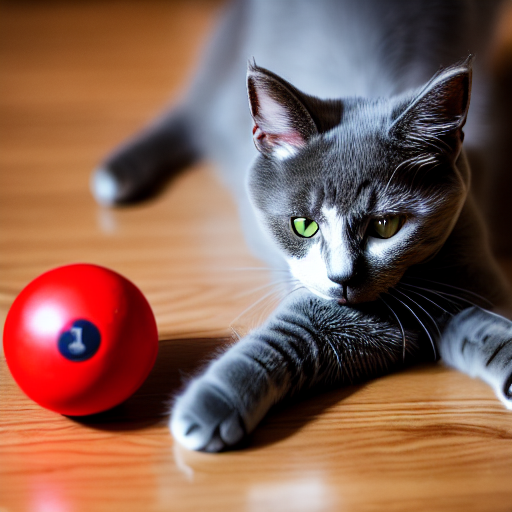}} \\
        \includegraphics{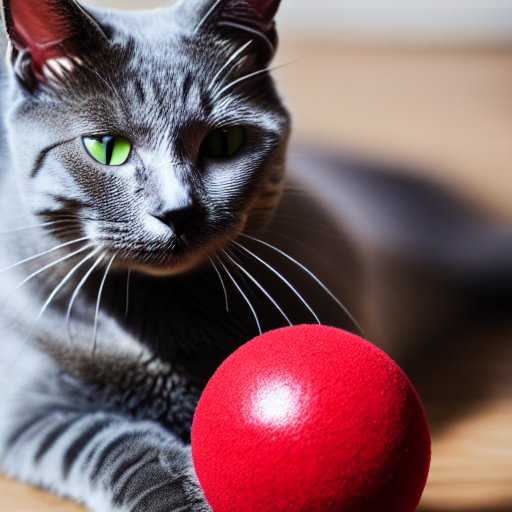} & \includegraphics{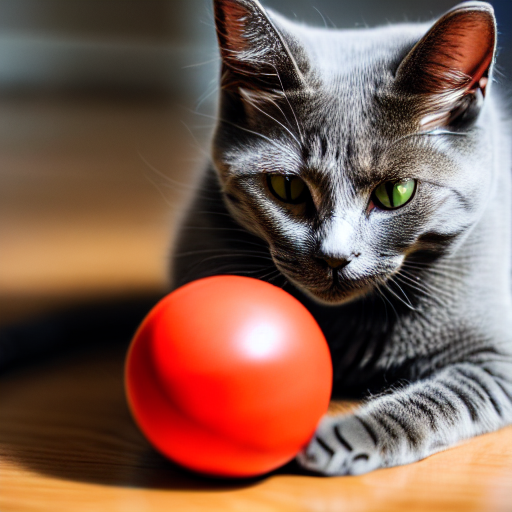} 
    \end{tabular}
    \end{adjustbox} &
    \begin{adjustbox}{max width=0.14\textwidth}
        \begin{tabular}{cc}
        \multicolumn{2}{c}{\includegraphics[scale=2]{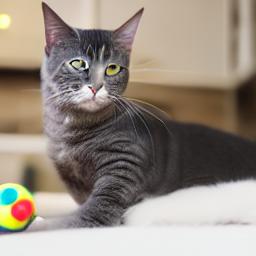}} \\
        \includegraphics{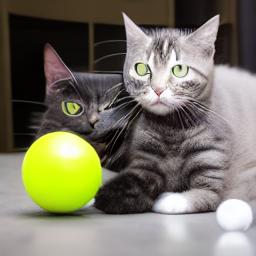} &
        \includegraphics{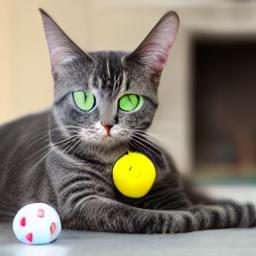} 
    \end{tabular}
    \end{adjustbox} &
    \begin{adjustbox}{max width=0.14\textwidth}
        \begin{tabular}{cc}
        \multicolumn{2}{c}{\includegraphics[scale=2]{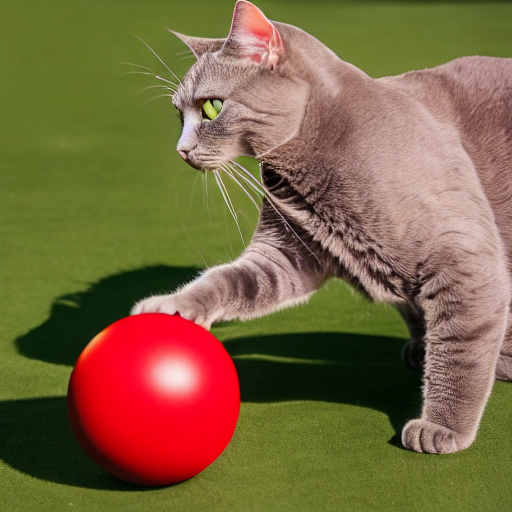}} \\
        \includegraphics{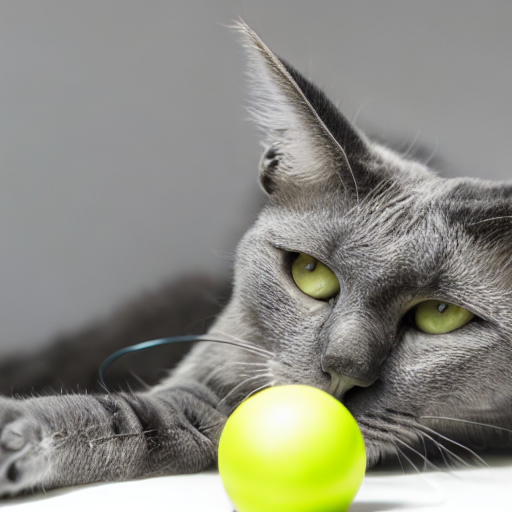} &
        \includegraphics{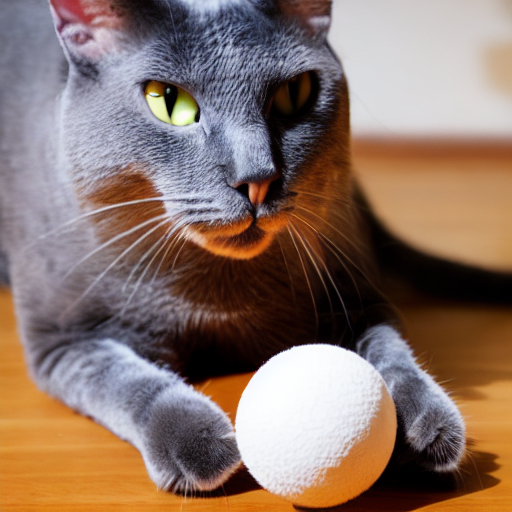}
    \end{tabular}
    \end{adjustbox} &
    \begin{adjustbox}{max width=0.14\textwidth}
    \begin{tabular}{cc}
        \multicolumn{2}{c}{\includegraphics[scale=2]{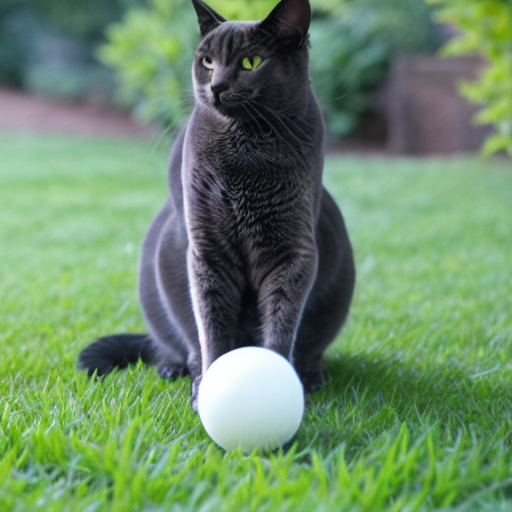}} \\
        \includegraphics{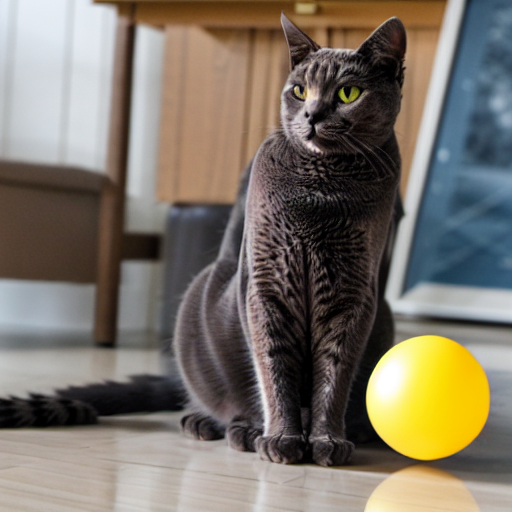} & \includegraphics{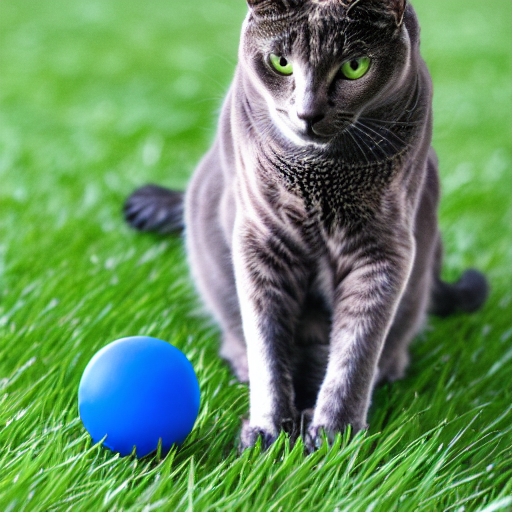} 
    \end{tabular}
    \end{adjustbox} &
    \begin{adjustbox}{max width=0.1\textwidth}
    \includegraphics{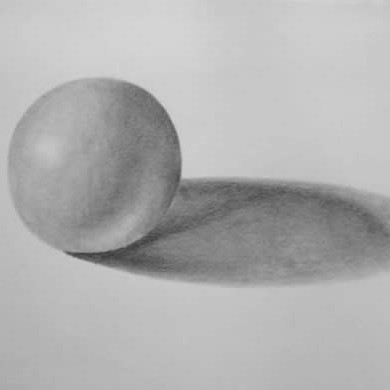}
    \end{adjustbox}\\
     & \multicolumn{3}{c}{A cat* is playing with a ball} & \multicolumn{2}{c}{}\\
     \\
    \begin{adjustbox}{max width=0.14\textwidth}
    \begin{tabular}{cc}
        \includegraphics{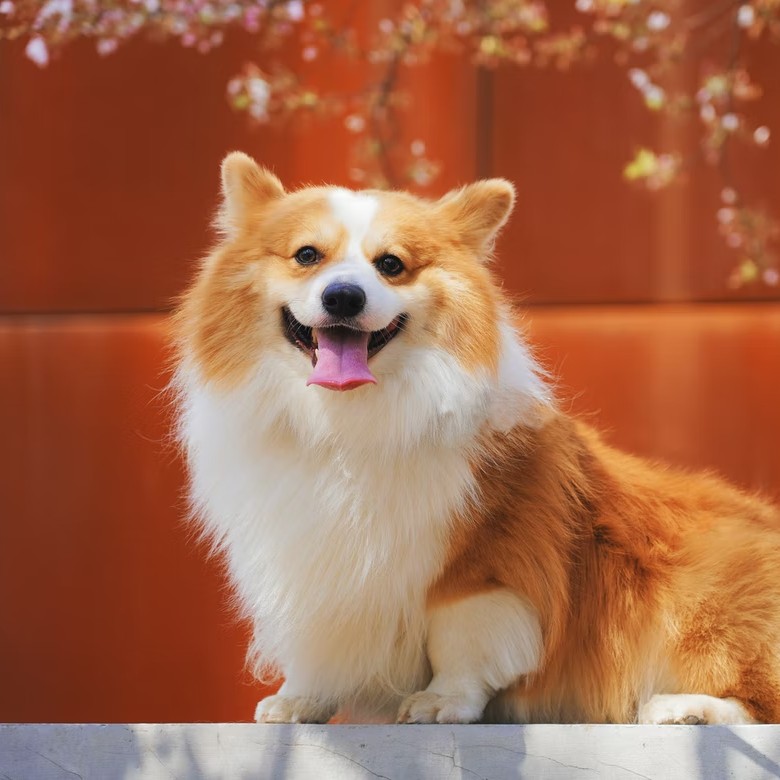} &
        \includegraphics{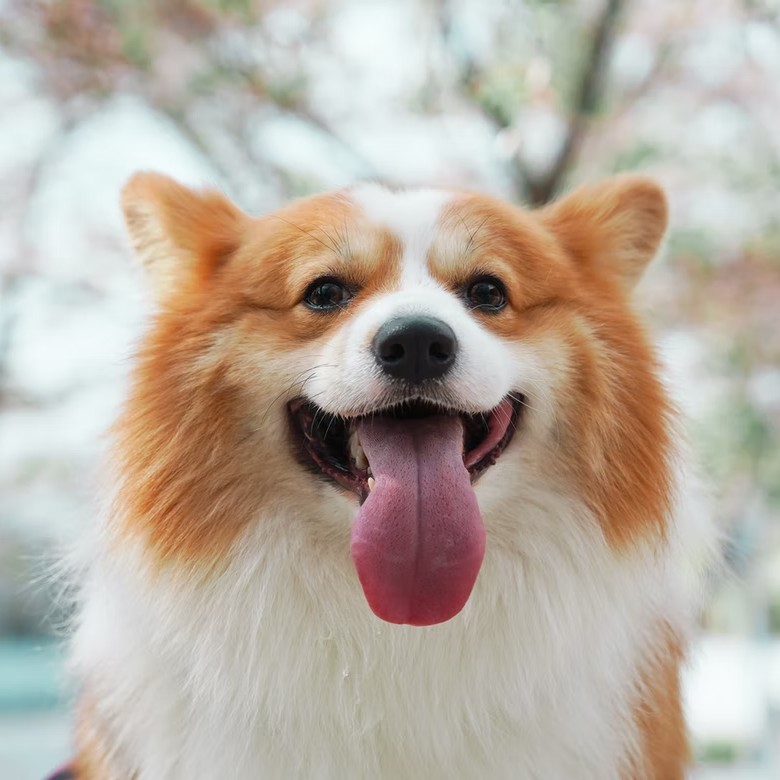}\\
        \includegraphics{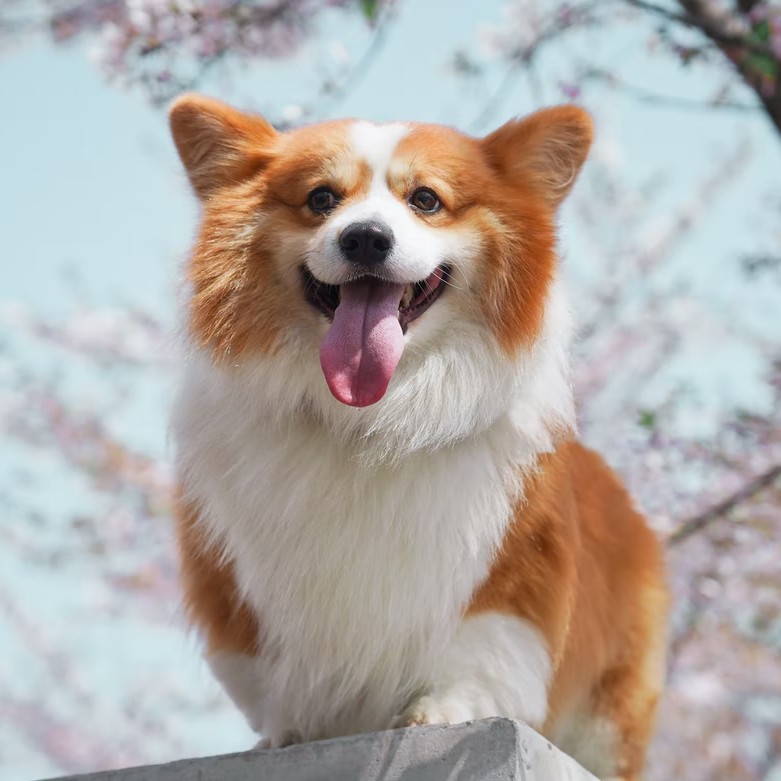} &
        \includegraphics{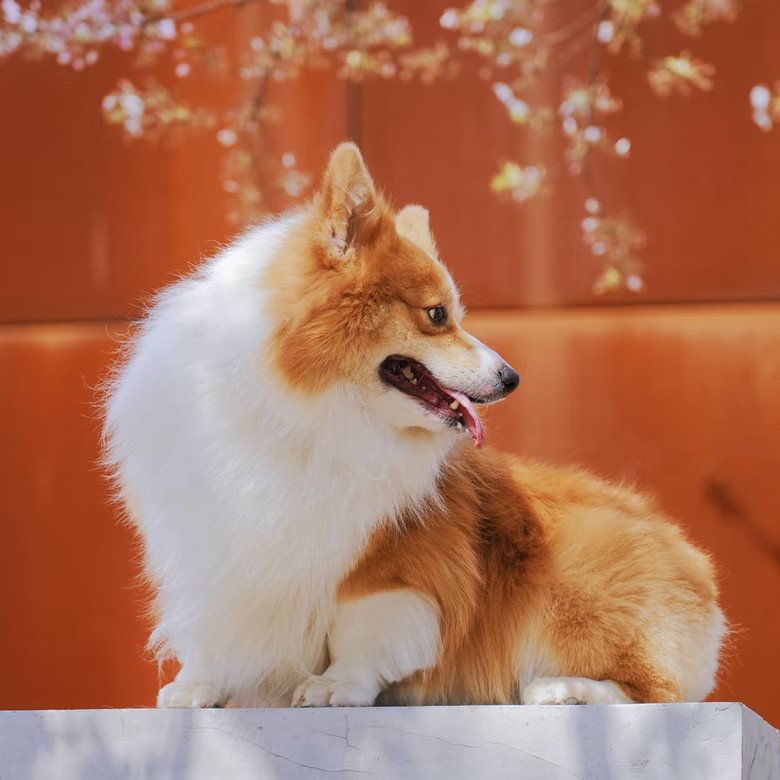}\\ 
    \end{tabular}
    \end{adjustbox}&
    \begin{adjustbox}{max width=0.14\textwidth}
    \begin{tabular}{cc}
        \multicolumn{2}{c}{\includegraphics[scale=2]{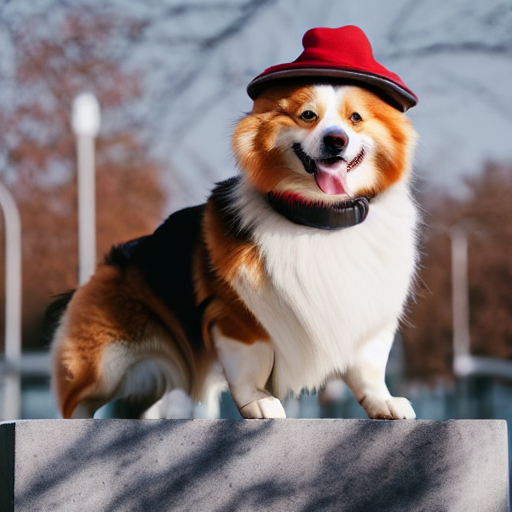}} \\
        \includegraphics{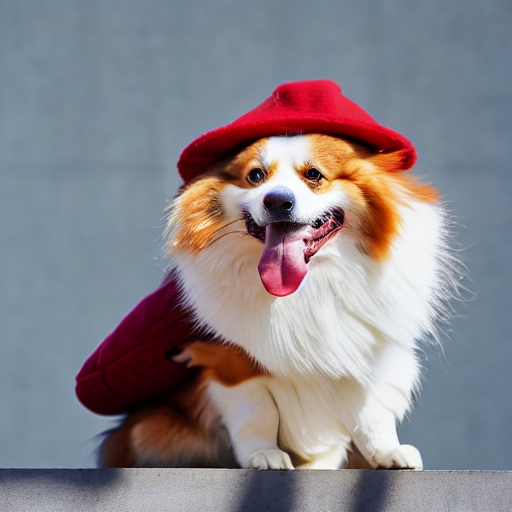} & 
        \includegraphics{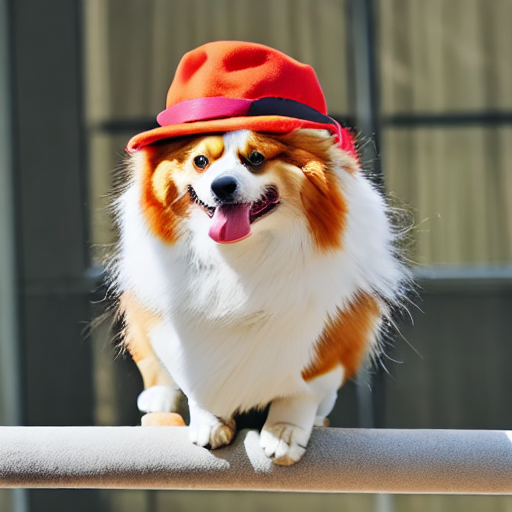} 
    \end{tabular}
    \end{adjustbox} &
    \begin{adjustbox}{max width=0.14\textwidth}
        \begin{tabular}{cc}
        \multicolumn{2}{c}{\includegraphics[scale=2]{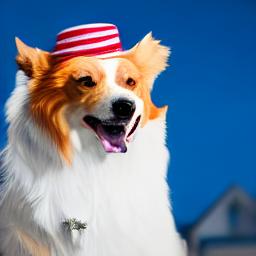}} \\
        \includegraphics{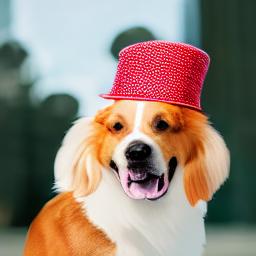} &
        \includegraphics{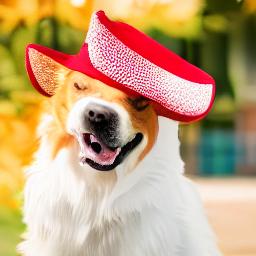} 
    \end{tabular}
    \end{adjustbox} &
    \begin{adjustbox}{max width=0.14\textwidth}
        \begin{tabular}{cc}
        \multicolumn{2}{c}{\includegraphics[scale=2]{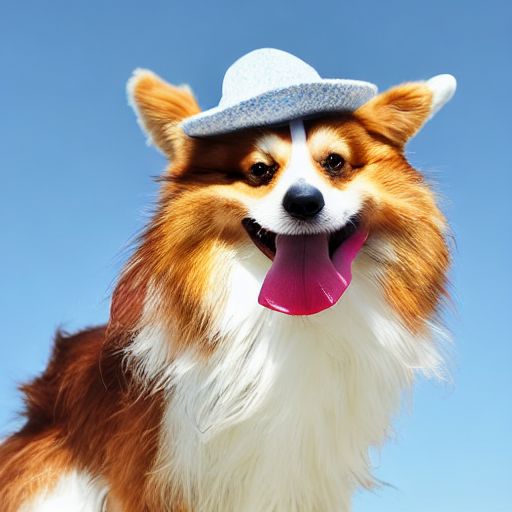}} \\
        \includegraphics{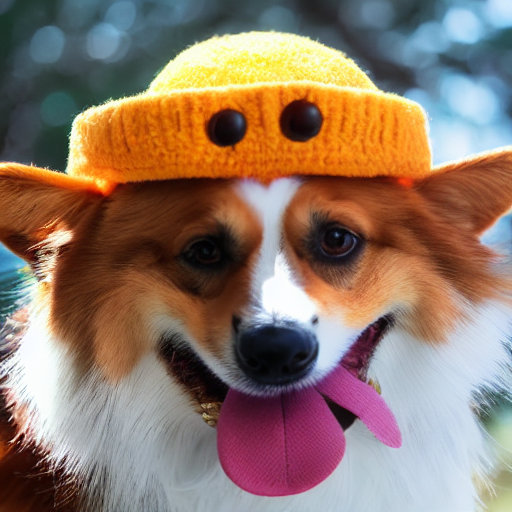} &
        \includegraphics{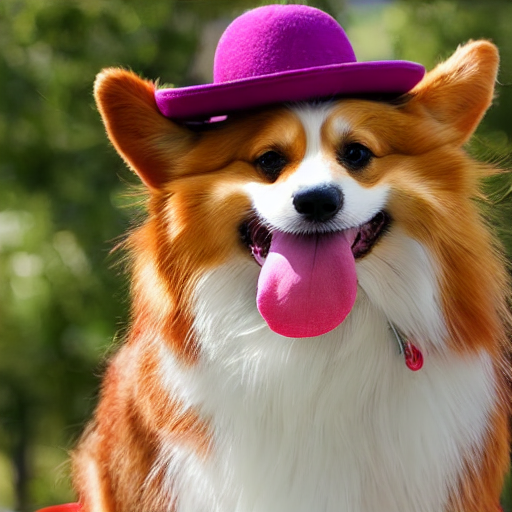}
    \end{tabular}
    \end{adjustbox} &
    \begin{adjustbox}{max width=0.14\textwidth}
    \begin{tabular}{cc}
        \multicolumn{2}{c}{\includegraphics[scale=2]{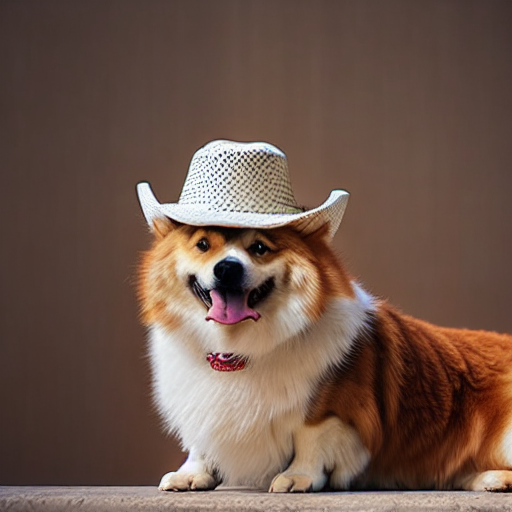}} \\
        \includegraphics{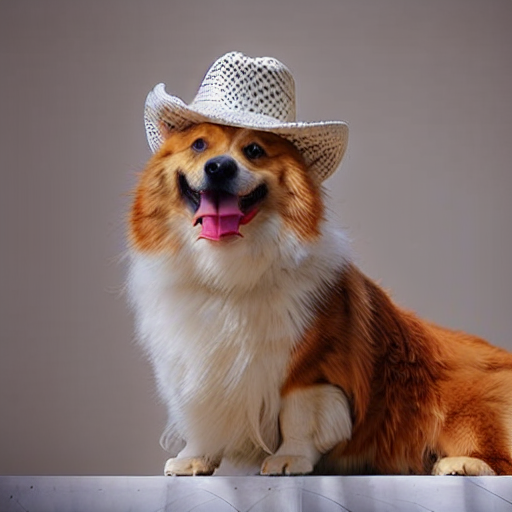} & \includegraphics{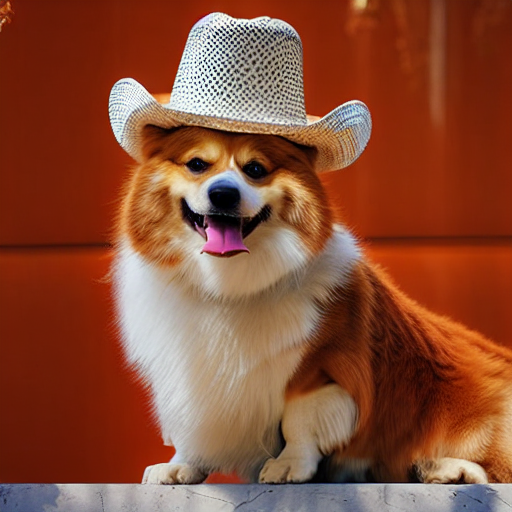} 
    \end{tabular}
    \end{adjustbox} &
    \begin{adjustbox}{max width=0.1\textwidth}
    \includegraphics{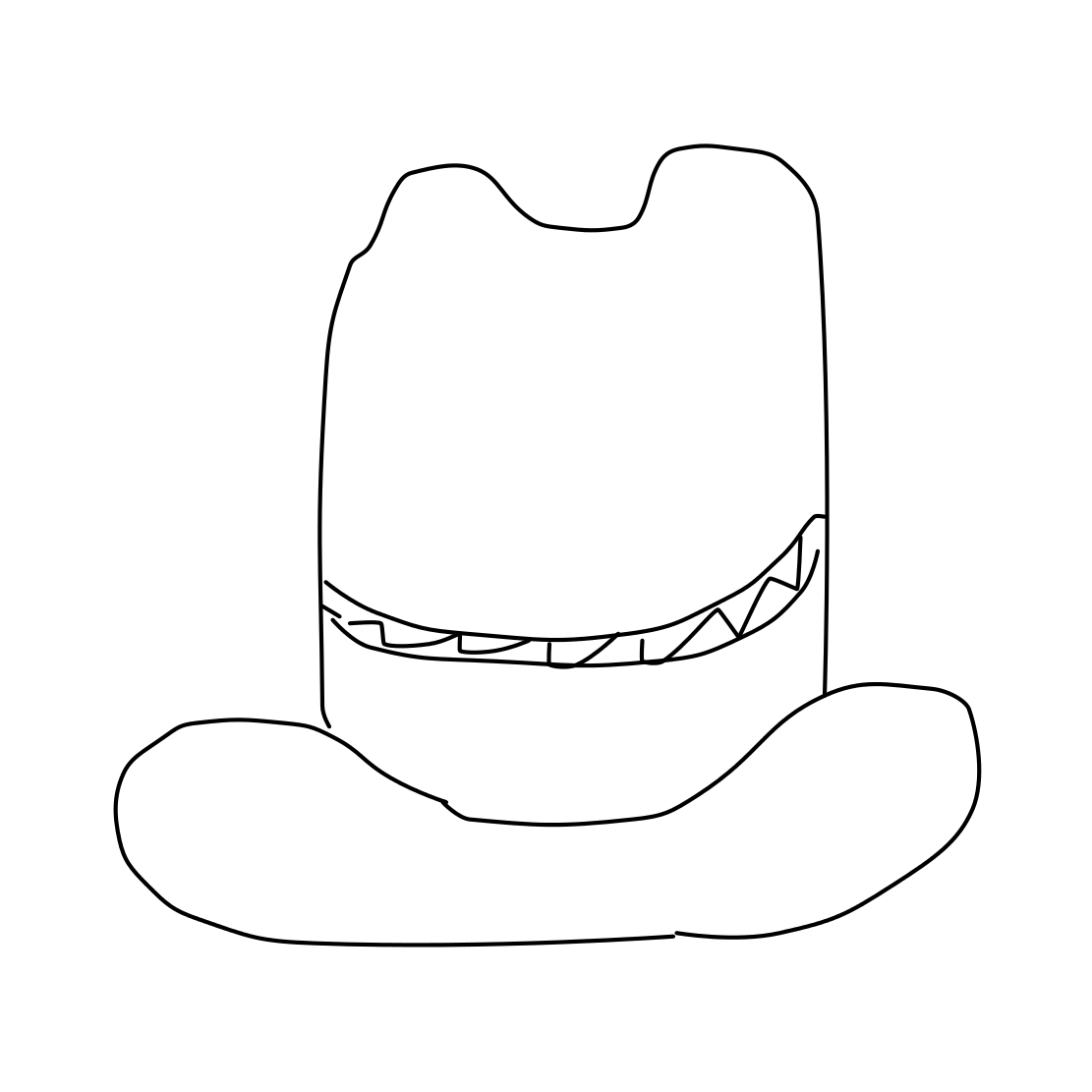}
    \end{adjustbox}\\
     & \multicolumn{3}{c}{A dog* is wearing a hat} & \multicolumn{2}{c}{}\\
     \\
    \begin{adjustbox}{max width=0.14\textwidth}
    \begin{tabular}{cc}
        \includegraphics{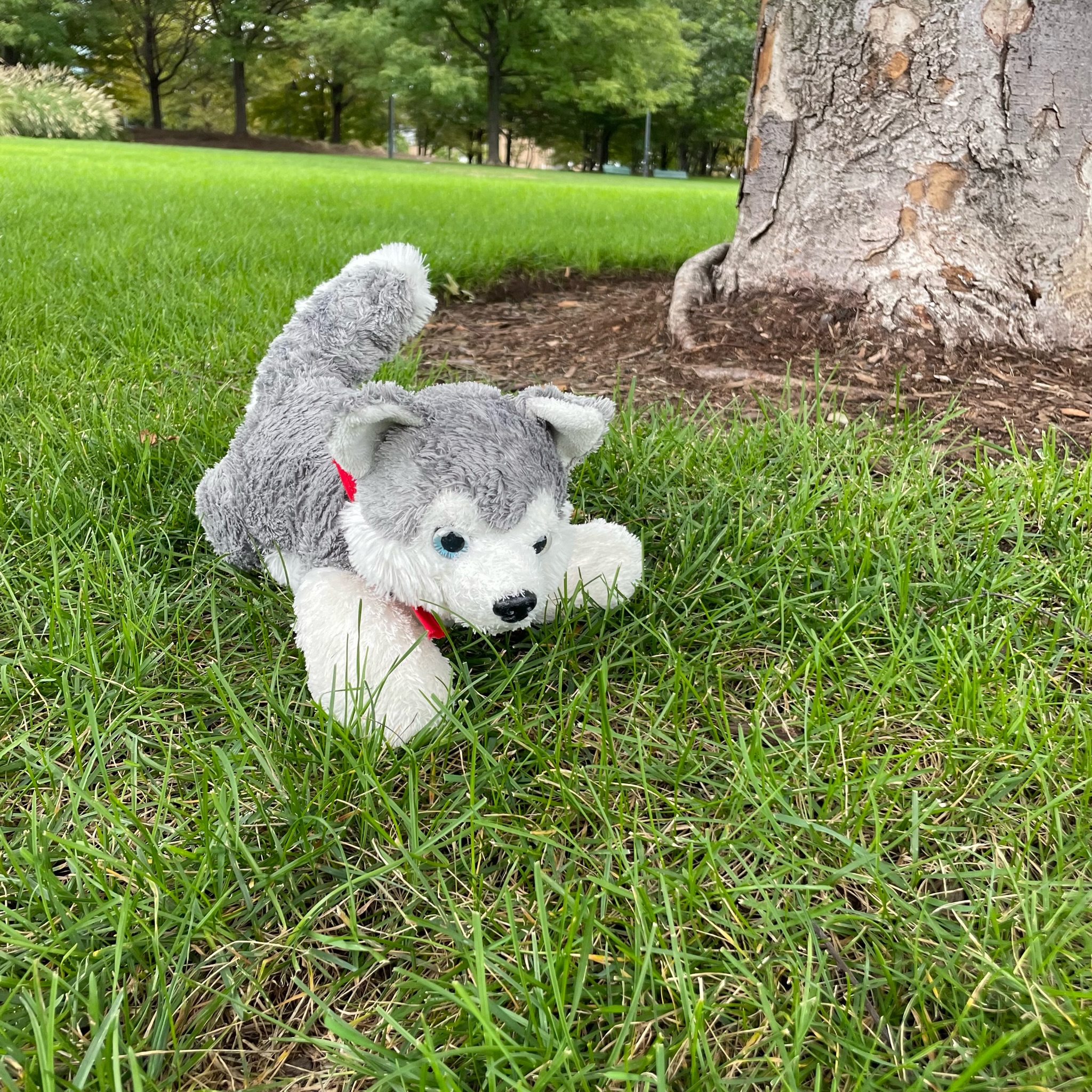} &
        \includegraphics{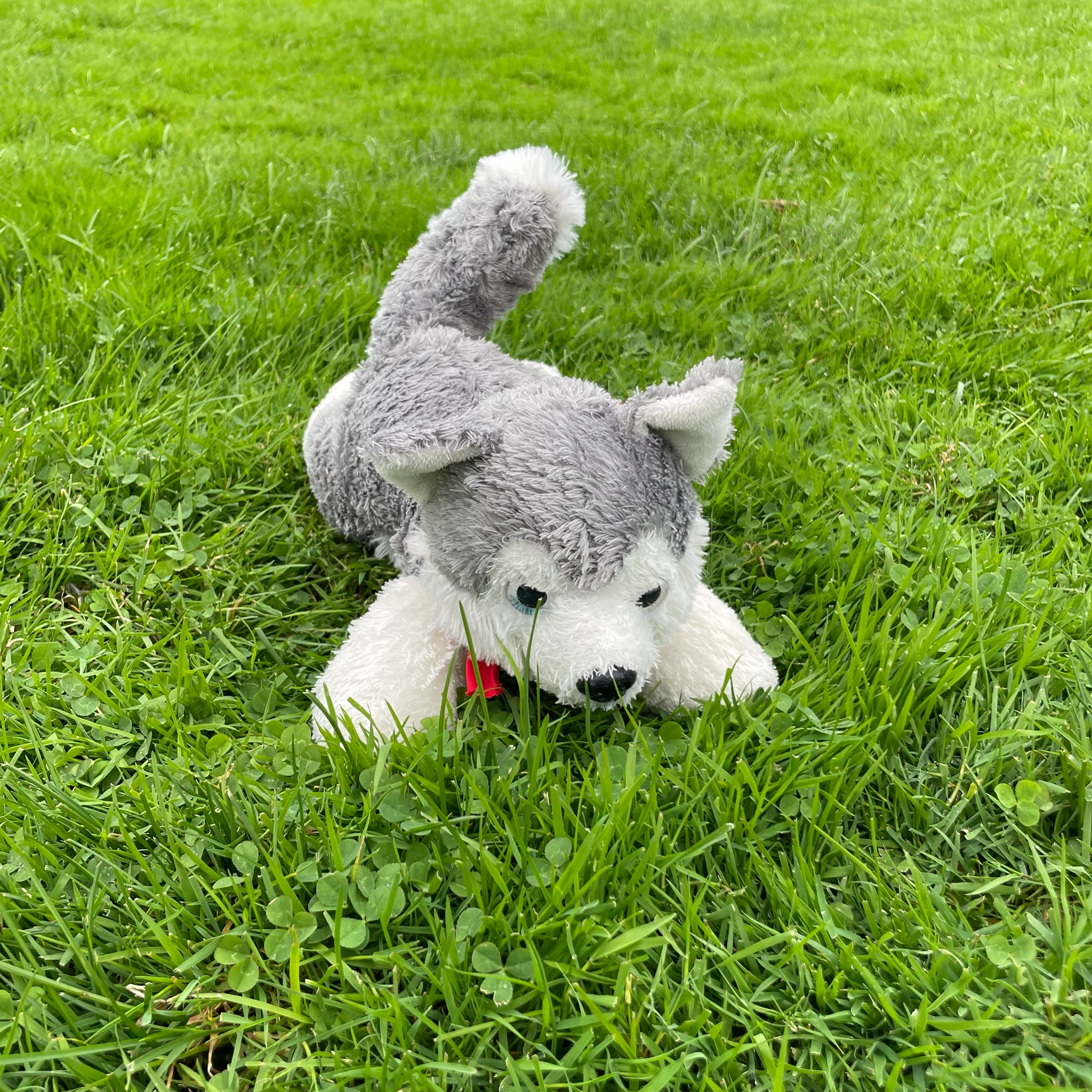}\\
        \includegraphics{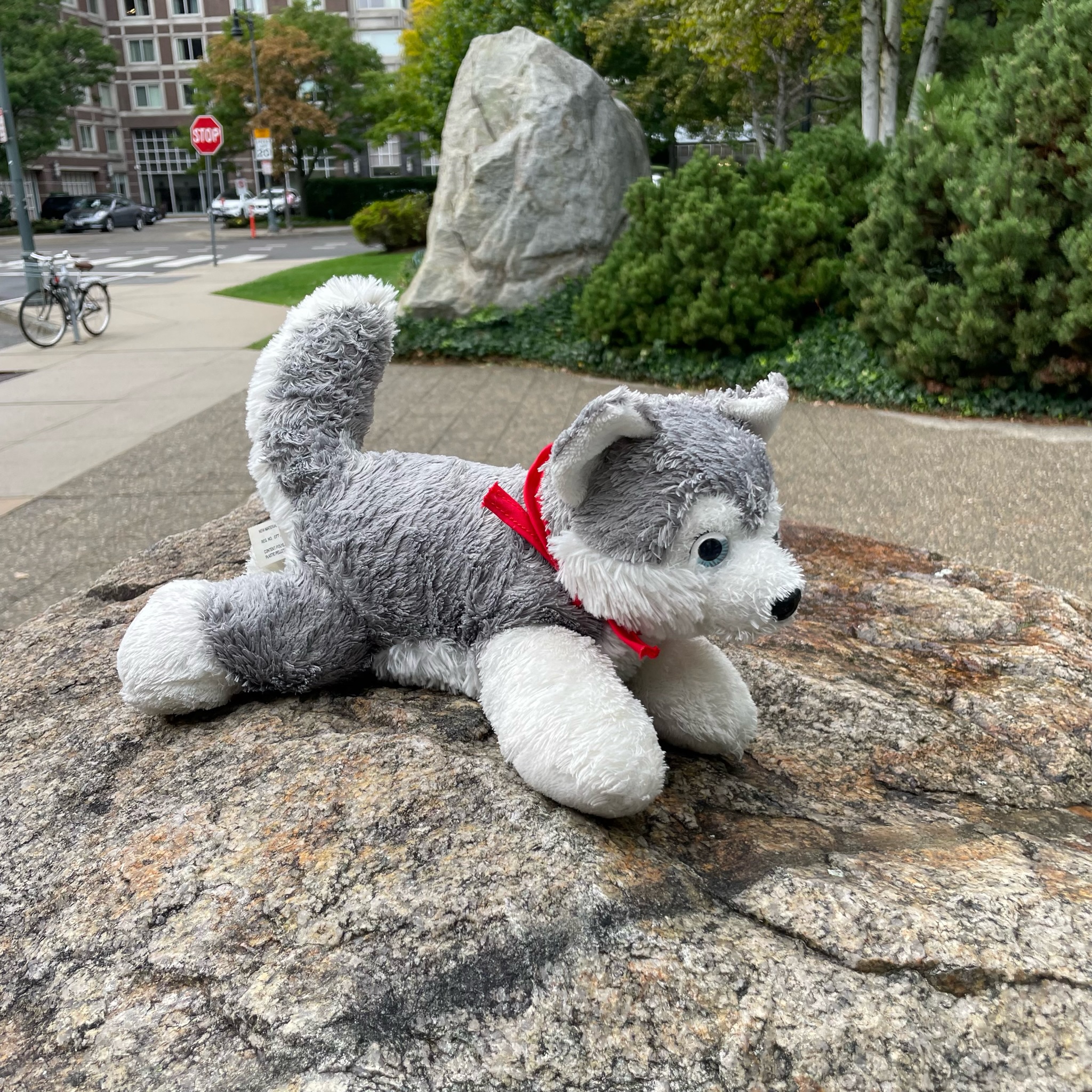} &
        \includegraphics{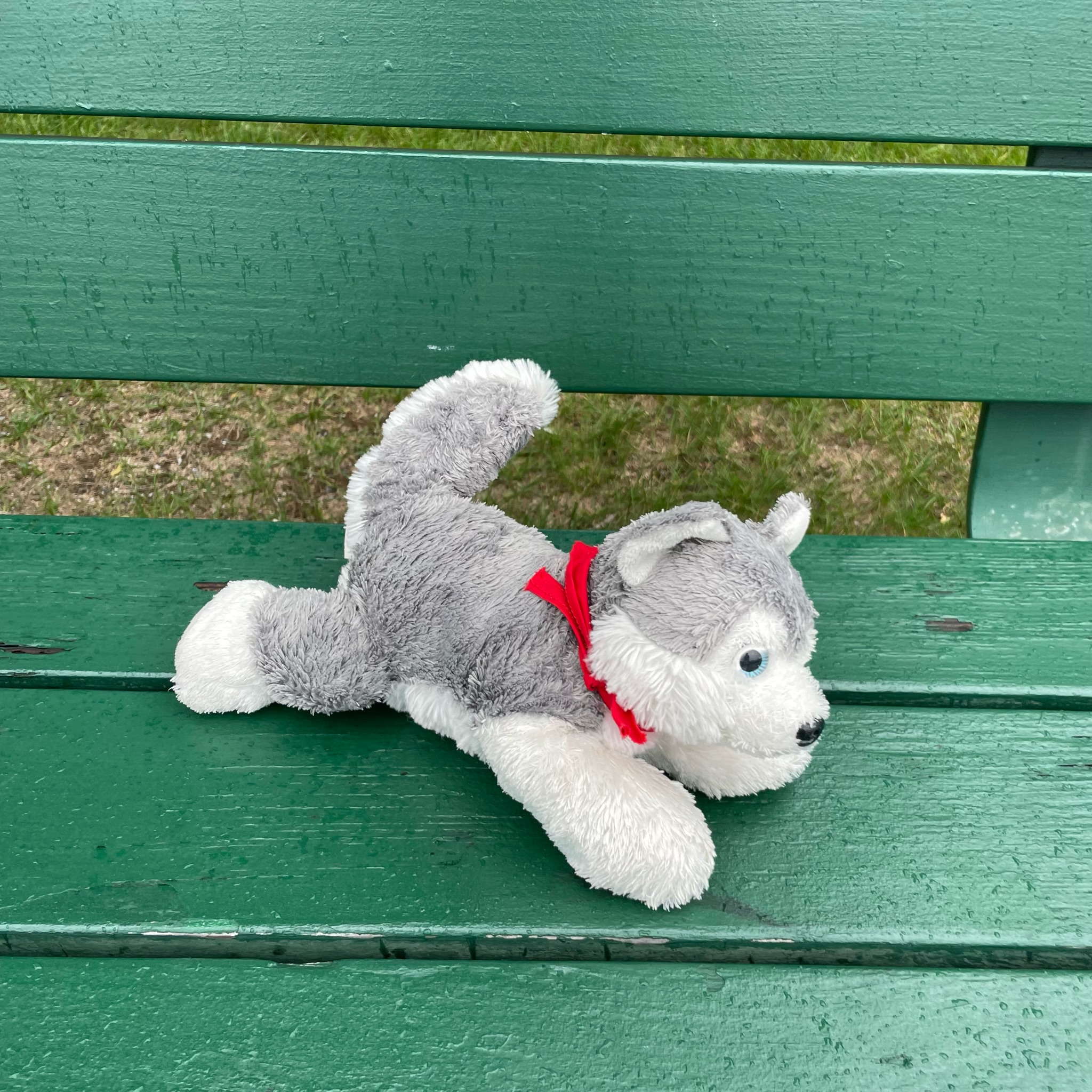}\\ 
    \end{tabular}
    \end{adjustbox}&
    \begin{adjustbox}{max width=0.14\textwidth}
    \begin{tabular}{cc}
        \multicolumn{2}{c}{\includegraphics[scale=2]{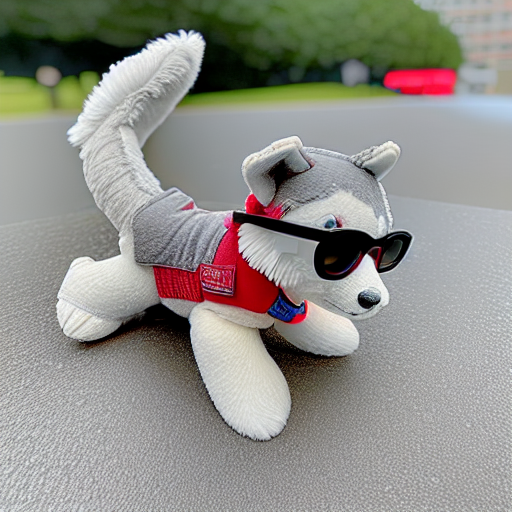}} \\
        \includegraphics{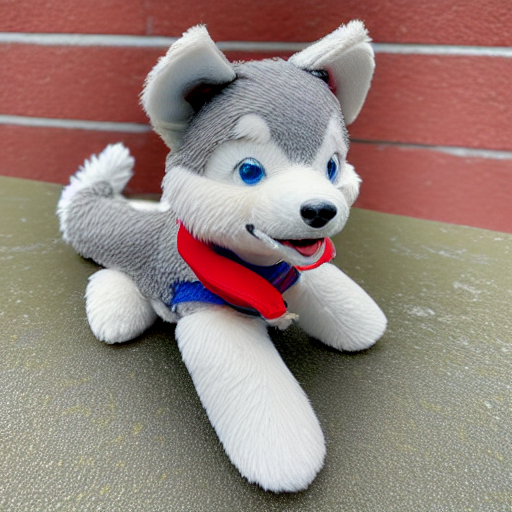} & 
        \includegraphics{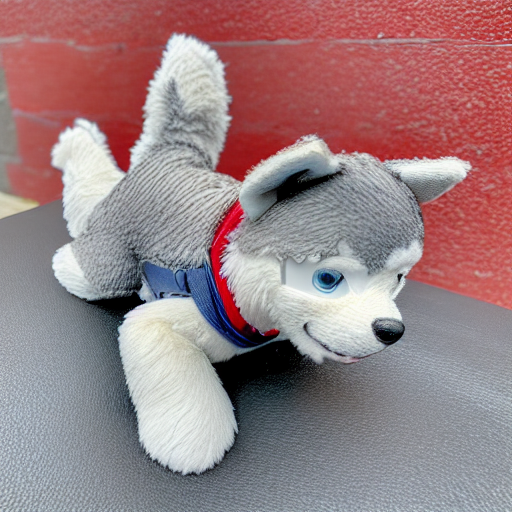} 
    \end{tabular}
    \end{adjustbox} &
    \begin{adjustbox}{max width=0.14\textwidth}
        \begin{tabular}{cc}
        \multicolumn{2}{c}{\includegraphics[scale=2]{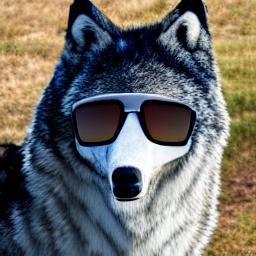}} \\
        \includegraphics{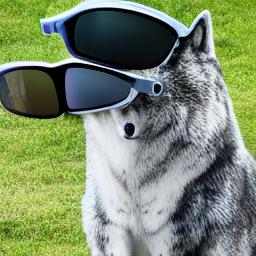} &
        \includegraphics{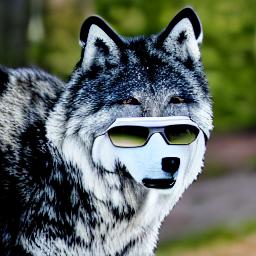} 
    \end{tabular}
    \end{adjustbox} &
    \begin{adjustbox}{max width=0.14\textwidth}
        \begin{tabular}{cc}
        \multicolumn{2}{c}{\includegraphics[scale=2]{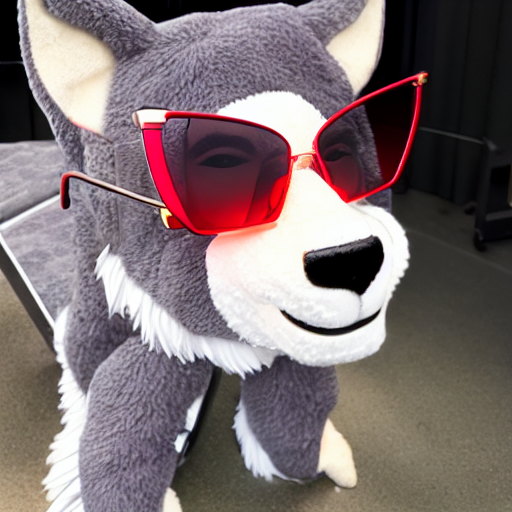}} \\
        \includegraphics{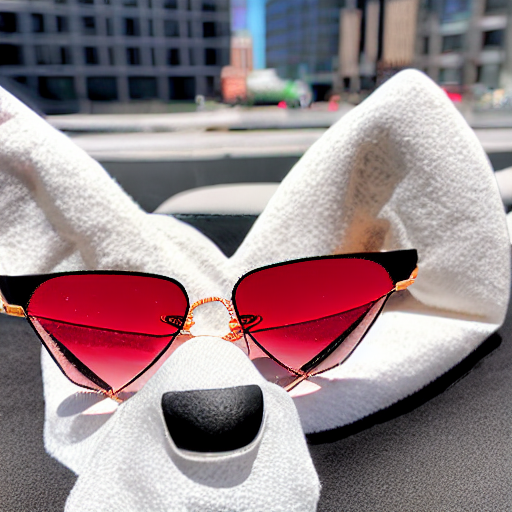} &
        \includegraphics{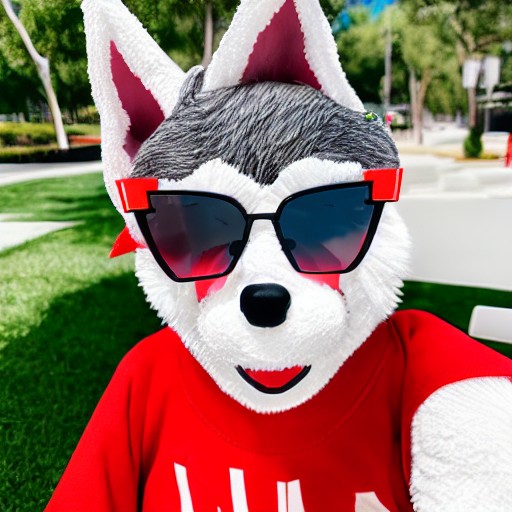}
    \end{tabular}
    \end{adjustbox} &
    \begin{adjustbox}{max width=0.14\textwidth}
    \begin{tabular}{cc}
        \multicolumn{2}{c}{\includegraphics[scale=2]{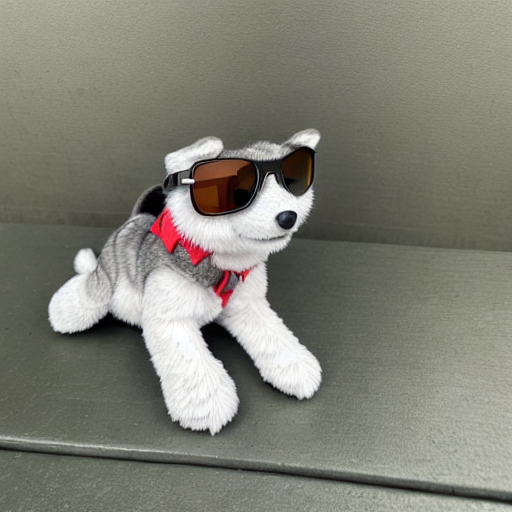}} \\
        \includegraphics{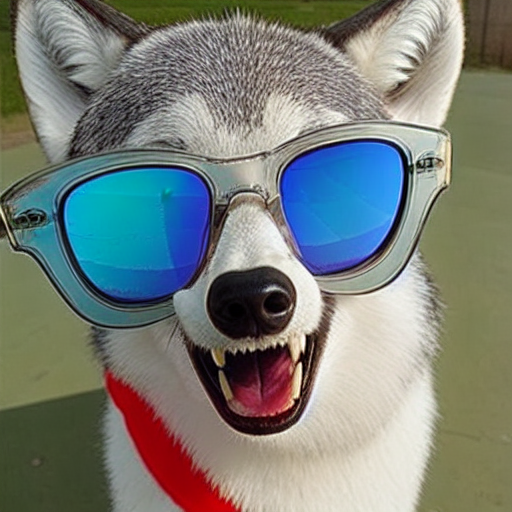} & \includegraphics{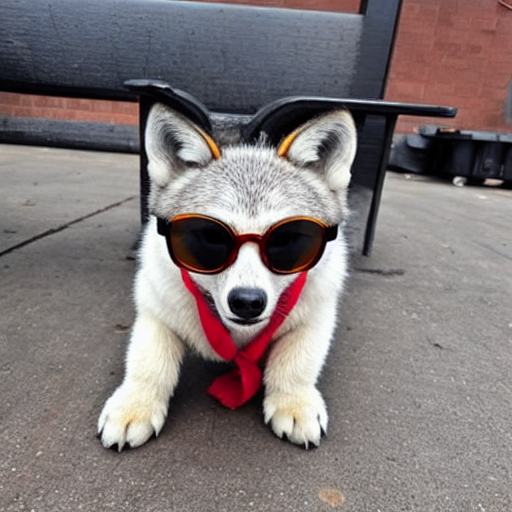} 
    \end{tabular}
    \end{adjustbox} &
    \begin{adjustbox}{max width=0.1\textwidth}
    \begin{tabular}{cc}
         \includegraphics[scale=1]{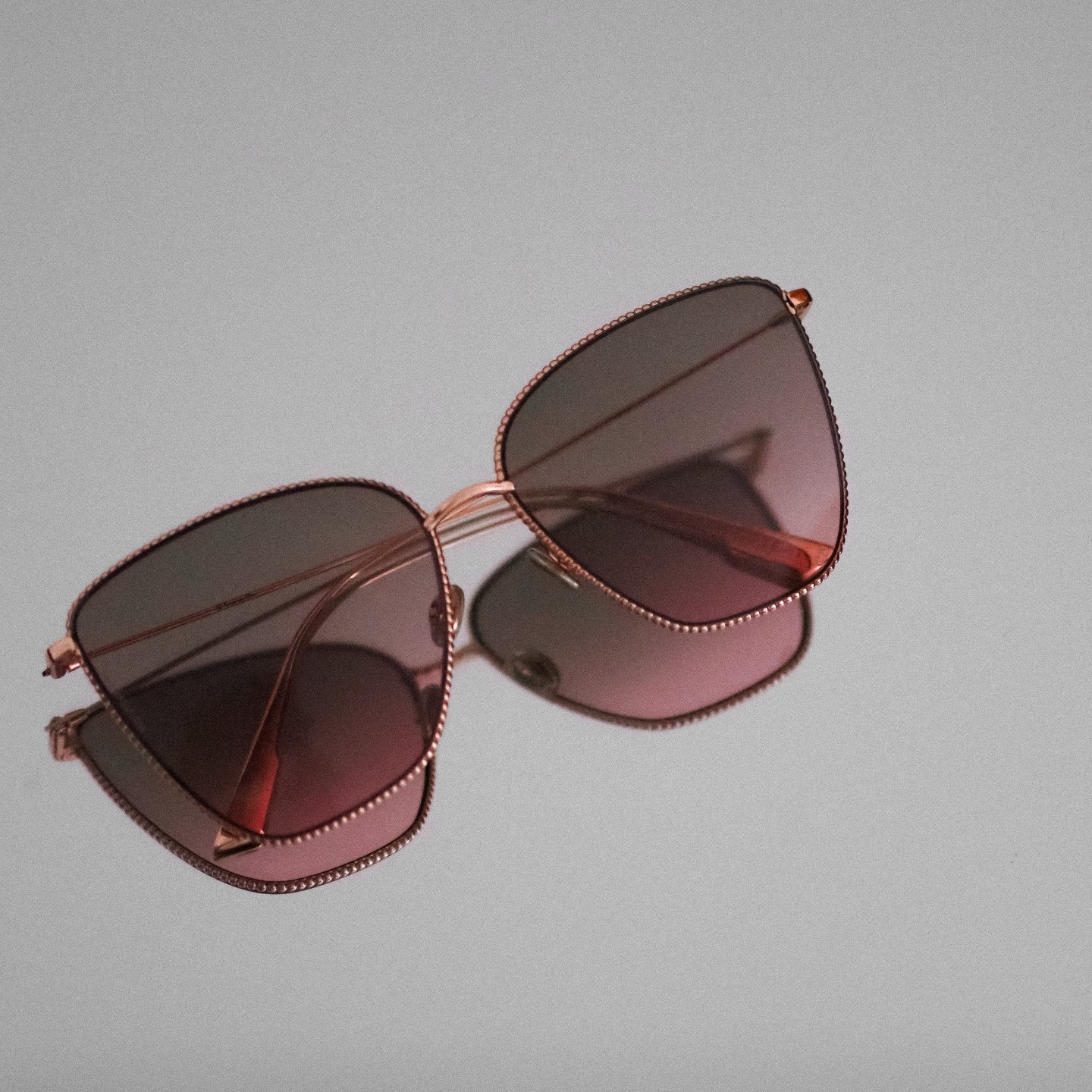}&  \includegraphics[scale=1.3]{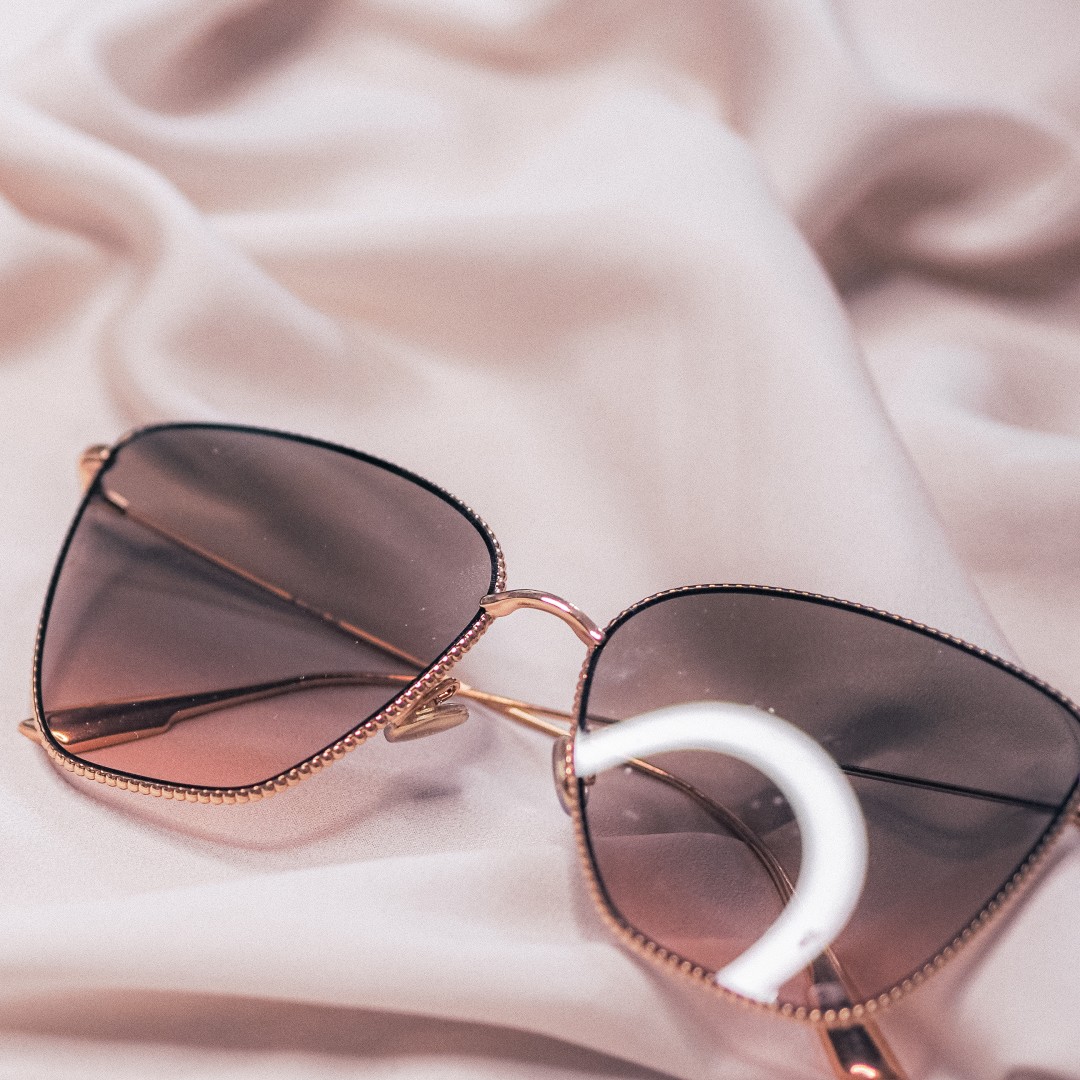}\\
         \\
         \multicolumn{2}{c}{\includegraphics[scale=7]{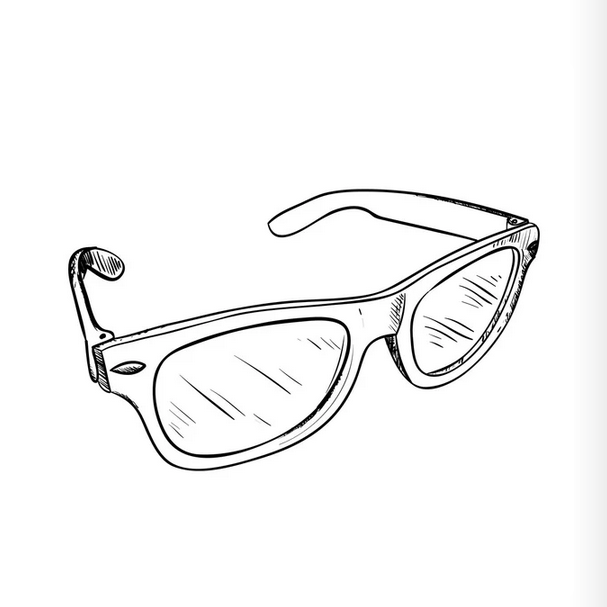}}
    \end{tabular}
    \includegraphics{sec/DF-eval/Condition/sunglass_s.png}
    \end{adjustbox}\\
     & \multicolumn{3}{c}{A wolf plusie* is wearing a sunglass*} & \multicolumn{2}{c}{}
     
    \end{tabular}
    
    \caption{Qualitative comparison with other approaches. The other methods shown here require a text prompt for conditioning. However, our method is able to generate results only using the sketch and a single image (the first image form each of the examples). For the images of plushies, \textit{Custom Diffusion} is trained with multiple concepts. The image of the concept sunglasses is provided in the right column.}
    \label{fig:DF-eval}
\end{figure*}

\section{Experiment}
We evaluate the stages of \textit{DiffMorph} framework, namely the sketch-to-image generation module (section~\ref{sec: condflow}), followed by the multi-concept image personalization model (section~\ref{sec: imgmorph}).
    \subsection{ConditionFlow}\label{sec: condflow}
        \noindent
        \textbf{Dataset:} 
        For the evaluation of the \textit{ConditionFlow} model with the previous \textit{ControlNet} model, both models were trained on a subset of the ImageNet~\cite{deng2009imagenet} dataset using stable diffusion~\cite{Rombach_2022_CVPR} weights. 
        After removing some overlapping classes, we selected 1725 classes, each containing 300 images resulting in about half a million images. Although we acknowledge that the total number of image-class combinations is considerably smaller than that of LAION-5B~\cite{schuhmann2022laion}, due to resource constraints, we adopted a smaller training set. Further, since our training required a sketch for every image, we converted the images into corresponding sketches using the Photo-Sketching~\cite{li2019photo} approach.
        
        \vspace{4mm}
        \noindent
        \textbf{Experimental Setup and Evaluation:} 
        The images selected from the ImageNet~\cite{deng2009imagenet} dataset are of non-uniform sizes. 
        Consequently, we cropped and transformed these to a uniform size of $512\times512$, and created a sketch dataset for these images. 
        To ensure a consistent comparison between both models, \textit{ConditionFlow} and \textit{ControlNet} were trained on a single 80GB Nvidia A100 GPU for 250 hours using the aforementioned dataset.
        For evaluation, the newly trained weights of \textit{ConditionFlow} and \textit{ControlNet} were applied to the TU-Berlin~\cite{eitz2012humans} sketch dataset as a condition to generate images for comparison. 
        We employed the Frechet Inception Distance (FID)~\cite{heusel2017gans} to measure the distribution distance across sketches, along with text-image evaluation metrics such as CLIP scores~\cite{radford2021learning} and CLIP aesthetic scores~\cite{schuhmann2022laion} presented in Table~\ref{tab:CF-eval}.

        \vspace{4mm}
        \noindent
        \textbf{Ablation Studies:} 
        In Figure~\ref{fig:architecture}, three out of four skip connections (marked in red) were removed from the \textit{ControlNet} architecture, which serves as the baseline method.
        Before arriving at this baseline, various settings were empirically explored, involving the removal of different combinations of skip connections. 
        The attempt was to train the model in a way that heavily relies on the given sketch as a condition rather than the input image. 
        \textit{One setting} involved removing all skip connections, while \textit{other setting} retained skip connections from encoder blocks 1 and 4 (refer to Figure~\ref{fig:architecture}(a)). 
        To save time, all four settings were trained with a limited number of images (approximately 120) and only three classes.
        Among these, our finalized \textit{ConditionFlow} model yielded the best results. 
        Notably, in the setting where all skip connections from the input encoder block were removed, the generated outputs exhibited unrealistic colors, and outputs from the other setting did not match the standard. 
        Refer to Figure~\ref{fig:abl-cf} for the generated outputs from different settings.

    \begin{table}[h]
    \centering
    \caption{Evaluation for image generation conditioned by sketch. We have reported FID, CLIP classification accuracy, and CLIP aesthetic Score.}
    \label{tab:CF-eval}
    \begin{adjustbox}{max width=\columnwidth}
        
        \begin{tabular}{lccc}
        \hline
             & FID$~\downarrow$ & \begin{tabular}[c]{@{}c@{}}CLIP Classification \\ Accuracy$~\uparrow$\end{tabular} & \begin{tabular}[c]{@{}c@{}}CLIP Aesthetic \\ Scor$e~\uparrow$\end{tabular} \\ \hline
            ControlNet & 26.76 & 82.21\% & 4.58 \\
            ConditionFlow & 19.96 & 91.14\% & 4.97 \\ \hline
        \end{tabular}
    \end{adjustbox}
    \end{table}

    \subsection{DiffMorph}\label{sec: imgmorph}
        \noindent
        \textbf{Dataset:} 
        For image morphing, we utilized two types of datasets. 
        As mentioned earlier, unlike other methods that require multiple images for each concept, our approach only requires one image per concept. 
        To facilitate a comparison with previous state-of-the-art (SOTA) papers, a few examples from their datasets were selected. 
        Additionally, for the evaluation of our proposed method, a subset of images was sourced from the internet, while others were photographed by us. 
        Regarding sketches, a portion was selected from the TU-Berlin~\cite{eitz2012humans} dataset, and the remainder was hand-drawn by us.

\begin{figure}
    \centering
    \scalebox{0.9}{
    \begin{tabular}{ccc}
    
        Input Image & Input Sketch & Final Image \\
        \includegraphics[width=0.22\columnwidth]{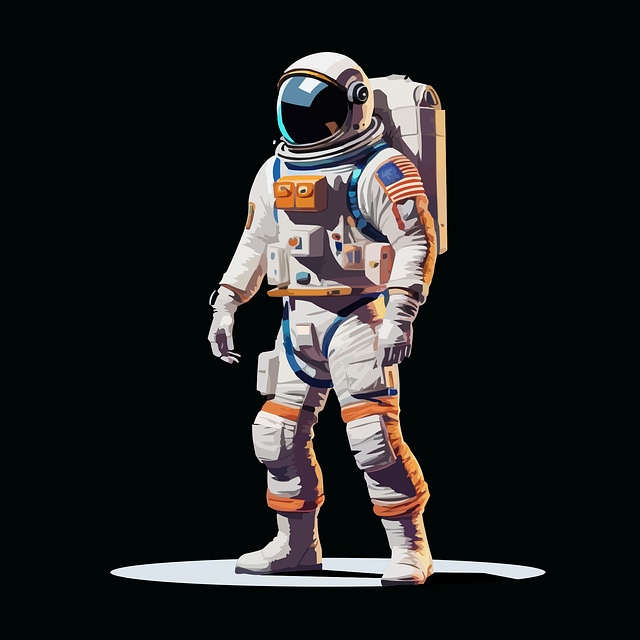} & 
        \includegraphics[width=0.22\columnwidth]{sec/Images/guitar.png} &
        \includegraphics[width=0.22\columnwidth]{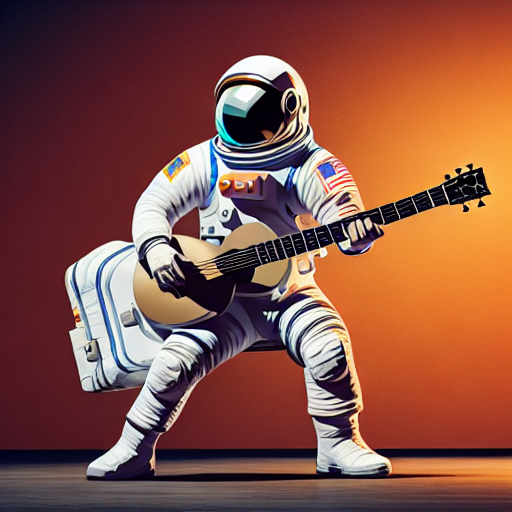} \\
        \includegraphics[width=0.22\columnwidth]{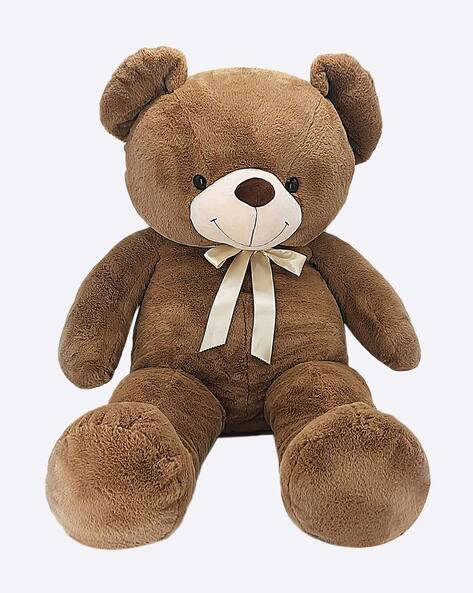} & 
        \includegraphics[width=0.22\columnwidth]{sec/DF-eval/Condition/hat_s.png} &
        \includegraphics[width=0.22\columnwidth]{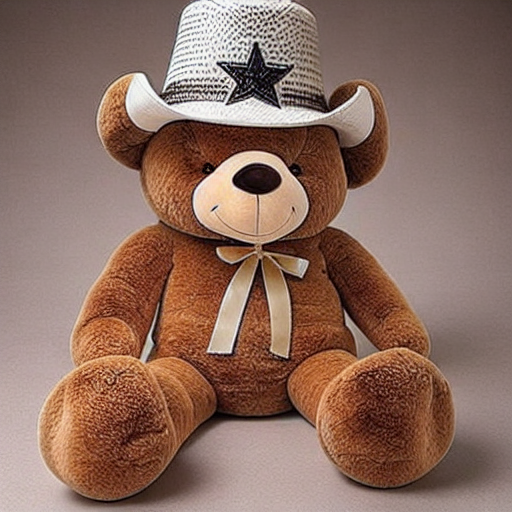} \\
        \includegraphics[width=0.22\columnwidth]{sec/Images/astronaut.png} & 
        \includegraphics[width=0.22\columnwidth]{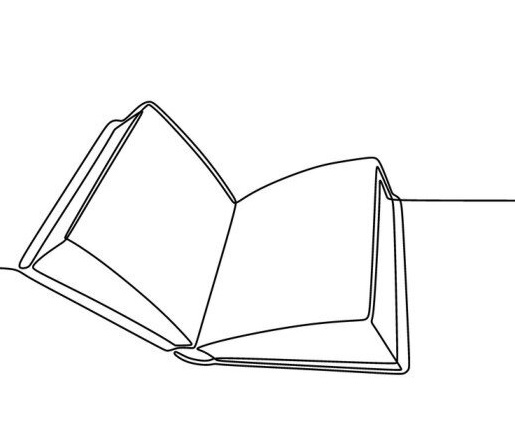} &
        \includegraphics[width=0.22\columnwidth]{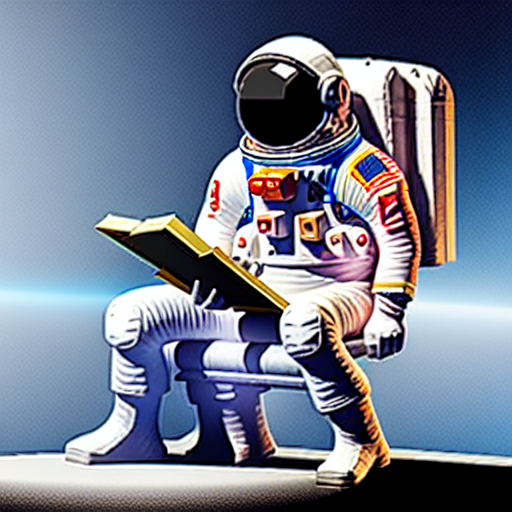} \\
        \includegraphics[width=0.22\columnwidth]{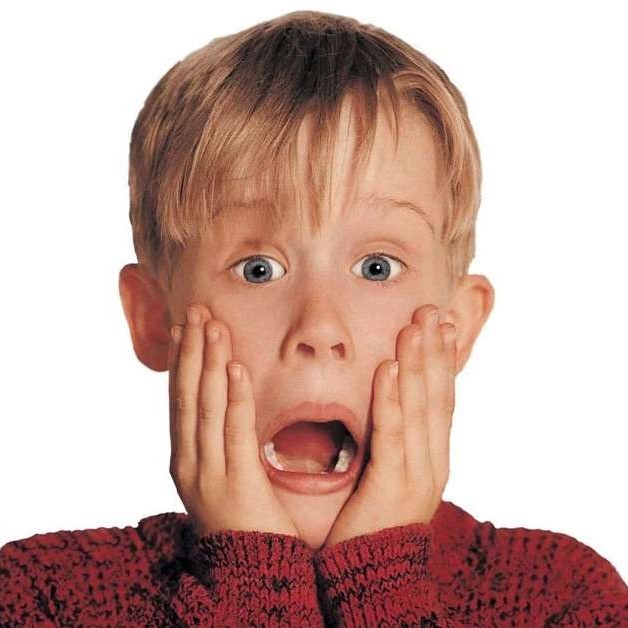} & 
        \includegraphics[width=0.22\columnwidth]{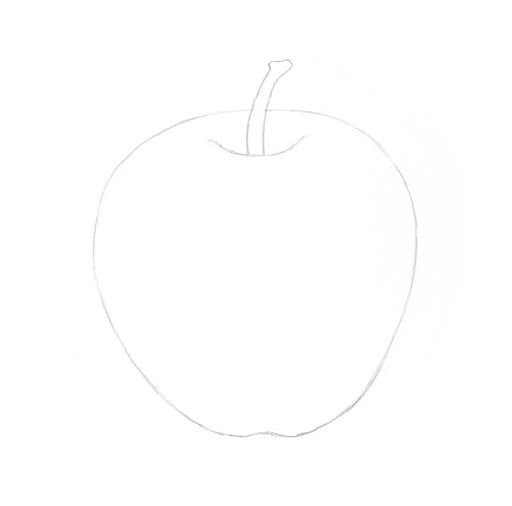} &
        \includegraphics[width=0.22\columnwidth]{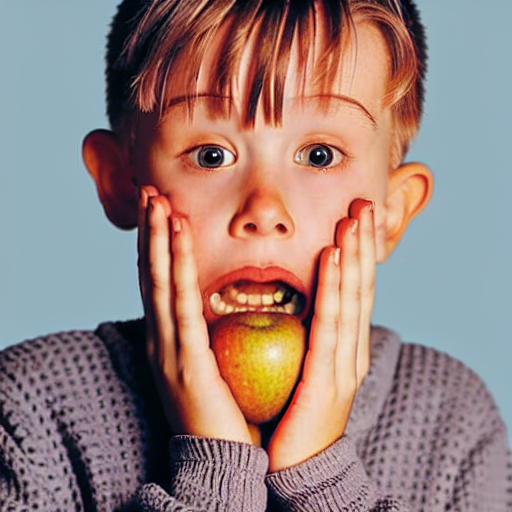} \\
        \includegraphics[width=0.22\columnwidth]{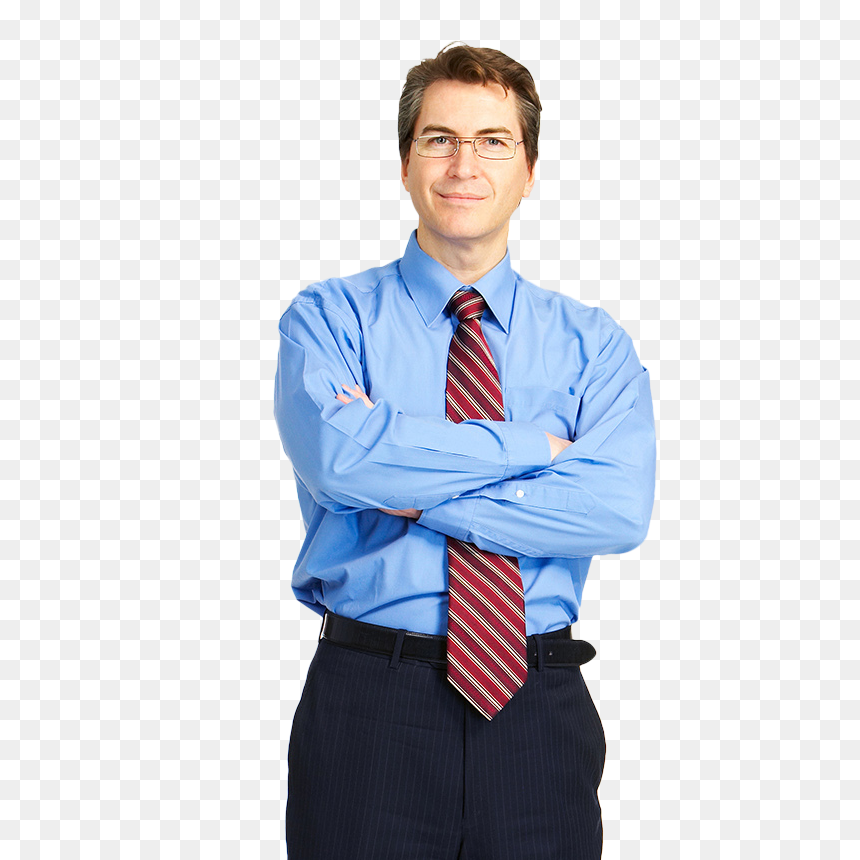} & 
        \includegraphics[width=0.22\columnwidth]{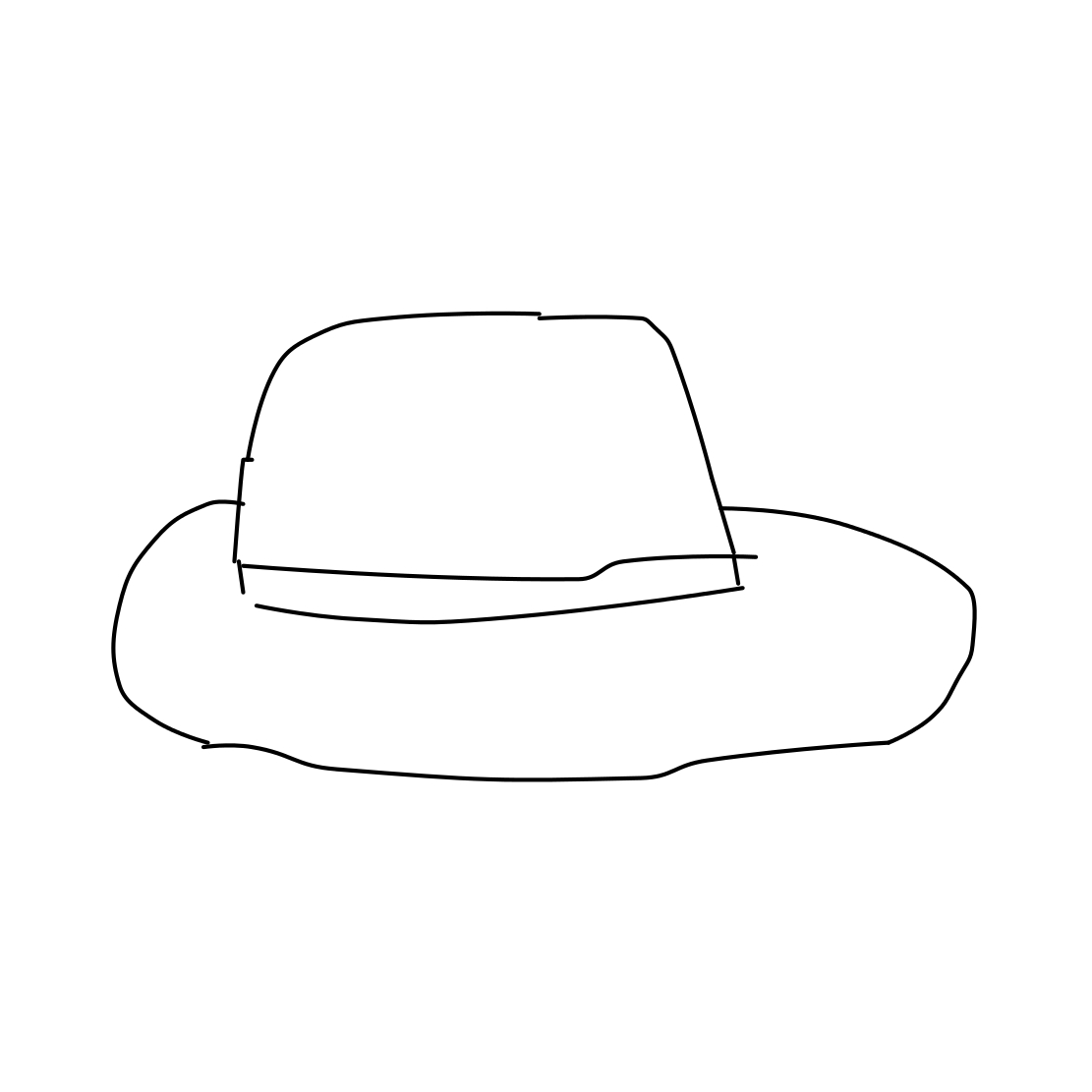} &
        \includegraphics[width=0.22\columnwidth]{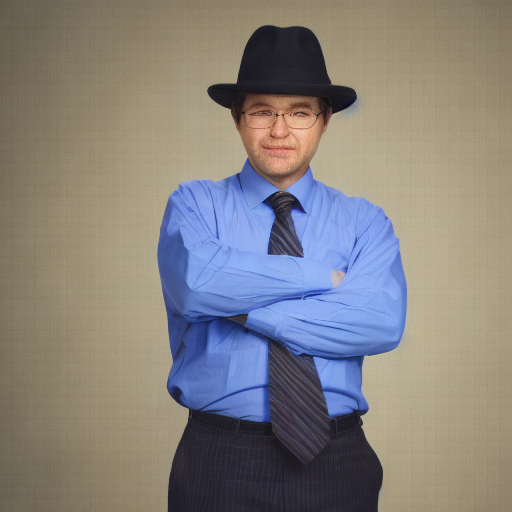}
    
    \end{tabular}
    }
    \begin{tabular}{cccc}
        \includegraphics[width=0.22\columnwidth]{sec/new/person.png} &
        \includegraphics[width=0.22\columnwidth]{sec/new/hat2.png} &
        \includegraphics[width=0.22\columnwidth]{sec/new/book.jpg} &
        \includegraphics[width=0.22\columnwidth]{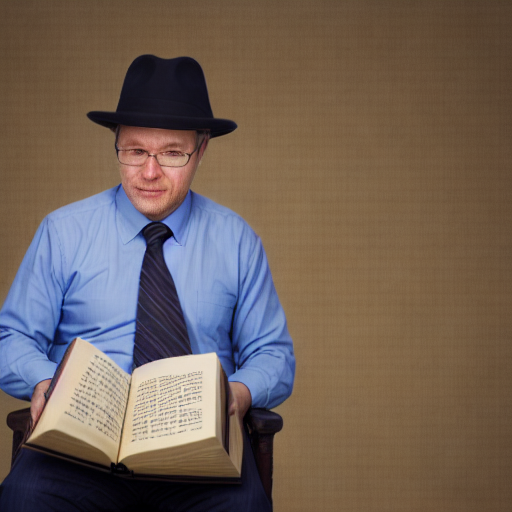}
    \end{tabular}    

    \caption{Some more outputs from DiffMorph method}
    \label{fig:DF-examp}
\end{figure}

        \vspace{4mm}
        \noindent
        \textbf{Experimental Setup and Evaluation:}
        Our experiment on the custom image generation method was executed on the same hardware mentioned before.
        The training time for each concept is approximately 1.5 minutes, and the overall generation time from start to finish is 4-5 minutes. 
        Although some prior SOTA papers, including DreamBooth~\cite{ruiz2023dreambooth} and Textual Inversion~\cite{daras2022multiresolution} focus on a single concept, in our comparison study we include these for comparison with our proposed method and other multi-concept image customization models, such as Custom Diffusion~\cite{kumari2023multi}. 
        For DreamBooth, we used the implementation by XavierXiao~\cite{XiaoBooth}
        Our method does not aim to drastically alter the image layout but rather to displace certain parts to accommodate secondary concepts and morph them together.
        Please refer to Figure~\ref{fig:DF-eval} for a comparison with other custom image generation methods. 
        Additionally, we have presented more of our outputs in Figure~\ref{fig:DF-examp}.

    \begin{table}[h]
    \centering
    \caption{Comparison metric between other model customization methods. We have reported FID, and CLIP Alignment Scores.}
    \label{tab:DM-eval}
    \begin{adjustbox}{max width=\columnwidth}
        
        \begin{tabular}{@{}l|cccc@{}}
            \toprule
             & DreamBooth~\cite{ruiz2023dreambooth} & Textual Inversion~\cite{gal2022image} & Custom Diffusion~\cite{kumari2023multi} & DiffMorph \\ \midrule
            FID Score ($\downarrow$) & 20.51 & 20.36 & 17.21 & 16.35  \\
            Class Alignment ($\uparrow$) & 67\% & 78.1\% & 79.5\% & 81.3\% \\
            Image Alignment ($\uparrow$) & 82.7\% & 74.8\% & 77.5\% & 79.7\% \\ \bottomrule
        \end{tabular}
    \end{adjustbox}
    \end{table}

\begin{figure}
    \centering
    \includegraphics[width=\columnwidth]{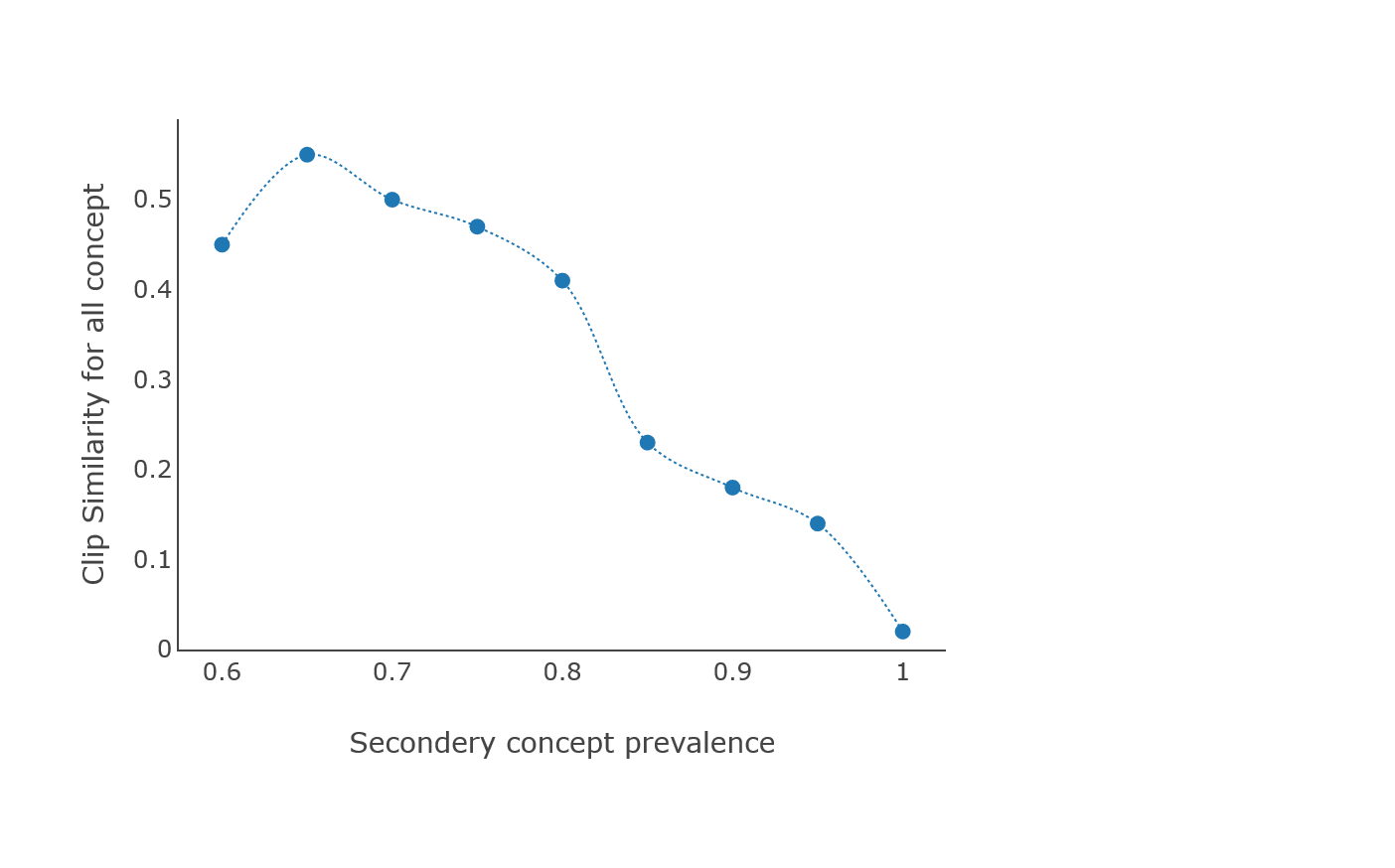}
    \caption{Ablation Study of DiffMorph. We tested different ratios of the prevalence of all of the concepts and generated the clip score regarding each concept. If the CLIP score of any concept is less than 0.1, we are assigning zero for that particular generation.}
    \label{fig:DF-abl}
\end{figure}

        \vspace{4mm}
        \noindent
        \textbf{Ablation Studies:} 
        In our fine-tuning approach, we mitigate the effects of overfitting by limiting iterations of the fine-tuned process.
        We tested a range of different key parameters for reconstruction loss.
        We also found that unrestrained fine-tuning leads to overfitting and prevents fine-tuning multiple concepts.
        We conducted tests with different key parameters for fine-tuning multiple concepts. 
        Figure~\ref{fig:DF-abl} illustrates the CLIP scores of the images generated under various key parameters.
\section{Conclusion and Future-work}
In this paper, we present \textit{DiffMorph}, a customized image generation method conditioned with sketches.
Instead of using text prompts from the user, our method establishes a correlation between the initial image and a sequence of provided sketches for conditioning.
This approach generates a morphed image with a high level of visual fidelity, relying on the provided image inputs alone.
We fine-tune the stable diffusion-based denoiser model in a controlled manner to synthesize multiple concepts while constraining the denoiser model to prevent overfitting.

\vspace{4mm}
\noindent   
\textbf{Limitation and Future Work:}
Our method has several limitations. 
Firstly, since the system autonomously generates relationships between provided concepts, the resulting images may not always align with the user's expectations.
This situation mostly occurs when the generated relation does not align with the users, leading to an unexpected image generation.
However, this limitation could be overcome by providing multiple possible relations as a choice.
Instead of generating embeddings on the topmost relation, suggesting a few options to the user would lead to the desired image generation.
Secondly, users lack the ability to precisely specify the location for secondary concepts; this determination is made by the system based on the generated relationships.
We are currently working on a further method where the user can select the region in the primary image where the condition would be placed to provide better consistency.
Thirdly, introducing a drastic change in the position of the primary condition may lead to challenges in image generation.
Despite the outcomes presented earlier, there remains room for improvement in future iterations of our approach.
{
    \small
    \bibliographystyle{unsrt}
    \bibliography{main}

\begin{thebibliography}{10}

\bibitem{dhariwal2021diffusion}
Prafulla Dhariwal and Alexander Nichol.
\newblock Diffusion models beat gans on image synthesis.
\newblock {\em Advances in neural information processing systems}, 34:8780--8794, 2021.

\bibitem{kingma2021variational}
Diederik Kingma, Tim Salimans, Ben Poole, and Jonathan Ho.
\newblock Variational diffusion models.
\newblock {\em Advances in neural information processing systems}, 34:21696--21707, 2021.

\bibitem{Rombach_2022_CVPR}
Robin Rombach, Andreas Blattmann, Dominik Lorenz, Patrick Esser, and Bj\"orn Ommer.
\newblock High-resolution image synthesis with latent diffusion models.
\newblock In {\em Proceedings of the IEEE/CVF Conference on Computer Vision and Pattern Recognition (CVPR)}, pages 10684--10695, June 2022.

\bibitem{ruiz2023dreambooth}
Nataniel Ruiz, Yuanzhen Li, Varun Jampani, Yael Pritch, Michael Rubinstein, and Kfir Aberman.
\newblock Dreambooth: Fine tuning text-to-image diffusion models for subject-driven generation.
\newblock In {\em Proceedings of the IEEE/CVF Conference on Computer Vision and Pattern Recognition}, pages 22500--22510, 2023.

\bibitem{kumari2023multi}
Nupur Kumari, Bingliang Zhang, Richard Zhang, Eli Shechtman, and Jun-Yan Zhu.
\newblock Multi-concept customization of text-to-image diffusion.
\newblock In {\em Proceedings of the IEEE/CVF Conference on Computer Vision and Pattern Recognition}, pages 1931--1941, 2023.

\bibitem{cohen2022my}
Niv Cohen, Rinon Gal, Eli~A Meirom, Gal Chechik, and Yuval Atzmon.
\newblock “this is my unicorn, fluffy”: Personalizing frozen vision-language representations.
\newblock In {\em European Conference on Computer Vision}, pages 558--577. Springer, 2022.

\bibitem{pmlr-v37-sohl-dickstein15}
Jascha Sohl-Dickstein, Eric Weiss, Niru Maheswaranathan, and Surya Ganguli.
\newblock Deep unsupervised learning using nonequilibrium thermodynamics.
\newblock In {\em Proceedings of the 32nd International Conference on Machine Learning}, volume~37 of {\em Proceedings of Machine Learning Research}, pages 2256--2265, Lille, France, 07--09 Jul 2015. PMLR.

\bibitem{balaji2022ediffi}
Yogesh Balaji, Seungjun Nah, Xun Huang, Arash Vahdat, Jiaming Song, Karsten Kreis, Miika Aittala, Timo Aila, Samuli Laine, Bryan Catanzaro, et~al.
\newblock ediffi: Text-to-image diffusion models with an ensemble of expert denoisers.
\newblock {\em arXiv preprint arXiv:2211.01324}, 2022.

\bibitem{nichol2021glide}
Alex Nichol, Prafulla Dhariwal, Aditya Ramesh, Pranav Shyam, Pamela Mishkin, Bob McGrew, Ilya Sutskever, and Mark Chen.
\newblock Glide: Towards photorealistic image generation and editing with text-guided diffusion models.
\newblock {\em arXiv preprint arXiv:2112.10741}, 2021.

\bibitem{ramesh2022hierarchical}
Aditya Ramesh, Prafulla Dhariwal, Alex Nichol, Casey Chu, and Mark Chen.
\newblock Hierarchical text-conditional image generation with clip latents.
\newblock {\em arXiv preprint arXiv:2204.06125}, 1(2):3, 2022.

\bibitem{chung1967markov}
Kai~Lai Chung.
\newblock Markov chains.
\newblock {\em Springer-Verlag, New York}, 1967.

\bibitem{esser2021taming}
Patrick Esser, Robin Rombach, and Bjorn Ommer.
\newblock Taming transformers for high-resolution image synthesis.
\newblock In {\em Proceedings of the IEEE/CVF conference on computer vision and pattern recognition}, pages 12873--12883, 2021.

\bibitem{radford2021learning}
Alec Radford, Jong~Wook Kim, Chris Hallacy, Aditya Ramesh, Gabriel Goh, Sandhini Agarwal, Girish Sastry, Amanda Askell, Pamela Mishkin, Jack Clark, et~al.
\newblock Learning transferable visual models from natural language supervision.
\newblock In {\em International conference on machine learning}, pages 8748--8763. PMLR, 2021.

\bibitem{podell2023sdxl}
Dustin Podell, Zion English, Kyle Lacey, Andreas Blattmann, Tim Dockhorn, Jonas M{\"u}ller, Joe Penna, and Robin Rombach.
\newblock Sdxl: Improving latent diffusion models for high-resolution image synthesis.
\newblock {\em arXiv preprint arXiv:2307.01952}, 2023.

\bibitem{NEURIPS2022_ec795aea}
Chitwan Saharia, William Chan, Saurabh Saxena, Lala Li, Jay Whang, Emily~L Denton, Kamyar Ghasemipour, Raphael Gontijo~Lopes, Burcu Karagol~Ayan, Tim Salimans, Jonathan Ho, David~J Fleet, and Mohammad Norouzi.
\newblock Photorealistic text-to-image diffusion models with deep language understanding.
\newblock In {\em Advances in Neural Information Processing Systems}, volume~35, pages 36479--36494. Curran Associates, Inc., 2022.

\bibitem{dalle}
OpenAI.
\newblock Dall·e 2.
\newblock \url{https://openai.com/dall-e-2}, 2022.

\bibitem{midj}
Inc. Midjourney.
\newblock Midjourney.
\newblock \url{https://www.midjourney.com/}, 2022.

\bibitem{ruiz2023hyperdreambooth}
Nataniel Ruiz, Yuanzhen Li, Varun Jampani, Wei Wei, Tingbo Hou, Yael Pritch, Neal Wadhwa, Michael Rubinstein, and Kfir Aberman.
\newblock Hyperdreambooth: Hypernetworks for fast personalization of text-to-image models.
\newblock {\em arXiv preprint arXiv:2307.06949}, 2023.

\bibitem{kawar2023imagic}
Bahjat Kawar, Shiran Zada, Oran Lang, Omer Tov, Huiwen Chang, Tali Dekel, Inbar Mosseri, and Michal Irani.
\newblock Imagic: Text-based real image editing with diffusion models.
\newblock In {\em Proceedings of the IEEE/CVF Conference on Computer Vision and Pattern Recognition}, pages 6007--6017, 2023.

\bibitem{daras2022multiresolution}
Giannis Daras and Alexandros~G. Dimakis.
\newblock Multiresolution textual inversion, 2022.

\bibitem{gal2022image}
Rinon Gal, Yuval Alaluf, Yuval Atzmon, Or~Patashnik, Amit~H. Bermano, Gal Chechik, and Daniel Cohen-Or.
\newblock An image is worth one word: Personalizing text-to-image generation using textual inversion, 2022.

\bibitem{tewel2023key}
Yoad Tewel, Rinon Gal, Gal Chechik, and Yuval Atzmon.
\newblock Key-locked rank one editing for text-to-image personalization.
\newblock In {\em ACM SIGGRAPH 2023 Conference Proceedings}, pages 1--11, 2023.

\bibitem{choi2018stargan}
Yunjey Choi, Minje Choi, Munyoung Kim, Jung-Woo Ha, Sunghun Kim, and Jaegul Choo.
\newblock Stargan: Unified generative adversarial networks for multi-domain image-to-image translation.
\newblock In {\em Proceedings of the IEEE conference on computer vision and pattern recognition}, pages 8789--8797, 2018.

\bibitem{isola2017image}
Phillip Isola, Jun-Yan Zhu, Tinghui Zhou, and Alexei~A Efros.
\newblock Image-to-image translation with conditional adversarial networks.
\newblock In {\em Proceedings of the IEEE conference on computer vision and pattern recognition}, pages 1125--1134, 2017.

\bibitem{park2019semantic}
Taesung Park, Ming-Yu Liu, Ting-Chun Wang, and Jun-Yan Zhu.
\newblock Semantic image synthesis with spatially-adaptive normalization.
\newblock In {\em Proceedings of the IEEE/CVF conference on computer vision and pattern recognition}, pages 2337--2346, 2019.

\bibitem{chen2021pre}
Hanting Chen, Yunhe Wang, Tianyu Guo, Chang Xu, Yiping Deng, Zhenhua Liu, Siwei Ma, Chunjing Xu, Chao Xu, and Wen Gao.
\newblock Pre-trained image processing transformer.
\newblock In {\em Proceedings of the IEEE/CVF conference on computer vision and pattern recognition}, pages 12299--12310, 2021.

\bibitem{ramesh2021zero}
Aditya Ramesh, Mikhail Pavlov, Gabriel Goh, Scott Gray, Chelsea Voss, Alec Radford, Mark Chen, and Ilya Sutskever.
\newblock Zero-shot text-to-image generation.
\newblock In {\em International Conference on Machine Learning}, pages 8821--8831. PMLR, 2021.

\bibitem{saharia2022palette}
Chitwan Saharia, William Chan, Huiwen Chang, Chris Lee, Jonathan Ho, Tim Salimans, David Fleet, and Mohammad Norouzi.
\newblock Palette: Image-to-image diffusion models.
\newblock In {\em ACM SIGGRAPH 2022 Conference Proceedings}, pages 1--10, 2022.

\bibitem{richardson2021encoding}
Elad Richardson, Yuval Alaluf, Or~Patashnik, Yotam Nitzan, Yaniv Azar, Stav Shapiro, and Daniel Cohen-Or.
\newblock Encoding in style: a stylegan encoder for image-to-image translation.
\newblock In {\em Proceedings of the IEEE/CVF conference on computer vision and pattern recognition}, pages 2287--2296, 2021.

\bibitem{gal2022stylegan}
Rinon Gal, Or~Patashnik, Haggai Maron, Amit~H Bermano, Gal Chechik, and Daniel Cohen-Or.
\newblock Stylegan-nada: Clip-guided domain adaptation of image generators.
\newblock {\em ACM Transactions on Graphics (TOG)}, 41(4):1--13, 2022.

\bibitem{karras2019style}
Tero Karras, Samuli Laine, and Timo Aila.
\newblock A style-based generator architecture for generative adversarial networks.
\newblock In {\em Proceedings of the IEEE/CVF conference on computer vision and pattern recognition}, pages 4401--4410, 2019.

\bibitem{patashnik2021styleclip}
Or~Patashnik, Zongze Wu, Eli Shechtman, Daniel Cohen-Or, and Dani Lischinski.
\newblock Styleclip: Text-driven manipulation of stylegan imagery.
\newblock In {\em Proceedings of the IEEE/CVF International Conference on Computer Vision}, pages 2085--2094, 2021.

\bibitem{zhang2023adding}
Lvmin Zhang and Maneesh Agrawala.
\newblock Adding conditional control to text-to-image diffusion models.
\newblock {\em arXiv preprint arXiv:2302.05543}, 2023.

\bibitem{BANKMAN2009261}
Isaac~N. Bankman, Thomas~S. Spisz, and Sotiris Pavlopoulos.
\newblock {\em Chapter 15 - Two-Dimensional Shape and Texture Quantification}.
\newblock Academic Press, Burlington, second edition edition, 2009.

\bibitem{cciccek20163d}
{\"O}zg{\"u}n {\c{C}}i{\c{c}}ek, Ahmed Abdulkadir, Soeren~S Lienkamp, Thomas Brox, and Olaf Ronneberger.
\newblock 3d u-net: learning dense volumetric segmentation from sparse annotation.
\newblock In {\em Medical Image Computing and Computer-Assisted Intervention--MICCAI 2016: 19th International Conference, Athens, Greece, October 17-21, 2016, Proceedings, Part II 19}, pages 424--432. Springer, 2016.

\bibitem{he2016deep}
Kaiming He, Xiangyu Zhang, Shaoqing Ren, and Jian Sun.
\newblock Deep residual learning for image recognition.
\newblock In {\em Proceedings of the IEEE conference on computer vision and pattern recognition}, pages 770--778, 2016.

\bibitem{brooks2023instructpix2pix}
Tim Brooks, Aleksander Holynski, and Alexei~A Efros.
\newblock Instructpix2pix: Learning to follow image editing instructions.
\newblock In {\em Proceedings of the IEEE/CVF Conference on Computer Vision and Pattern Recognition}, pages 18392--18402, 2023.

\bibitem{speer2017conceptnet}
Robyn Speer, Joshua Chin, and Catherine Havasi.
\newblock Conceptnet 5.5: An open multilingual graph of general knowledge.
\newblock In {\em Proceedings of the AAAI conference on artificial intelligence}, volume~31, 2017.

\bibitem{deng2009imagenet}
Jia Deng, Wei Dong, Richard Socher, Li-Jia Li, Kai Li, and Li~Fei-Fei.
\newblock Imagenet: A large-scale hierarchical image database.
\newblock In {\em 2009 IEEE conference on computer vision and pattern recognition}, pages 248--255. Ieee, 2009.

\bibitem{schuhmann2022laion}
Christoph Schuhmann, Romain Beaumont, Richard Vencu, Cade Gordon, Ross Wightman, Mehdi Cherti, Theo Coombes, Aarush Katta, Clayton Mullis, Mitchell Wortsman, et~al.
\newblock Laion-5b: An open large-scale dataset for training next generation image-text models.
\newblock {\em Advances in Neural Information Processing Systems}, 35:25278--25294, 2022.

\bibitem{li2019photo}
Mengtian Li, Zhe Lin, Radomir Mech, Ersin Yumer, and Deva Ramanan.
\newblock Photo-sketching: Inferring contour drawings from images.
\newblock In {\em 2019 IEEE Winter Conference on Applications of Computer Vision (WACV)}, pages 1403--1412. IEEE, 2019.

\bibitem{eitz2012humans}
Mathias Eitz, James Hays, and Marc Alexa.
\newblock How do humans sketch objects?
\newblock {\em ACM Transactions on graphics (TOG)}, 31(4):1--10, 2012.

\bibitem{heusel2017gans}
Martin Heusel, Hubert Ramsauer, Thomas Unterthiner, Bernhard Nessler, and Sepp Hochreiter.
\newblock Gans trained by a two time-scale update rule converge to a local nash equilibrium.
\newblock {\em Advances in neural information processing systems}, 30, 2017.

\bibitem{XiaoBooth}
Inc. XavierXiao.
\newblock Dreambooth on stable diffusion.
\newblock \url{https://github.com/XavierXiao/Dreambooth-Stable-Diffusion}, 2022.

\end{thebibliography}
}

\end{document}